\documentclass[review]{elsarticle}

\usepackage{lineno,hyperref}
\modulolinenumbers[5]

\journal{Journal of \LaTeX\ Templates}

\graphicspath{{figure/}}   

\usepackage{amsmath,amsfonts,amssymb,bm}
\usepackage{subfigure}
\usepackage{multirow}
\usepackage{booktabs}
\usepackage{lscape}
\usepackage{algorithm, algorithmic}
\usepackage{color} 
\usepackage{appendix}

\begin{document}

\begin{frontmatter}

\title{{Joint Deep Reversible Regression Model and Physics-Informed Unsupervised Learning} for Temperature Field Reconstruction}

\author{Zhiqiang Gong}
\author{Weien~Zhou}
\author{Jun~Zhang}
\author{Wei~Peng}
\author{Wen Yao}

\address{Defense Innovation Institute, Academy of Military Sciences, Beijing 100000, China}



\begin{abstract}
{Temperature monitoring over heat source components in  engineering systems, such as the  energy system, electronic equipments, becomes essential to  guarantee the working performance of these components.}
However, prior methods, which mainly use the interpolate estimation to reconstruct {the overall temperature field} from limited monitoring points, require large amounts of temperature tensors for an accurate estimation. This {may affect the availability and reliability of the system. 
To solve the problem, this work develops a novel reconstruction method which joints the deep reversible regression model and
physics-informed unsupervised learning for temperature field reconstruction of heat-source systems (TFR-HSS).}
{Firstly,} we {define the TFR-HSS mathematically,  numerically model the system with discrete grids}, and hence transform the task as an image-to-image regression problem.
 Then, this work develops the deep reversible regression model which can better {learn physical information,} {especially over the area near the boundaries of the system}.
Finally, {this work proposes the physics-informed reconstruction loss with the physical characteristics of the system and learns the deep model without labelled samples.}
{Experimental studies have conducted over typical two-dimensional heat-source systems to validate the effectiveness of the proposed method. 
 Under the proposed  method, the mean average error of the constructed temperature field can achieve about 0.1K, 50\% lower than other methods. Besides, the proposed method takes 5.2 ms per sample for inference which can provide real-time predictions.
}
\end{abstract}


\end{frontmatter}


\section{Introduction}
\label{intro}
Heat management plays an important role in heat-source systems where heat may be generated internally \cite{04,05}, especially over systems consisting of components (electronic devices) with smaller size and higher power density \cite{ij_1,ij_2}. It can provide the working  status, and even improve the life time and reliability of the system.
Temperature monitoring, which can monitor and provide the real-time operating temperature, tends to be an irreplaceable process in heat management system.
Generally, temperature transducer \cite{tensor1,tensor2} is used as the engineering device  to collect and convert the temperature information to available signal as output.
However, too many tranducers would  severely influence the availability and reliability of the system  and sharply increase the  monitoring cost. Generally, only limited number of transducers can be assigned for real-time temperature monitoring in real-world system.  This would multiply the difficulties to obtain the whole temperature field of the heat-source systems.
 {Therefore, {\it Temperature field reconstruction task of heat-source systems (TFR-HSS)}  \cite{06, tfrd_gong} has become an urgent but challenging task in engineering systems, such as printed circuit  boards (PCBs) in electronic  equipments \cite{nr14}, honeycomb panels in spacecraft \cite{nr12}, energy system \cite{nr13}, launch vehicle \cite{nr15}, and others.}

Traditional methods utilize the interpolation methods for temperature field reconstruction \cite{01}.
Interpolation is a type of estimation to reconstruct the temperature of interest area within the range of a discrete set of known monitoring temperature points, such as the linear interpolation, polynomial interpolation, spline interpolation \cite{interpolation}. These interpolation methods mainly take advantage of the fixed local and global relative  correlation, and are generally fast to calculate and easy to operate. However, they cannot be adaptive to learn the correlation between the interest points and the monitoring  points. Their performance is guaranteed by the prior knowledge. Besides, they ignore the physical characteristics of the temperature field and  thus provide the constructed temperature field with high prediction error.

Faced with these circumstances, {most of the  researches take advantage of surrogate modeling \cite{r2} based on traditional regression models \cite{ij_5},  such as  derivative Gaussian filter \cite{nr2}, Gaussian process regression (Kriging method) \cite{nr7, kriging}, support vector regression (SVR)\cite{nr10, svr},  polynomial regression \cite{polynomial},  {Gaussian Radial
Basis Functions (GRBF)-based} kernel regression \cite{03}, gappy proper orthogonal decomposition (GOP) \cite{02}, Artificial Neural Networks (ANNs) \cite{r1, nr6, nr8}, in the literature of field reconstruction.}
{Benefited from machine learning,} these surrogate models can learn the nonlinear physical correlation adaptively and present stronger generalization and more universal.
Nevertheless, these methods which consist of limited number of parameters cannot be well applied to the ultra-dimensional TFR-HSS task.  For example, {the Random Vector Functional link (RVFL) network in \cite{r1}, as well as the GMDH neural network in \cite{nr9, nr11}, is only with single hidden layer feedforward neural network which has limited ability for  end-to-end thermal approximation.} Therefore, exploring surrogate models with higher representational ability and better performance in describing linear and nonlinear physical information is essential for current ultra-dimensional task.


In recent years, deep neural networks (DNNs), as a non-linear ultra-dimensional fitting \cite{ener}, have achieved good performance in extracting high-level information in many fields, such as image classification \cite{gong2019,gong2021}.  {As a representative, convolutional neural networks (CNNs) \cite{nr3, nr5}, which can extract both the local and global information and describe the complex physical correlation between different input pixels \cite{nr4}, have already been applied in thermal analysis of electronic systems \cite{r3}.}
Considering the ultra-dimension and nonlinear physical characteristics of the mapping from observation data to temperature field, deep models with potential ability to learn the latent complex physical correlation is an effective method  for the specific task \cite{tfrd_gong}.
In this work, due to the good performance, the CNNs will be used as the deep surrogate model to extract {the physical  correlation \cite{tfrd_gong, ij_6}.}

{However, the state-of-the-art CNN models for TFR-HSS task in prior works, such as \cite{tfrd_gong}, mainly apply the existing vanilla deep regression models. There still exist two difficulties faced in prior deep surrogate models in TFR-HSS task.}
\begin{itemize}
\item Appropriate numerical  modeling which can fit for deep surrogate models is required for TFR-HSS task. Besides, a proper deep surrogate model which can capture physical characteristics of temperature field is also needed to reconstruct the field given limited monitoring points.
\item  General CNNs are data-driven where large amounts of labeled samples are required {for the training of the deep model \cite{nr1}}. However, for TFR-HSS task, the temperature field corresponding to the given monitoring value is expensive to obtain, sometimes even unavailable.
\end{itemize}
These problems would make it more challenging to apply deep surrogate models for TFR-HSS task.

To solve the first problem, this work models the heat-source system as a two-dimensional domain and {discretize the domain with $N\times N$ grid.}  Thus, the information from monitoring points and the temperature field of the system can be numerically modelled as $N\times N$ matrix, respectively.  Then, the TFR-HSS task is transformed as the optimization problem to find the mapping from the matrix of monitoring points to the whole temperature field.
The problem can be seen as the image-to-image regression  problem and solved by deep regression models.
{However, due to the law of forward and backward propagation of convolutional layers in general models \cite{nn} and the law of heat conduction \cite{heattransfer}, general models, such as the fully convolutional networks (FCN) \cite{fcn}, cannot provide expected reconstruction performance, especially in the area near the upside and rightside boundaries.
Considering both the laws and fitting the deep models for the task, this work develops the reversible regression models for this specific TFR-HSS task which can solve the field reconstruction near boundaries  through twice encoder-decoder process and the diagonal flip operation.}

To solve the latter one, this work utilizes the physical characteristics of temperature field and develops the physics-informed deep learning methods to train the deep model unsupervisedly for the TFR-HSS task.
Generally, model-based deep learning methods, which can use the model prior of the data for the training of the deep model, is an effective way for the deep learning with limited samples \cite{model,gong_manifold}.
These model priors, such as the statistical property, topological structure correlation, can assist the training of deep models. 
As a specific form of model priors, physics properties can also be used as the model prior for the training of the deep surrogate model.
Many prior works have focused on such physics prior for the learning of neural networks to solve the engineering problem, such as the elastodynamics \cite{physics_2}, fluid dynamics \cite{physics_5}, thermochemical curing process \cite{physics_7}. Even though the  applied fields vary from each other, the main idea of such problem is to formulate the training loss based on the partial differential equation (PDE) from the physical process \cite{physics_4, physics_6}.
{Faced with the TFR-HSS task, few efforts has been made to construct the physics-informed reconstruction loss and provide the reconstruction model without labelled samples.}

{As for the task, the physical properties,} including the heat conduction, boundary conditions, spatial smoothness, can be used for the training of the surrogate model. 
First, considering the heat conduction correlation between different points in the system, this work develops the Laplace loss which constraints the predicted temperature field with the steady-state conduction equation with heat sources.
Then, the boundary conditions are constrainted by the formulated BC loss. Besides, the point loss is  developed to ensure the consistency of the temperature value of monitoring points in the  constructed temperature field. More importantly, the total variation loss is used to guarantee the  one order spatial smoothness of the reconstructed temperature field.  These four losses assist the deep surrogate model to learn the physical  correlation of the system. Hence, joint learning with these four losses can make the surrogate model provide impressive reconstruction performance for the TFR-HSS task.

{Overall, 
this work develops a novel reconstruction method which joints the deep reversible regression model and physics-informed unsupervised learning for TFR-HSS task. The deep reversible regression model takes advantage of the powerful representational ability of deep model and the specific reversible regression module to extract the thermal property of the system. Besides, this work develops a physics-informed reconstruction loss, which can learn the deep model unsupervisedly. It can alleviate the dependence of massive labelled samples for the training of deep model and provide the reconstruction model without labelled samples.}

To sum up, this work makes the following contributions.
\begin{itemize}
\item This work defines the temperature field reconstruction of heat-source systems (TFR-HSS) task from real-world engineering applications and provides mathematical formulation as well as the  numerical modeling of the task.
\item  {This work proposes a novel reversible regression model which joints the reversible module and deep regression model to better represent the thermal property of the system for TFR-HSS task.}
\item {This work develops the physics-informed reconstruction loss (PIRL) for TFR-HSS task,  which can train the deep model without labelled samples}.
\end{itemize}
Moreover, the experiments over typical two-dimensional heat-source systems have been conducted and the  comparison results with the most recent methods have demonstrated the superiority of the proposed method for TFR-HSS task.

The remainder of this paper is arranged as follows. In Section \ref{sec:definition}, the mathematical form of the TFR-HSS task is defined. Section \ref{sec:method} numerically models the TFR-HSS task, and {further develops the novel reconstruction method which joints the proposed deep reversible regression model and the developed physics-informed reconstruction loss.} The experimental studies over typical two-dimensional heat-source systems are presented to validate the effectiveness of the proposed method in Section \ref{sec:experiments}. Finally, we conclude this paper and point out future directions in Section \ref{sec:conclusions}.

\section{Temperature Field Reconstruction of Heat-Source Systems (TFR-HSS)}
\label{sec:definition}
This work focuses on the heat-source systems where heat may be generated internally and the principles of heat transfer are mainly concerned with.
The heat-source systems can be  modelled as a two-dimensional domain where each electronic component is defined as the heat source with different shapes and  thermal conduction occurs over this specific domain \cite{square1, ij_3}. Given the system, the TFR-HSS task aims to reconstruct the overall temperature field with specified temperature values from monitoring tensors and is an important part of real-time health detection system of electronic equipment in engineering.

Given a heat-source system with $\Lambda$ heat sources and $m$ transducers are placed on the layout domain for temperature monitoring.
Generally, the mathematical formulation of the TFR-HSS task can be formulated as
\begin{equation}\label{eq:optimization}
T^* = \arg\min\limits_{T}(\sum\limits_{i\in[1,m]}|T(x_{s_i},y_{s_i}|\phi_1,\cdots,\phi_\Lambda)-f(x_{s_i},y_{s_i})|) ,
\end{equation}
where $T(\cdot)$ describes the reconstructed temperature field of the system, and $f(\cdot)$ is the monitoring temperature value. $\phi_i(i=1,2,\cdots,\Lambda)$ represents the intensity distribution  of the $i$-th heat source.
For simplicity, the intensity of each heat source is uniformly distributed and set to a constant value.
$O_1,O_2,\cdots,O_m$ denote the monitoring points where $(x_{s_i},y_{s_i})$ describes the positions of the $O_i$ monitoring point and $m$ is the number of the points.

Therefore, the key process of this specific task is to obtain the optimal temperature field using Eq. \ref{eq:optimization} while satisfying the physical equation of thermal conduction. As for the two-dimensional heat conduction, the steady-state satisfies the Laplace equation, which can be formulated as
\begin{equation}\label{eq:laplace0}
\frac{\partial}{\partial x}(\lambda\frac{\partial T}{\partial x})+\frac{\partial}{\partial y}(\lambda\frac{\partial T}{\partial y})+\sum\limits_{i=1}^\Lambda \phi_i(x,y)=0,
\end{equation}
where $\lambda$ represents the thermal conductivity of the domain. 
In addition to the thermal condution equation, physical properties over boundary conditions should also be constrained and can be generally written as
\begin{equation}
 T=T_0 \  \text{or} \   {\lambda\frac{\partial T}{\partial {\bf n}}=0} \  \text{or} \
{\lambda \frac{\partial T}{\partial {\bf n}}=h(T-T_0)},
\end{equation}
where $T_0$ is a constant temperature value, ${\bf n}$ denotes the (typically exterior) normal to the boundary, and $h$ represents the convective heat transfer coefficient. The three boundaries are known as the Dirichlet boundary conditions (Dirichlet BCs) where $T_0$ is the isothermal boundary temperature, the Neumann boundary conditions (Neumann BCs) where zero heat flux is exchanged, and the Robin boundary conditions (Robin BCs)  which combines the Dirichlet BCs and Neumann BCs.
Overall, the TFR-HSS task can be transformed as the following optimization problem:
\begin{equation}\label{eq:problem}
\begin{aligned}
\min\limits_{T}&(\sum\limits_{i\in[1,m]}|T(x_{s_i},y_{s_i}|\phi_1,\cdots,\phi_\Lambda)-f(x_{s_i},y_{s_i})|) \\
s.t. &\ \frac{\partial}{\partial x}(\lambda\frac{\partial T}{\partial x})+\frac{\partial}{\partial y}(\lambda\frac{\partial T}{\partial y})+\sum\limits_{i=1}^\Lambda \phi_i(x,y)=0 \\
&T=T_0 \  \text{or} \   \lambda{\frac{\partial T}{\partial {\bf n}}=0} \  \text{or} \
{ \lambda\frac{\partial T}{\partial {\bf n}}=h(T-T_0)}.
\end{aligned}
\end{equation}

\begin{figure}[t]
\centering
\includegraphics[width=0.8\linewidth]{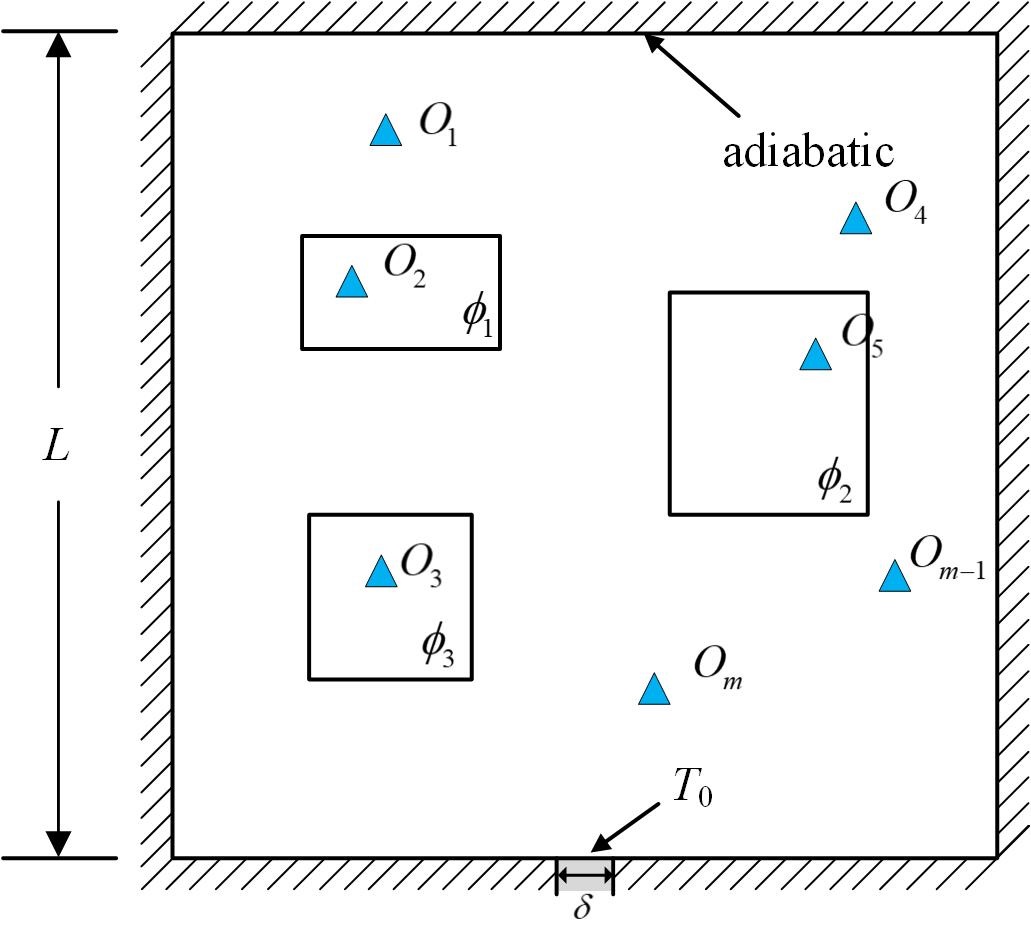}
	\caption{The illustration of the region $\Omega$ of the heat-source systems with $m$ sensors for temperature monitor.}
	\label{fig:definition}
\end{figure}

As a representative but without loss of generality, the volume-to-point(VP) heat conduction problem \footnote{{Another representive problem, e.g. volume-to-boundary (VB), can be seen as a specific ones of VP problems.}} in a two-dimensional rectangular domain as Fig. \ref{fig:definition} shows is taken for the validation of the proposed method (just as \cite{chen2021} and \cite{square1}). In this problem, all the boundaries are adiabatic except  the  small patch of heat sink which is represented as $\delta$ in the figure. The objective of the reconstruction problem is to reconstruct the temperature field of the domain from the $m$ monitoring points.

Through mathematical modeling, the TFR-HSS task can be transformed to the continuous optimistic problem as Eq. \ref{eq:problem} shows.
To solve this optimization, this work numerically models the problem, {and develops a novel reconstruction method for the task.}
Following we will introduce the numerical  modeling process, {the construction of the deep reversible regression model and  the physics-informed reconstruction loss for temperature field reconstruction in detail.}

\section{Proposed Method}
\label{sec:method}

 For convenience, $\Omega$, $\Omega_e$, $\Omega_l$, and $\Omega_b$ are used to describe the square domain, the layout area without heat sources laid on, the area with heat sources laid on, and the boundary area, respectively.
The size of the domain and the length of the heat sink is described by $L$ and $\delta$ separately.


\subsection{Numerical  Modeling for TFR-HSS Task}\label{subsec:obser}

\begin{figure}[t]
\centering
 \subfigure[]{\label{fig:layout}\includegraphics[width=0.31\linewidth]{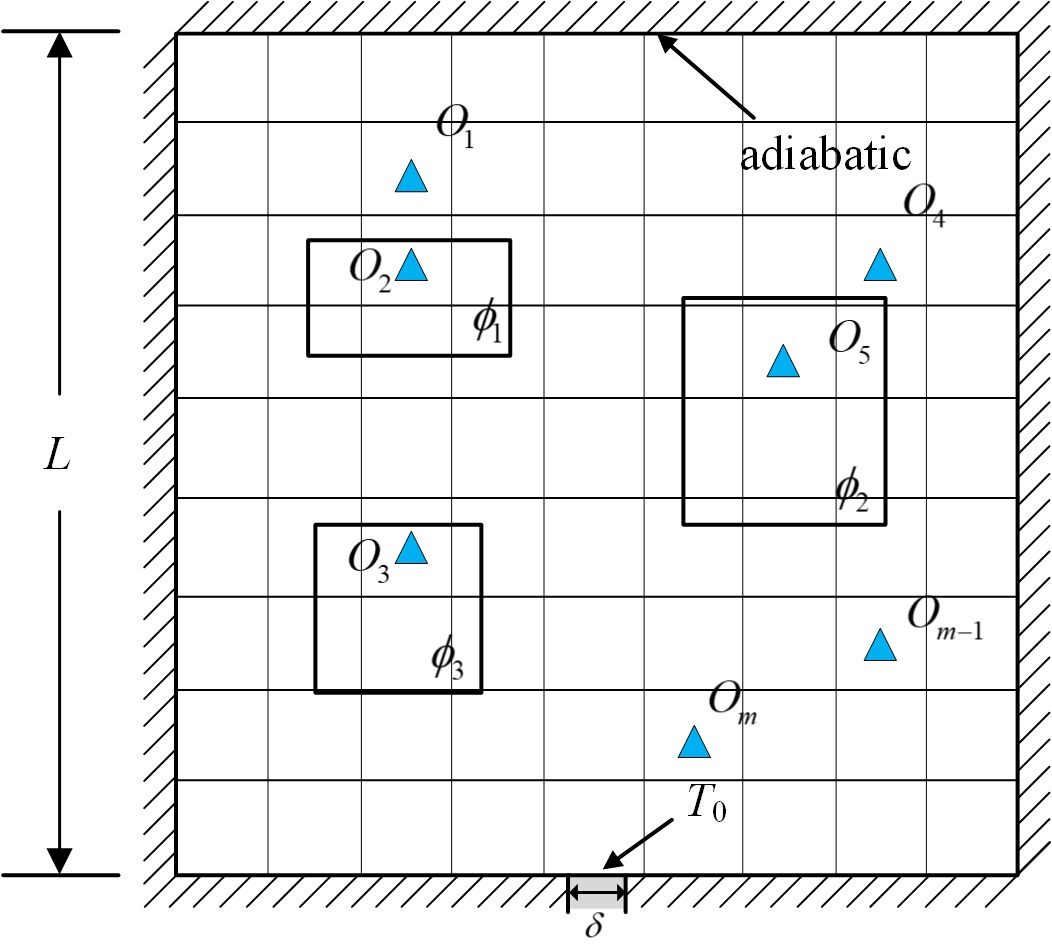}}
 \subfigure[]{\label{fig:monitoring}\includegraphics[width=0.31\linewidth]{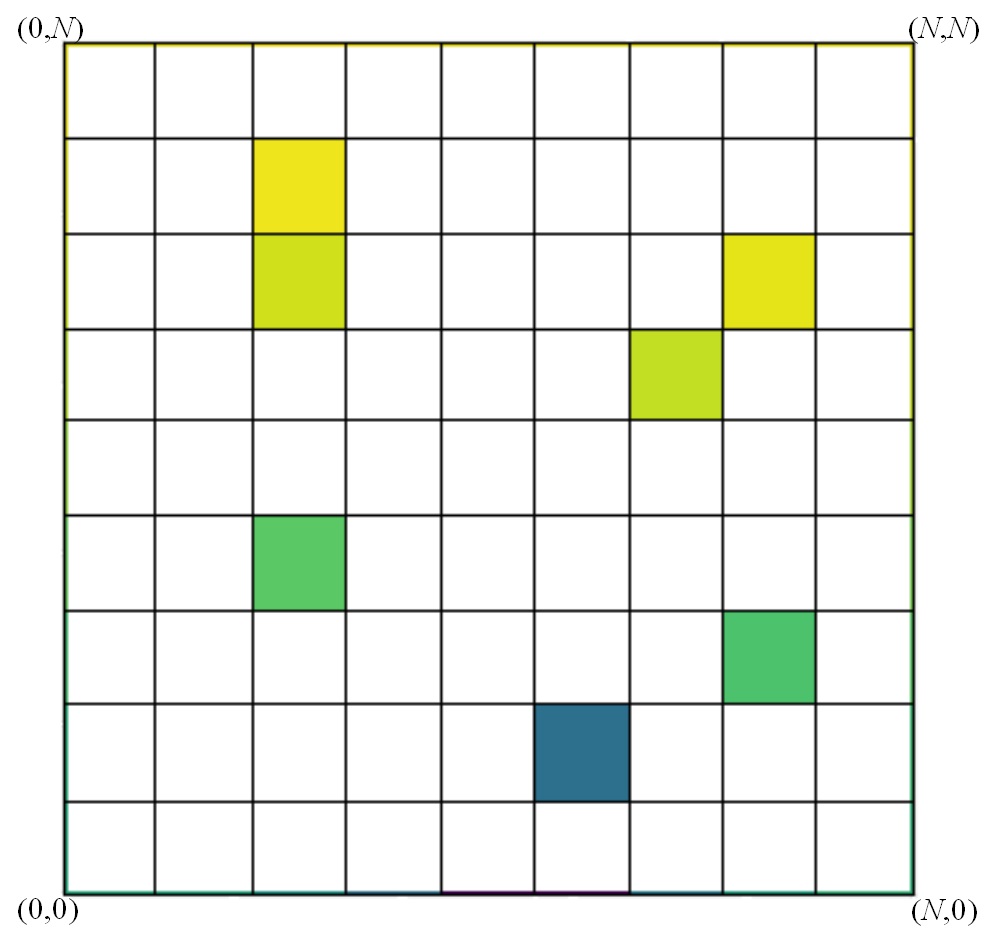}}
 \subfigure[]{\label{fig:reconstructed_field}\includegraphics[width=0.35\linewidth]{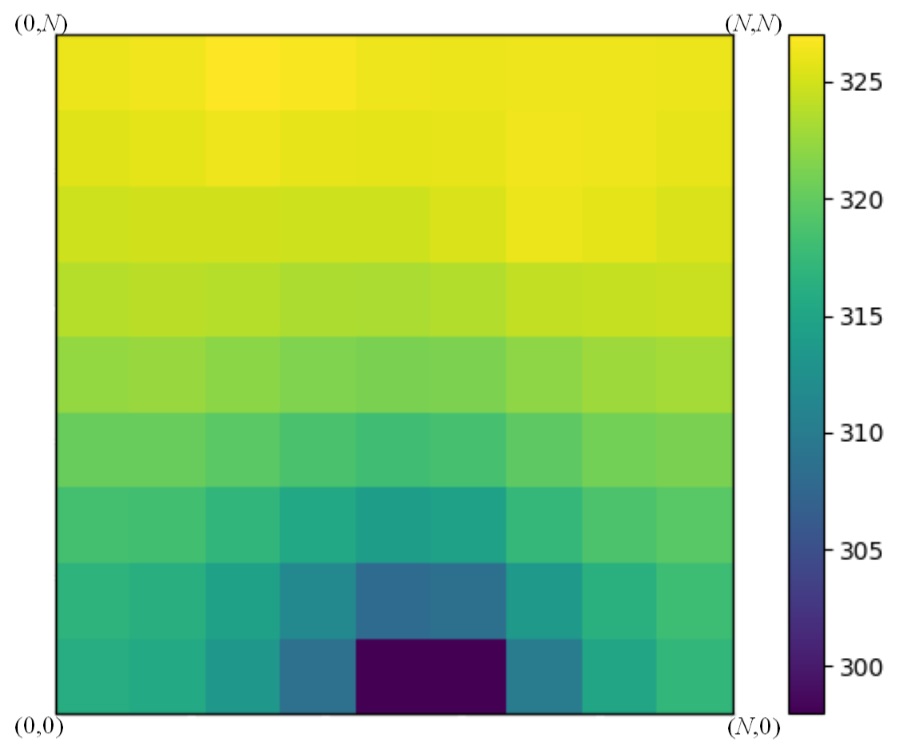}}
   \caption{Numerical  modeling for TFR-HSS task. (a) {Layout board with $N\times N$ grid}; (b) {$N\times N$ discreted monitoring matrix}; (c) {$N\times N$ discreted temperature field}.}
\label{fig:mesh}
\end{figure}

In order to facilitate the computing process, {numerical  modeling of the TFR-HSS task \cite{tfrd_gong}} is necessary.
At first, just as Fig. \ref{fig:layout} shows,
the layout domain is meshed by $N\times N$ grid. The area within a certain grid is supposed to share a constant temperature value.
The monitoring points are arranged in the grids to obtain the temperature of  the grids.
Then,  two-dimensional $N\times N$ matrix $f$ can be obtained to describe monitoring information and used as the input to reconstruct the overall temperature field in the layout domain. As Fig. \ref{fig:monitoring} shows, the discreted monitoring matrix $f$ is filled with the monitoring temperature value at the monitoring point and the remainder points will be filled with the constant value $T_0$. 

Through numerical  modeling, the objective of TFR-HSS task is to obtain the temperature value of  other grids in the matrix.
Then, as Fig. \ref{fig:reconstructed_field} shows, the output of the task is  the reconstructed temperature field which has been numerically modelled as a $N\times N$ matrix $T$. 
Therefore, the task can be seen as a discrete optimization problem which tries to find the surrogate mapping $\Phi$ from monitoring matrix $f$  to temperature field $T$, and it can be written as
\begin{equation}
f{\xrightarrow{\Phi}}T.
\end{equation}
Since $f$ and $T$ are both $N\times N$ matrix, the problem can be seen as the  regression problem. General image-to-image regression methods can be applied as the {deep surrogate model for the TFR-HSS task \cite{chen2021}.}

However, due to special characteristics of the  TFR-HSS task,  vanilla deep regression methods usually  cannot  work well for the task. Through fully  consideration of these characteristics, this work develops a novel reversible regression model for TFR-HSS task. Besides,  due to the high cost to obtain labelled training samples, this work develops the physics-informed training loss based on the physical properties of TFR-HSS task  which can learn the deep model unsupervisedly. Following we will introduce the proposed method in detail.

\subsection{Reversible Regression Model}

Due to the calculation order of convolution operation in  vanilla deep regression models, the reconstructed temperature field usually has jagged boundary temperature, especially  the upside and rightside of the reconstructed temperature field.
The reason is that  the convolutional operation over the input is conducted orderly, and the final calculated area cannot capture enough physical information to update the temperature under only one encoder-decoder process. 

\begin{figure}[t]
\centering
\includegraphics[width=0.8\linewidth]{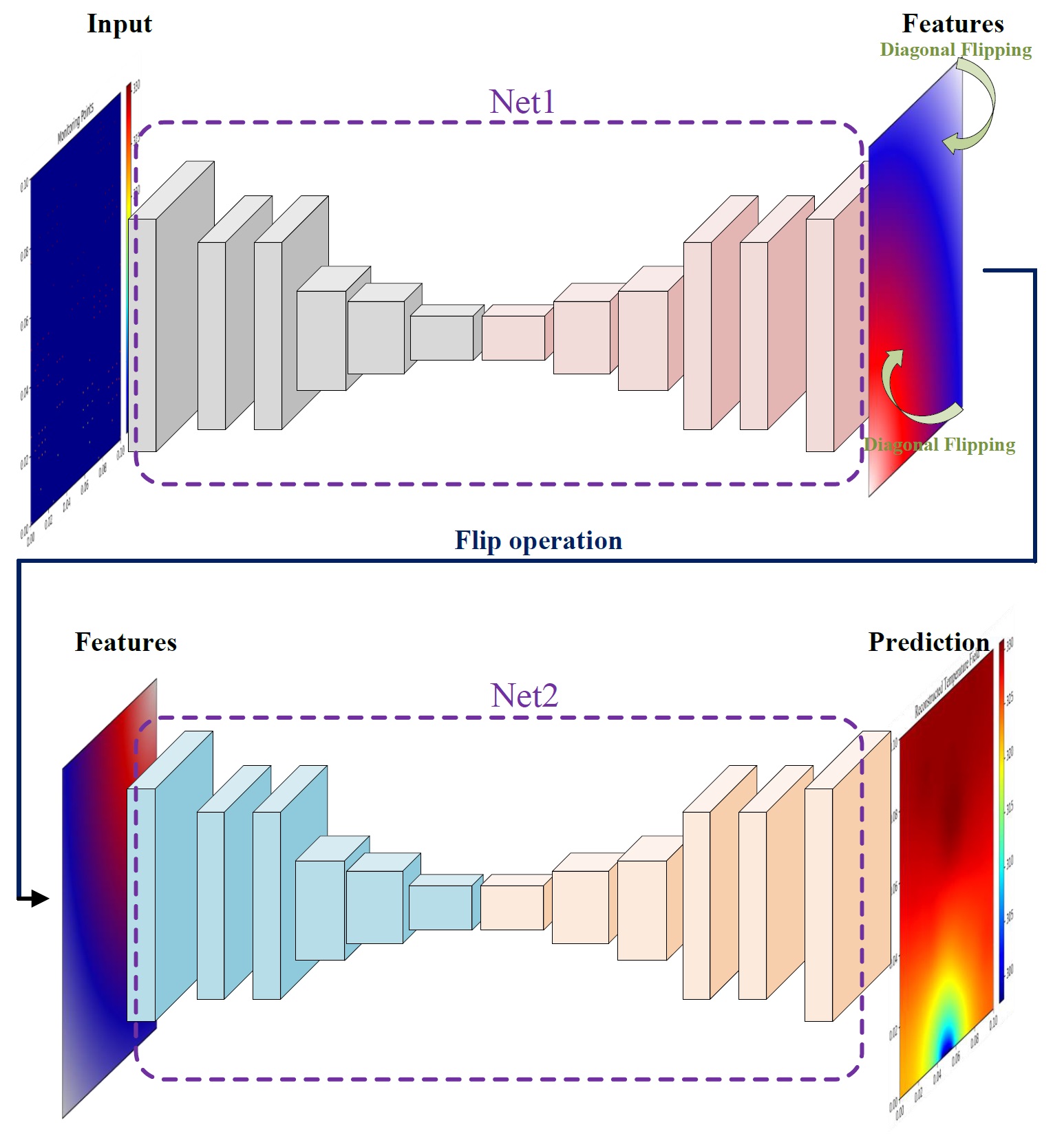}
\caption{The architecture of proposed reversible regression model for the TFR-HSS task. Here, $Net_1$ and $Net_2$ represent the base models in the proposed model which can have the same or different structures.}
	\label{fig:net}
\end{figure}

Taking these characteristics of TFR-HSS task into consideration, this work develops the reversible regression model which consists of two encoder-decoder operations and one flip operation to guarantee the reconstruction performance of the surrogate model. 
Fig. \ref{fig:net} presents the architecture of the proposed model.
As the figure shows, the proposed model, {formulated} as $Net_1-Net_2$, is divided into two parts, e.g. $Net_1$ and $Net_2$. Both $Net_1$ and $Net_2$ are independent  encoder-decoder models. Between $Net_1$ and $Net_2$, the flip operation with diagonal flipping is conducted to make the computing process reversible. Such operation  could make the reversible regression model better fit for the TFR-HSS task. {This is because through diagonal flipping operation, the upside and rightside of the field are transformed to the downside and leftside one. Therefore, the physical information all over the field can be captured and  the whole temperature field can be well reconstructed.}

The proposed reversible regression model is easy to implement and can provide remarkable reconstruction performance for TFR-HSS task. Benefit from the flip operation and two successive encoder-decoder processes, the temperature field can be reconstructed well with limited monitoring points,  including that over the boundaries. 

\subsection{Physics-Informed Reconstruction Loss}\label{subsec:loss}

{For complex physical systems, labelled samples are difficult to obtain or even unavailable, and therefore, unsupervised learning with physical information without labelled samples becomes a feasible and promising way for modeling the physical systems \cite{nr16, nr17}.
For the task at hand, considering the characteristics of the TFR-HSS task,} this work develops the novel physics-informed reconstruction loss for the task {and trains the reversible regression model unsupervisedly.} The loss consists of four significantly different loss terms which describe different physical characteristics from different perspectives.

 {\bf Point loss:} This work formulates the point loss $L_{point}$ {based on MSE loss \cite{nr18}} to make the predicted temperature  satisfying the temperature over monitoring points. Then, $L_{point}$ can be formulated as
\begin{equation}\label{eq:point}
L_{point}=\sum_{i=1}^m|T(x_{s_i},y_{s_i})-f(x_{s_i},y_{s_i})|^2_2,
\end{equation}
where $|\cdot|_2^2$ represents the $L_2$ norm.

 {\bf BC loss:} In addition to point loss  for utilizing the information from monitoring points, the boundary conditions in Eq. \ref{eq:problem} also have significant effects on the temperature field. To use the boundary conditions for reconstruction, this work designs the
BC loss $L_{bc}$ to ensure the physical properties of the boundaries of the temperature field.
Since this work mainly considers the Romann B.C. and Dirichlet B.C., the $L_{bc}$ is formulated over this two kinds of boundary conditions separately.
As introduces in Section \ref{sec:definition},
the Romann B.C. satisfies
$|\frac{\partial T}{\partial {\bf n}}(x,y)|_{(x,y)\in \Omega_b^{romann}}=0$, 
and the Dirichlet B.C. satisfies
$|T(x,y)-T_0|_{(x,y)\in \Omega_b^{dirichlet}}=0$.
Therefore, the continuous form of the $L_{bc}$ loss can be formulated as
\begin{equation}
L_{bc}=|T(x,y)-T_0|_{(x,y)\in \Omega_b^{dirichlet}}+|\frac{\partial T}{\partial {\bf n}}(x,y)|_{(x,y)\in \Omega_b^{romann}},
\end{equation}
where $\Omega_b^{dirichlet}$, $\Omega_b^{romann}$ describes the boundary regions satisfying the Dirichlet B.C. and the Romann B.C., respectively.
Here, the BC loss is also processed discretely.
 Then, the BC loss over the Dirichlet B.C. can be written as
\begin{equation}
L_{bc}^{Dirichlet}=\sum\limits_{(x_i,y_j)\in \Omega_b^{dirichlet}}|T(x_i,y_j)-T_0|.
\end{equation}
For Romann B.C., the temperature over the boundary should satisfy 
$T(x_i,y_N)-T(x_i,y_{N-1})=0$
or $T(x_i,y_2)-T(x_i,y_{1})=0$
or $T(x_N,y_j)-T(x_{N-1},y_{j})=0$
or $T(x_2,y_j)-T(x_1,y_{j})=0 (i,j = 1,2,\cdots,N)$.
In this work, the BC loss over Romann B.C. is implemented through replicate padding of the temperature field with one dimension over the surrounded boundaries. Since the physical property with Romann B.C. can be satisfied through replicate padding operation, the $L_{bc}$ loss could mainly concern about the physical property over the Dirichlet B.C., and it can be written as
\begin{equation}\label{eq:bc}
L_{bc}=\sum\limits_{(x_i,y_j)\in \Omega_b^{dirichlet}}|T(x_i,y_j)-T_0|.
\end{equation}

 {\bf Laplace loss:} For the heat-source systems, the thermal conduction satisfies the two-dimensional laplace equation as Eq. \ref{eq:laplace0} shows. It builds the relationship between different points inside the heat-source systems and can guide the reconstruction of the temperature field with the help of the monitoring points.
In order to  utilize such kind of physical information,  the laplace loss $L_{laplace}$ is built to preserve the physical characteristics of the predicted temperature on the domain.
Based on Eq. \ref{eq:laplace0},  the laplace loss can be formulated as
\begin{equation}\label{eq:losslaplace_0}
L_{laplace}=\sum_{(x,y)\in \Omega}|\frac{\partial}{\partial x}(\lambda\frac{\partial T}{\partial x})+\frac{\partial}{\partial y}(\lambda\frac{\partial T}{\partial y})+\sum\limits_{i=1}^\Lambda \phi_i(x,y)|,
\end{equation}
where $\lambda$ is set to 1 in this work, namely constant thermal conductivity is assumed. Then Eq. \ref{eq:losslaplace_0} can be reformulated as
\begin{equation}
L_{laplace}=\sum_{(x,y)\in \Omega}|\frac{\partial}{\partial x}(\frac{\partial T}{\partial x})+\frac{\partial}{\partial y}(\frac{\partial T}{\partial y})+\sum\limits_{i=1}^\Lambda \phi_i(x,y)|.
\end{equation}
For the current task, the real-time intensity of heat sources in the system remains unknown, and therefore in this work, the laplace loss is constructed by the thermal conduction characteristics over the domain without the heat sources laid where the intensity can be seen as zero for simplicity. Then, the continuous form of the final laplace loss $L_{laplace}$ can be written as
\begin{equation}
L_{laplace}=\sum_{(x,y)\in \Omega_e}|\frac{\partial}{\partial x}(\frac{\partial T}{\partial x})+\frac{\partial}{\partial y}(\frac{\partial T}{\partial y})|,
\end{equation}
where $\Omega_e$ describes the domain without heat source placed.
Without exception, the laplace loss is implemented discretely. Based on the difference equation, 
$\displaystyle{\frac{\partial}{\partial x}(\frac{\partial T}{\partial x})}$ at $(x_i,y_j)$ can be calculated as
\begin{equation}
\begin{aligned}
\frac{\partial}{\partial x}(\frac{\partial T}{\partial x})|_{x=x_i}=&\frac{\partial}{\partial x}(\frac{T(x_{i+1},y_j)-T(x_{i},y_j)}{x_{i+1}-x_{i}}) \\
=&\frac{\frac{T(x_{i+1},y_j)-T(x_{i},y_j)}{x_{i+1}-x_{i}}-\frac{T(x_{i},y_j)-T(x_{i-1},y_j)}{x_{i}-x_{i-1}}}{x_{i}-x_{i-1}}.
\end{aligned}
\end{equation}
Since the temperature field is calculated by the uniform square mesh,
the equation can be reformulated as
\begin{equation}
\frac{\partial}{\partial x}(\frac{\partial T}{\partial x})|_{x=x_i}=\frac{T(x_{i+1},y_j)+T(x_{i-1},y_j)-2T(x_{i},y_j)}{{\Delta x}^2},
\end{equation}
where $\Delta x=x_{i+1}-x_i$.
Similarly, $\displaystyle{\frac{\partial}{\partial y}(\frac{\partial T}{\partial y})|_{y=y_j}}$ can be reformulated as
\begin{equation}
\frac{\partial}{\partial y}(\frac{\partial T}{\partial y})|_{y=y_j}=\frac{T(x_{i},y_{j+1})+T(x_{i},y_{j-1})-2T(x_{i},y_j)}{{\Delta y}^2},
\end{equation}
where $\Delta y=y_{j+1}-y_j$.
Therefore, the discrete form of $L_{laplace}$ can be written as
\begin{equation}\label{eq:laplace}
\begin{aligned}
L_{laplace}=&\sum_{(x_i,y_j)\in \Omega_e}|\frac{\partial}{\partial x}(\frac{\partial T}{\partial x})|_{x=x_i}+\frac{\partial}{\partial y}(\frac{\partial T}{\partial y})|_{y=y_j}| \\
=&\sum_{(x_i,y_j)\in \Omega_e}|\frac{T(x_{i+1},y_j)+T(x_{i-1},y_j)-2T(x_{i},y_j)}{{\Delta x}^2}+ \\
&\ \ \ \ \ \ \ \ \ \ \ \ \frac{T(x_{i},y_{j+1})+T(x_{i},y_{j-1})-2T(x_{i},y_j)}{{\Delta y}^2}|.
\end{aligned}
\end{equation}
In the numerical modeling process, we set $\Delta=\Delta x=\Delta y$.
Denote $D_{x_i,y_j}=T(x_{i+1},y_j)+T(x_{i-1},y_j)+T(x_{i},y_{j+1})+T(x_{i},y_{j-1})-4T(x_{i},y_j)$.
Then, $L_{laplace}$ can be reformulated as
\begin{equation}
L_{laplace}=\sum_{(x_i,y_j)\in \Omega_e}|\frac{D_{x_i,y_j}}{{\Delta}^2}|.
\end{equation}
 Interestingly, $D_{x_i,y_j}$ is the typical two-dimensional difference format and can be seen as a special form of convolutional operation. Therefore, in this work, $L_{laplace}$ can be implemented using convolutional operation and $L_1$ norm.

{\bf TV loss:} Generally speaking, the temperature field changes gentlely and there is no drastic changes in steady-state temperature field. Considering such property, this work further uses the {total variation (TV) regularization \cite{tv}} for the temperature field reconstruction task. The
TV regularization $L_{tv}$ encourages the spatial smoothness  of the reconstructed temperature field, and  further helps the training process with 1-D physical property.
It mainly considers the one-order gradient information of the temperature field and can be formulated as 
\begin{equation}\label{eq:regularization}
L_{tv}=\int_{\Omega}(\frac{\partial T}{\partial x}(x,y)^2+\frac{\partial T}{\partial y}(x,y)^2)^\frac{\rho}{2},
\end{equation}
where $\rho$ describes the order of the TV regularization.
The gradient information describes the local changes of the temperature field and can be re-written as the relationship of neighboring points of the field discretely. Therefore, Eq. \ref{eq:regularization} can be calculated by  
\begin{equation}\label{eq:tv}
\begin{aligned}
L_{tv}=\sum_{i=1}^N\sum_{j=1}^N ((T(x_i,&y_{j+1})-T(x_i,y_j))^2+ \\
&(T(x_{i+1},y_j)-T(x_i,y_j))^2)^\frac{\rho}{2}.
\end{aligned}
\end{equation}
In this work, $\rho$ is set to 2 and the used $L_{tv}$ can be written as
\begin{equation}\label{eq:tv}
\begin{aligned}
L_{tv}=\sum_{i=1}^N\sum_{j=1}^N ((T(x_i,&y_{j+1})-T(x_i,y_j))^2+ \\
&(T(x_{i+1},y_j)-T(x_i,y_j))^2).
\end{aligned}
\end{equation}

{\bf Overall loss:} Based on Eqs. \ref{eq:point}, \ref{eq:bc}, \ref{eq:laplace}, and \ref{eq:tv}, the final physics-informed reconstruction loss (PIRL) can be formulated as
\begin{equation}\label{eq:loss}
L=L_{point}+\alpha L_{bc}+\beta L_{laplace}+\gamma L_{tv},
\end{equation}
where $\alpha, \beta, \gamma$ stands for the tradeoff parameters.

\subsection{Evaluation Metrics for the Reconstruction Performance}

{To thoroughly evaluate the reconstruction performance of the proposed method,} this work designs the following four metrics based on the {temperature field information we concern about \cite{tfrd_gong}}, namely the mean absolute error (MAE), the maximum of component-constrained absolute error (M-CAE), the component-constrained mean absolute error (CMAE), and the boundary-constrained mean absolute error (BMAE).

{\bf Mean absolute error (MAE)} calculates the mean absolute error of the whole reconstructed temperature field, which can be formulated as
 \begin{equation}\label{eq:mae}
E_{MAE}=\frac{1}{N^2}\sum\limits_{i=1}^N\sum\limits_{j=1}^N|T(x_i,y_j)-\hat{T}(x_i,y_j)|,
\end{equation}
where $\hat{T}$ represents the real temperature field obtained by numerical simulation.

{\bf Component-constrained mean absolute error (CMAE)} calculates the mean absolute error of area with heat sources laid on and it can be written as
\begin{equation}\label{eq:cmae}
E_{CMAE}=\frac{1}{|\Omega_l|}\sum\limits_{(x_i,y_j)\in\Omega_l}|T(x_i,y_j)-\hat{T}(x_i,y_j)|,
\end{equation}
where $\Omega_l$ represents the area with heat sources laid on.

{\bf Maximum of Component-constrained absolute error (M-CAE)} calculates the maximum absolute error of the whole reconstructed temperature field, which can be formulated as
\begin{equation}\label{eq:mcae}
E_{M-CAE}=\max\limits_{(x_i,y_j)\in \Omega_l}|T(x_i,y_j)-\hat{T}(x_i,y_j)|.
\end{equation}

{\bf Boundary-constrained mean absolute error (BMAE)} calculates the mean absolute error over the boundary area and it can be written as
\begin{equation}\label{eq:bmae}
E_{BMAE}=\frac{1}{|\Omega_b|}\sum\limits_{(x_i,y_j)\in\Omega_b}|T(x_i,y_j)-\hat{T}(x_i,y_j)|,
\end{equation}
where $\Omega_b$ denotes the boundary regions.

Overall, the MAE measures the overall reconstruction performance of the proposed method which describes the difference between the differential equation modeling and the reconstructed field with the proposed method. M-CAE, CMAE, and BMAE indicate local errors which are used as supplements of MAE where the M-CAE, CMAE describes the reconstruction performance of local area  of heat sources, and the BMAE concerns the reconstruction performance over the boundaries.  If not specified, the reconstruction performance in the following describes the performance evaluated over such four evaluation metrics.

\subsection{Implementation of the Proposed Method for TFR-HSS task}

{The overall flowchart of the proposed method is showed in Fig. \ref{fig:nflowchart}. The pseudocode of the training process of the proposed method is given in Algorithm \ref{algorithm:1}. The implementation can be divided into three parts: training process, prediction process, and evaluation process.

\begin{figure}[t]
\centering
\includegraphics[width=0.99\linewidth]{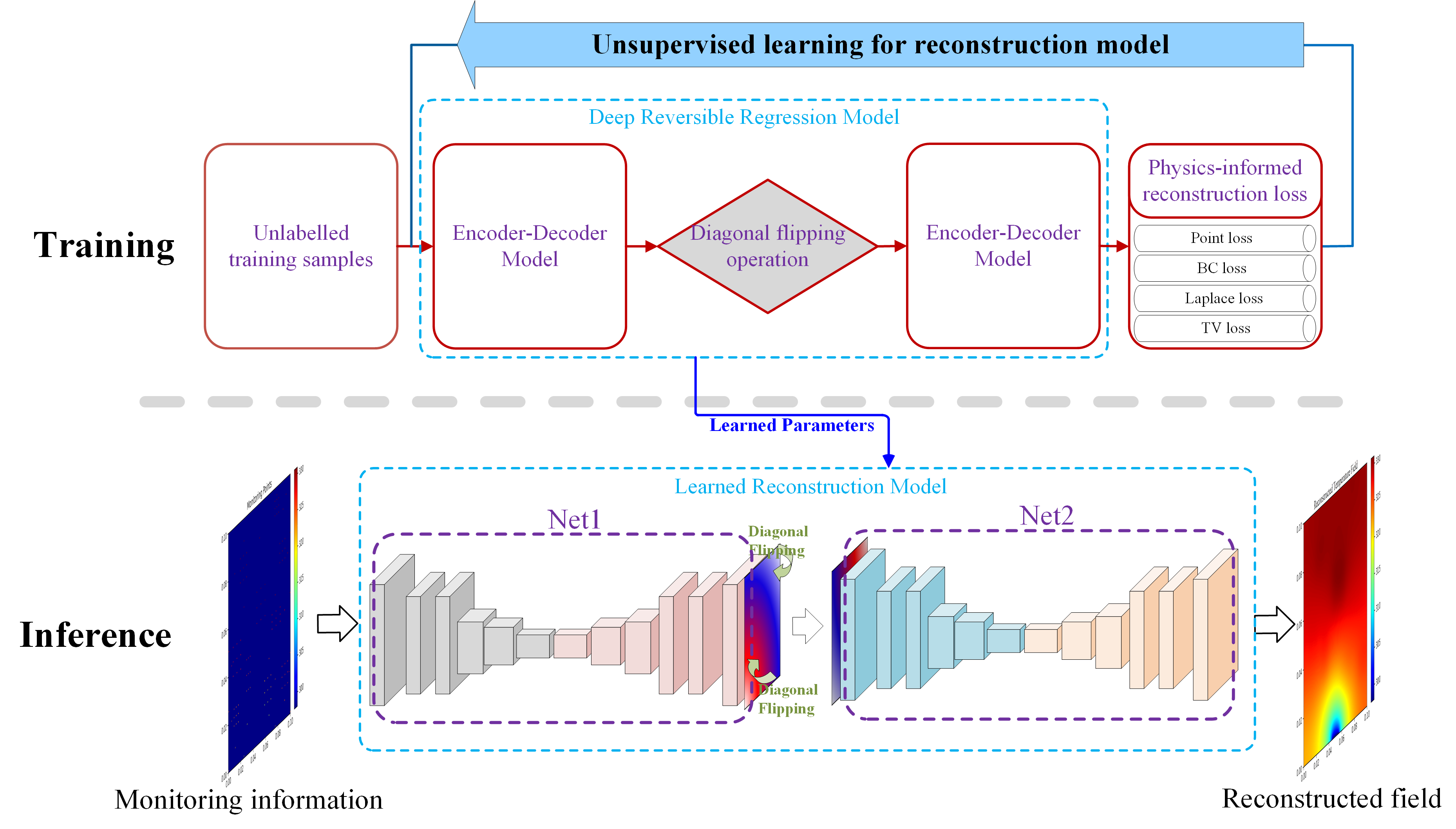}
\caption{The flowchart of the proposed method for temperature field reconstruction of heat source systems.}
	\label{fig:nflowchart}
\end{figure}
}

{\bf Step 1:} Prepare the training samples, construct the reversible regression model as surrogate mapping to extract the physical information and set the hyper-parameters.

{\bf Step 2:} Train the reversible regression model unsupervisedly following step 2-11 in Algorithm \ref{algorithm:02} and provide the optimized surrogate mapping $\Phi^*$.

{\bf Step 3:} Predict the temperature field with obtained surrogate mapping $\Phi^*$.

{\bf Step 4:} Evaluate the surrogate mapping $\Phi^*$ under the given metrics.

\begin{algorithm}[t]
\renewcommand{\algorithmicrequire}{\textbf{Input:}}
\renewcommand{\algorithmicensure}{\textbf{Output:}}
\caption{The framework of the proposed method for TFR-HSS task}\label{algorithm:02}
\begin{algorithmic}[1]
\REQUIRE Training samples $\{f_1, f_2, \cdots, f_n\}$, Testing samples $\{f_{t_1}, f_{t_2}, \cdots, f_{t_k}\}$, surrogate mapping $\Phi$, hyperparameter $\alpha, \beta, \gamma$.
\ENSURE $\Phi^*$
\STATE // Training process: Train the model
\WHILE{not converge}
\STATE Reconstruct the temperature field with surrogate mapping by $T_i=\Phi(f_i) (i=1,2,\cdots,n)$.
\STATE Compute the point loss $L_{point}$ using Eq. \ref{eq:point}.
\STATE Compute the laplace loss $L_{laplace}$ using Eq. \ref{eq:laplace}.
\STATE Compute the TV loss $L_{tv}$ using Eq. \ref{eq:tv}.
\STATE Compute the bc loss $L_{bc}$ using Eq. \ref{eq:bc}.
\STATE Compute the training loss $L$ using Eq. \ref{eq:loss}.
\STATE Update $\Phi$ using training loss $L$ by {\bf auto-grad}.
\ENDWHILE
\STATE Provide the optimized surrogate mapping $\Phi^*$.
\STATE // Prediction process: Predict temperature field with trained model
\STATE Predict the temperature field of testing samples using $T_{t_i}=\Phi^*(f_{t_i}) (i=1,2,\cdots,k)$.
\STATE // Evaluation process: Evaluate the performance of trained model
\STATE Evaluate surrogate mapping $\Phi^*$ under MAE, CMAE, M-CAE, and BMAE using Eq. \ref{eq:mae}-\ref{eq:bmae}.
\STATE {\bf return} $\Phi^*$.
\end{algorithmic}
\label{algorithm:1}
\end{algorithm}

\section{Experimental Studies}
\label{sec:experiments}

\subsection{Experimental Setups}

{\bf Datasets: }
To validate the effectiveness of the proposed method for temperature field reconstruction, this work constructs four typical thermal analysis datasets with different heat sources and boundary conditions, which are described as Data A, Data B, Data C, Data D, respectively.

The domain of the systems is denoted as $0.1m \times 0.1m$ board. 
For convenience but without loss of generality,  the heat sinks with 0.01m are used as the boundary conditions for heat dissipation. Specifically, for Data A, Data B, Data C, Data D, the heat sinks exist over the center of the upside, downside, leftside, rightside of the boundary, respectively.
Due to the page limitation, the layout information and characteristics of heat source components of different datasets are {showed} in the attachment.

\begin{figure}[t]
\centering
 \subfigure[Data A]{\label{fig:1a}\includegraphics[width=0.9\linewidth]{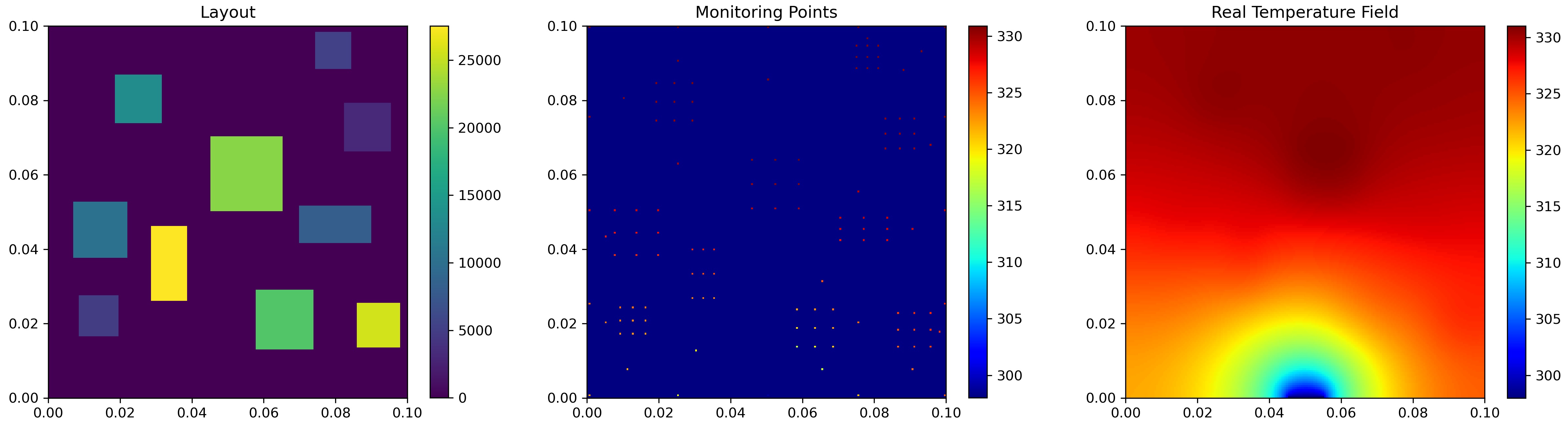}}
 \subfigure[Data B]{\label{fig:1b}\includegraphics[width=0.9\linewidth]{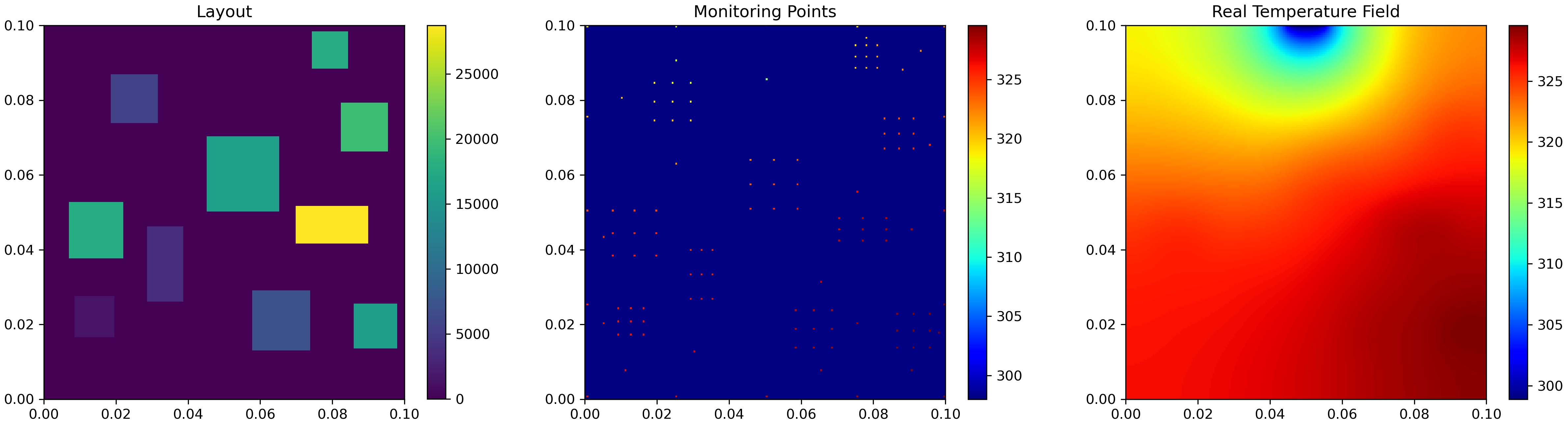}}
 \subfigure[Data C]{\label{fig:1c}\includegraphics[width=0.9\linewidth]{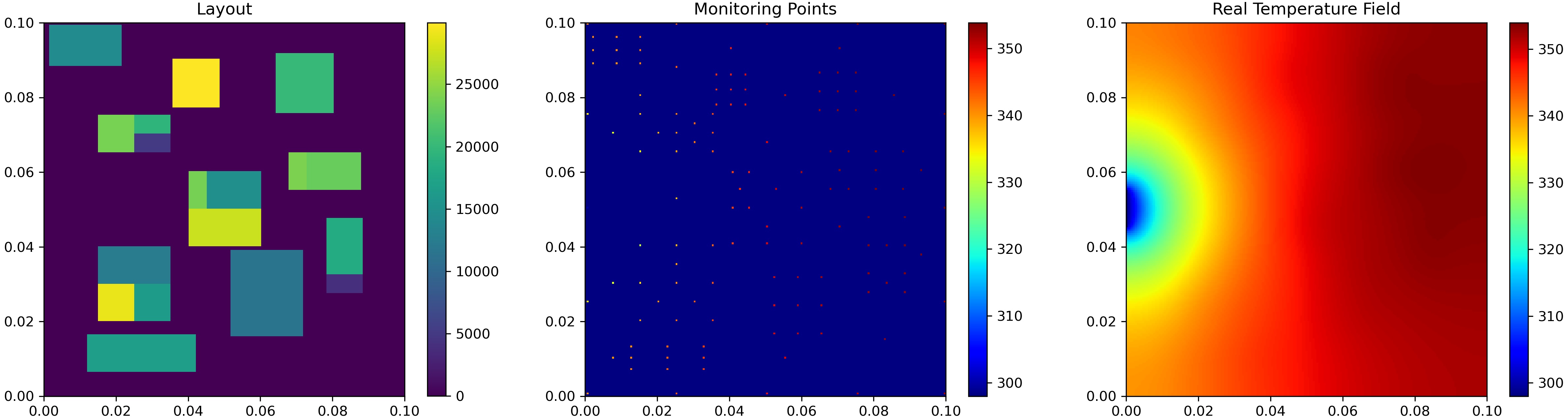}}
 \subfigure[Data D]{\label{fig:1d}\includegraphics[width=0.9\linewidth]{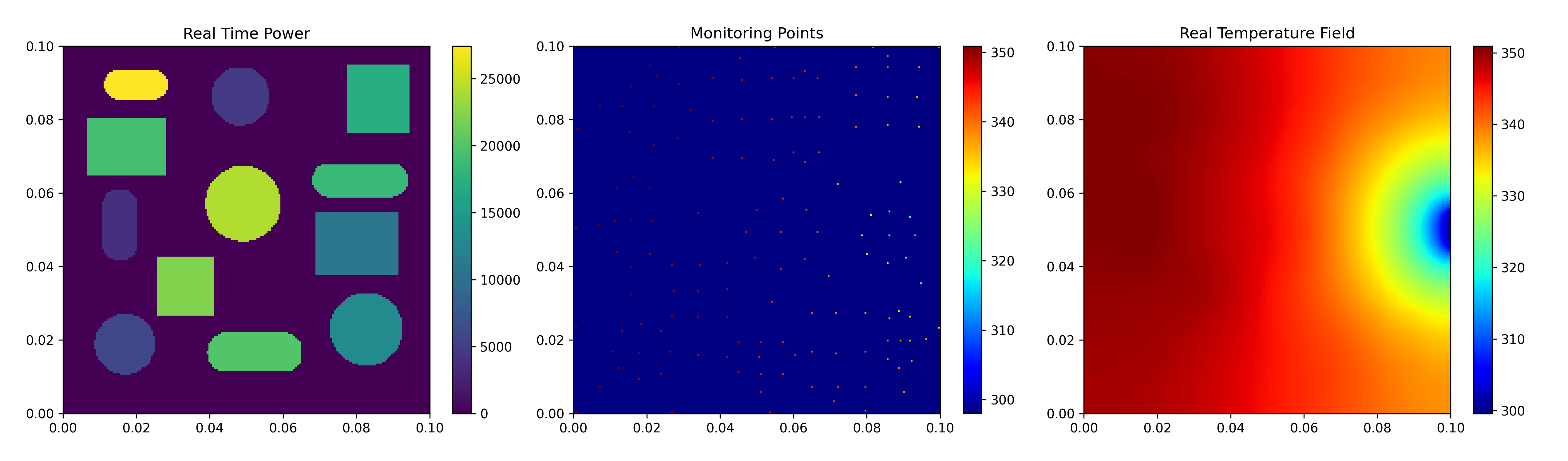}}
   \caption{Different simulation analysis datasets.}
\label{fig:data}
\end{figure}

As for the  monitoring points, Table \ref{table:point} lists the number of monitoring points for these datasets.  In this work, for temperature field reconstruction, the monitoring points are placed near the boundary, between the heat-sources and on the heat-sources, respectively. As the table shows, 16 monitoring points are placed near the boundary. 
18 points are put between components for Data A, B, D and 16 points are put for Data C.
For Data A, B, and D, nine monitoring points are placed on each component.
While for Data C, nine monitoring points are placed for general heat sources, twelve are placed for complex sources with three parts and ten are placed for complex sources with two parts.
Such complex heat sources in Data C describe the complex components made of several different materials in engineering, such as central processing units (CPUs).

 In engineering, the numerically simulated methods, such as the finite element method (FEM) and the finite difference method (FDM), which solve the heat conduction equations mathematically, are generally used
for accurate thermal analysis. Therefore, for convenience but without loss of generality, this work takes
such temperature field solved by finite element method (FEM) as groundtruth for evaluation of TFR-HSS task \footnote{{Other interested researchers can also use FDM instead.}}. Our data generator \footnote{\url{https://github.com/shendu-sw/recon-data-generator}.} is constructed over the FEniCS \footnote{\url{https://fenicsproject.org/}}, one of the finite element simulation software. It supports the calculation of temperature field over heat source systems with
different components, boundary conditions.

\begin{table}
\begin{center}
\caption{Number of monitoring points used in this work for different datasets. NB, BC and OC represent near the boundary, between components and on the components, respectively. }
\label{table:general_performance}
\begin{tabular}{| c | c|c| c | c | }
\hline
{\bf Positions}    & {NB} & {BC}   &  {OC} & Total \\
\hline\hline
Data A \& B & 16 & 18 & $9\times 10$ & 124 \\
Data C & 16 & 16 & $9\times 5 + 12\times 3+ 10\times 2$ & 133 \\
Data D & 16 & 18 & $9\times 12$ & 142 \\

\hline
\end{tabular}
\end{center}
\label{table:point}
\end{table}

{\bf Optimization and hyperparameters: }
Considering the orders of magnitude of the four losses, $\alpha, \beta, \gamma$ are set to $1e^{-3}, 1e^{-3}, 1e^{-2}$, respectively. {It should be noted that slight changes of these hyperparameters are permitted. However, excessive changes would make the training process unstable, or even not converged}.
For each simulation data, we choose 40000 samples for training process where 80\% is for training and 20\% for validation, and 10000 samples for testing.
The training epoch is set to 50.

{ {\bf Model architecture:}
If not specified, the SegNet-AlexNet \cite{segnet} (SegNet with AlexNet backbone) is used as the base regression model of proposed reversible regression model. Both $Net_1$ and $Net_2$ in the reversible regression model uses the SegNet-AlexNet \footnote{More details about the models are provided in attachment.}. The codes of reproducing the proposed method for TFR-HSS task are released at \url{https://github.com/shendu-sw/PIRL}.
}

{\bf Compute Infrastructure: }
A very common machine with a 2.8-GHz Intel(R) Xeon(R) Gold 6242 CPU, 256-GB memory, and NVIDIA GeForce RTX 3090 GPU was used to test the performance of the proposed method.



\subsection{ General Performance}

{At first, we present a brief overview of the merits of the proposed physics-informed deep surrogate learning method for TFR-HSS task.}

\begin{figure}
\centering
\subfigure[Training loss vs epochs]{\label{fig:toa}\includegraphics[width=0.49\linewidth]{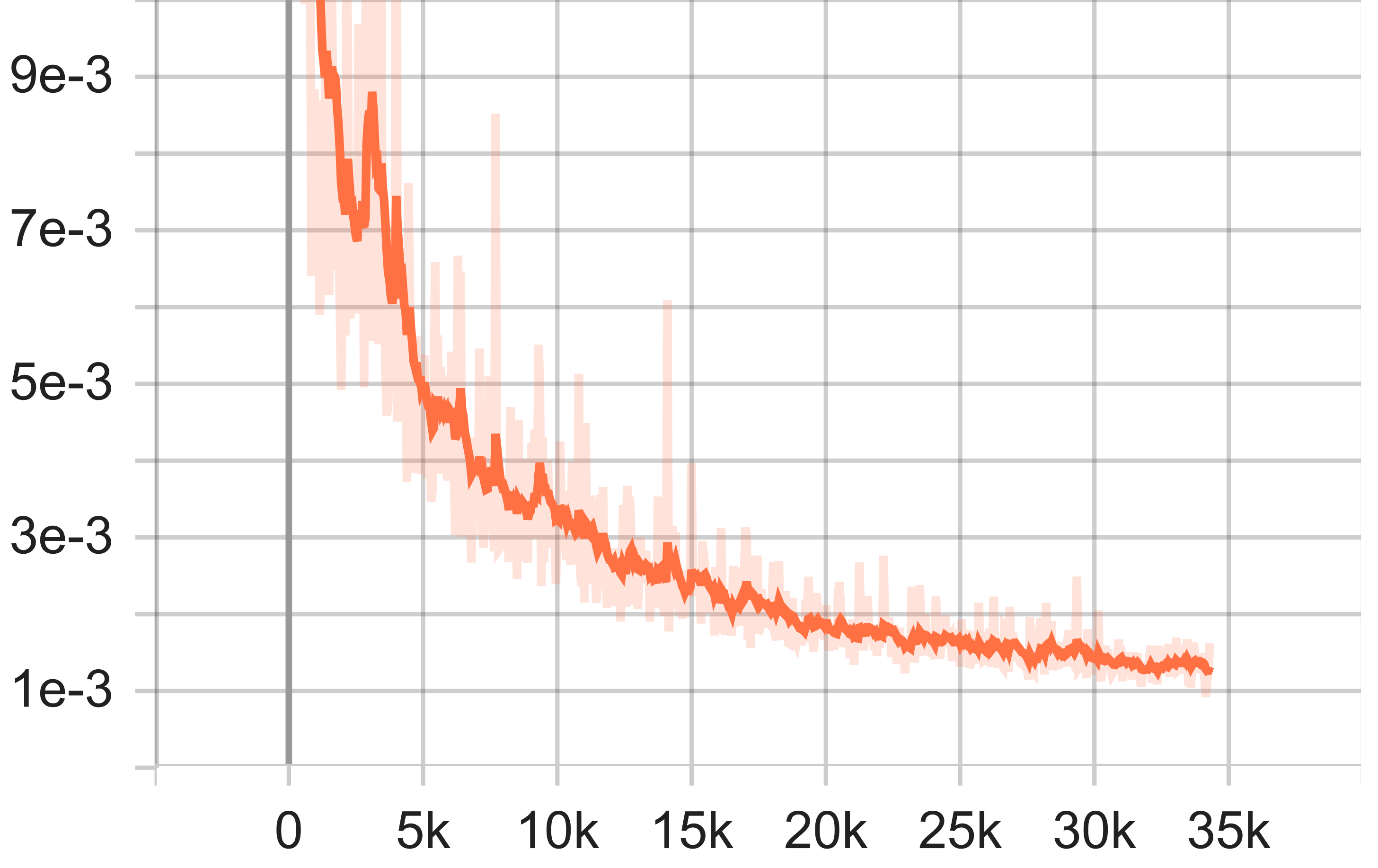}}
\subfigure[Validation loss vs epochs]{\label{fig:vob}\includegraphics[width=0.49\linewidth]{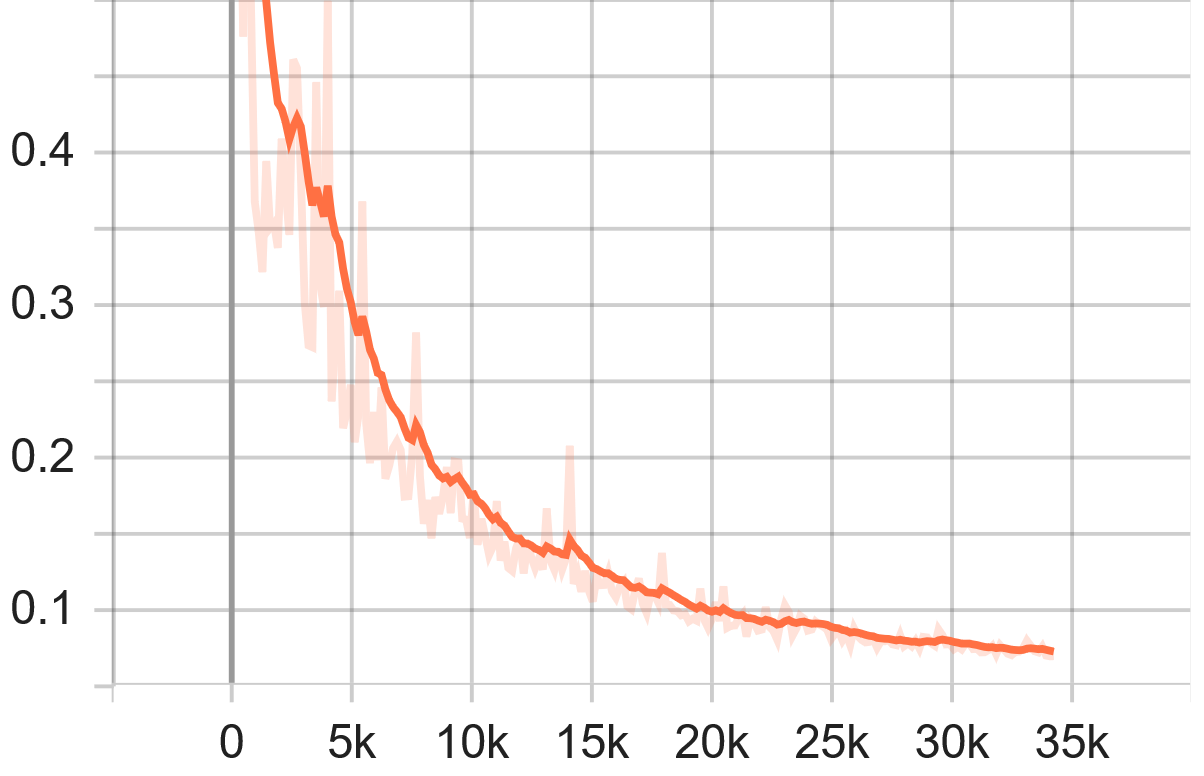}}
\subfigure[BC loss vs epochs]{\label{fig:bcc}\includegraphics[width=0.49\linewidth]{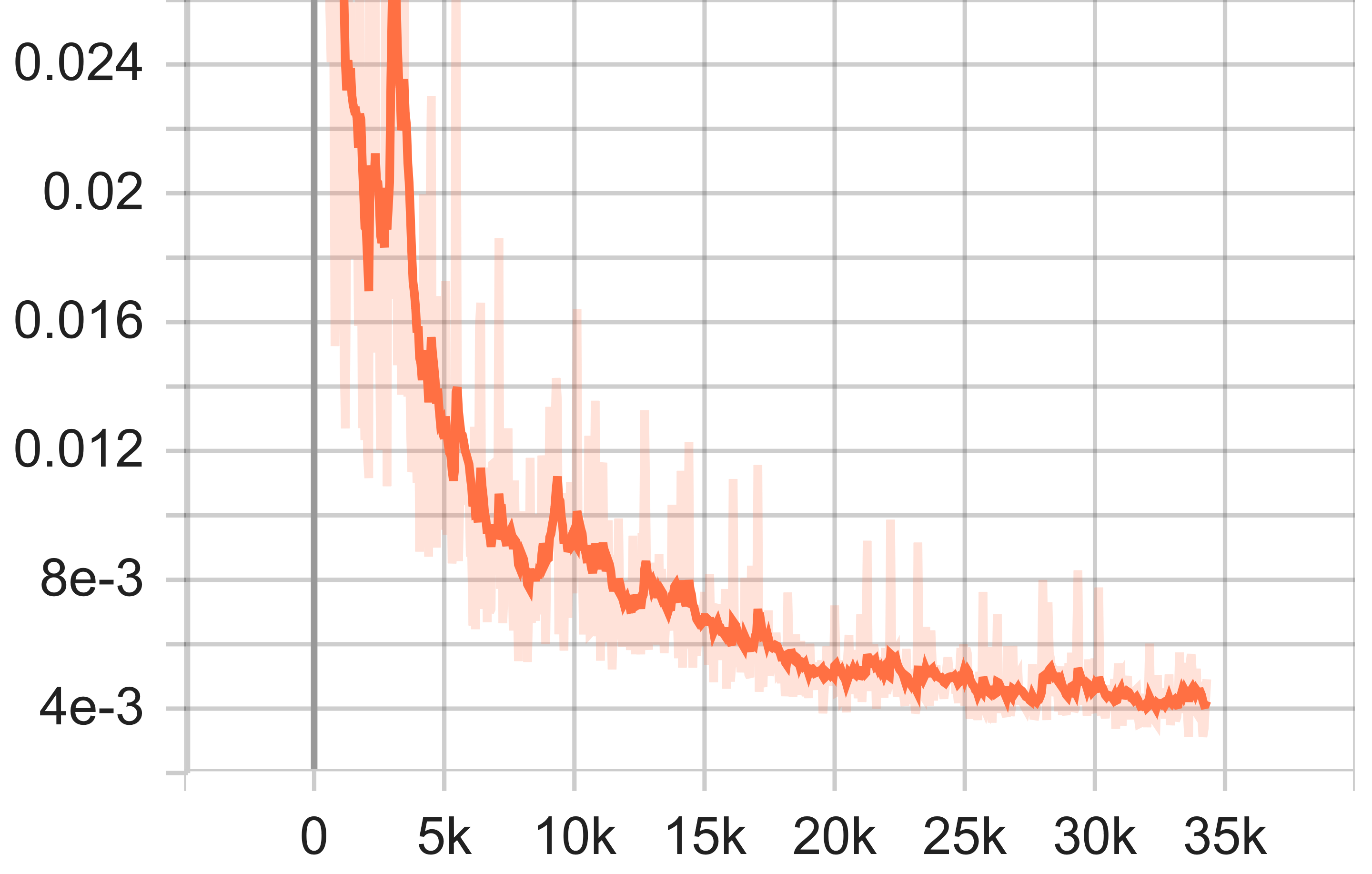}}
\subfigure[Laplace loss vs epochs]{\label{fig:lad}\includegraphics[width=0.49\linewidth]{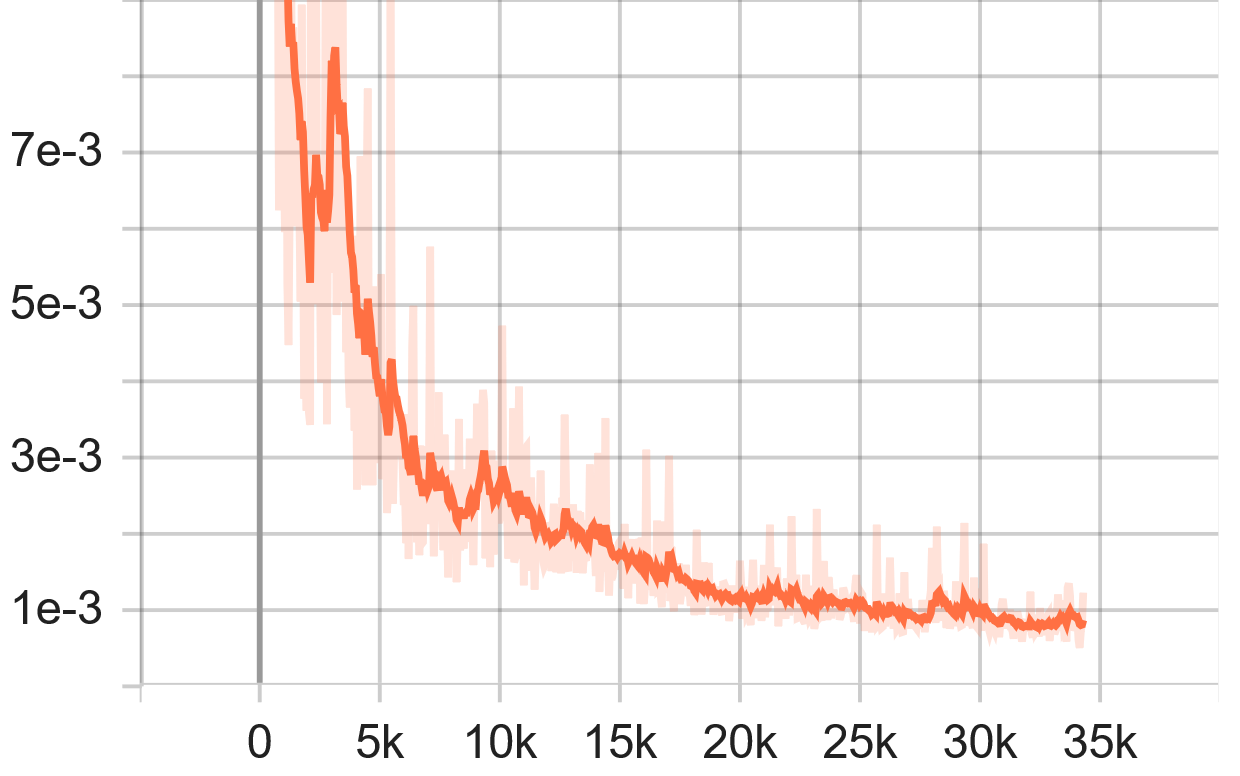}}
\subfigure[Point loss vs epochs]{\label{fig:ple}\includegraphics[width=0.49\linewidth]{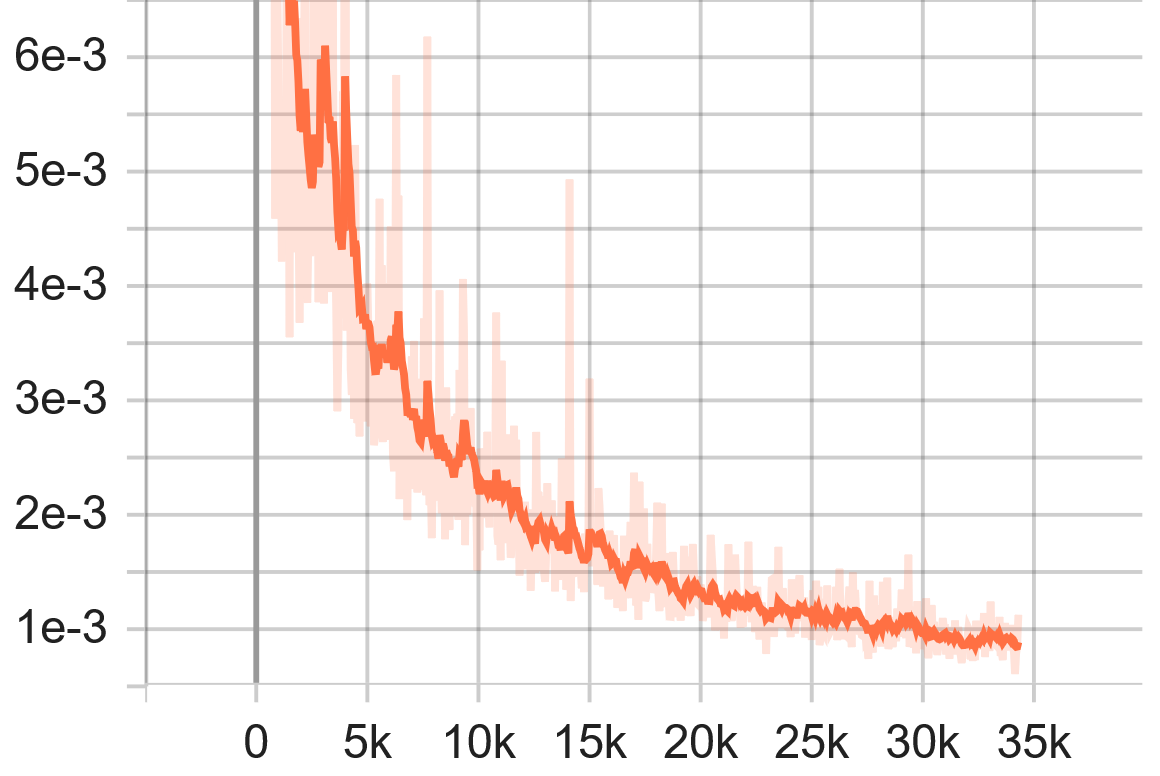}}
\subfigure[TV loss vs epochs]{\label{fig:tvf}\includegraphics[width=0.49\linewidth]{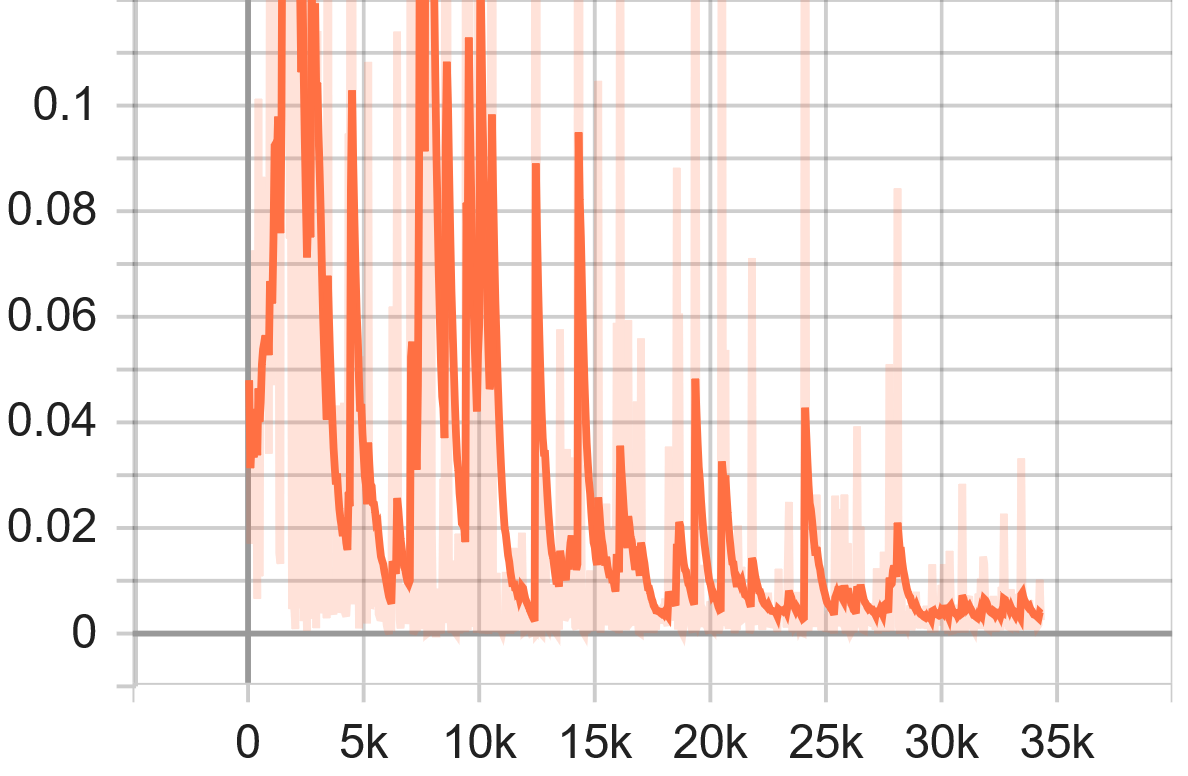}}
  \caption{Training curves of the proposed method over Data A under default experimental setups.}
\label{fig:training}

\end{figure}

 Fig. \ref{fig:training} presents the training curves vs epochs under the proposed method. 
The training curves in Fig. \ref{fig:toa} show that the proposed reconstruction loss can well train the {deep model unsupervisedly and promote} the convergence of the model. Compared Fig. \ref{fig:toa} with \ref{fig:vob},  we can find that the training process can be well generalized to validation data. 
Besides, Fig. \ref{fig:bcc} , \ref{fig:lad}, \ref{fig:ple}, and \ref{fig:tvf} show that these four losses can converge well and contribute to the training of the proposed model.

Table \ref{table:general_performance} lists the performance over Data A, B, C, D under different metrics and
Fig. \ref{fig:monitoring_9} presents several reconstruction examples of the proposed method over these datasets.
As the results show, the MAE of proposed method over all the datasets is less than 0.5K. Especially, over the components, the MAE is about 0.1K over all the datasets. Besides, the BMAE  is about or less than 1K over all the datasets. This means that the proposed method can better reconstruct the temperature both over the components and near the boundaries. As Fig. \ref{fig:monitoring_9} shows, the reconstruction error of only  few points near the boundaries is larger than 1K. This also indicates that the proposed reversible regression model can well work for the current task {without labelled samples}.

Obviously, the base models can significantly affect the performance of the proposed method. Therefore, following we will present the general reconstruction performance of the proposed method  with different surrogate models.

\begin{figure}
\centering
\subfigure[Data A]{\label{fig:1a}\includegraphics[width=0.9\linewidth]{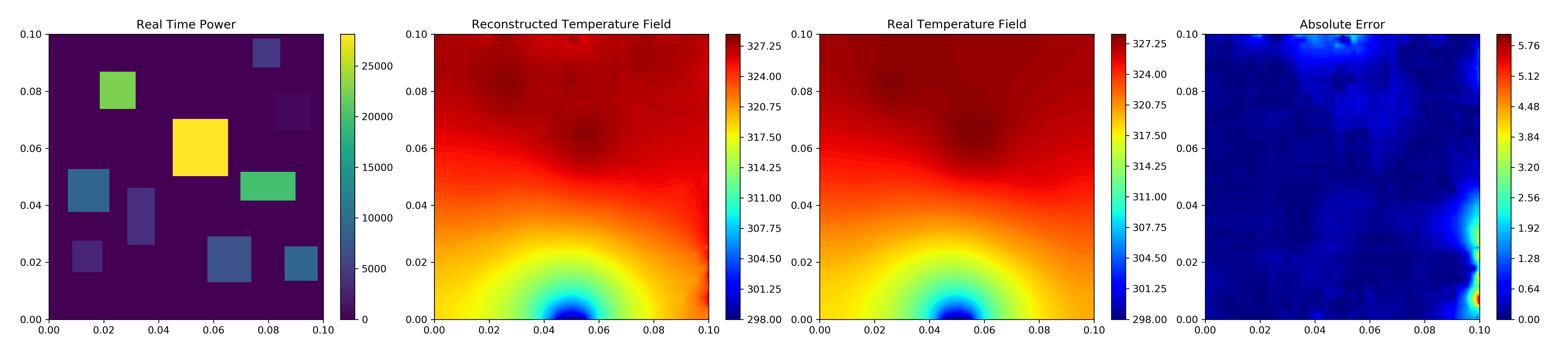}}
\subfigure[Data B]{\label{fig:1b}\includegraphics[width=0.9\linewidth]{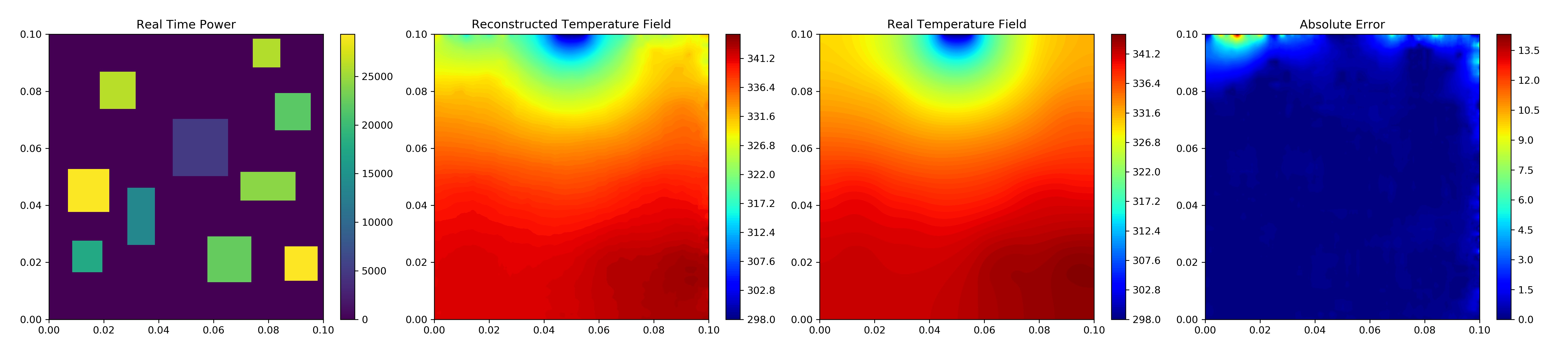}}
\subfigure[Data C]{\label{fig:1c}\includegraphics[width=0.9\linewidth]{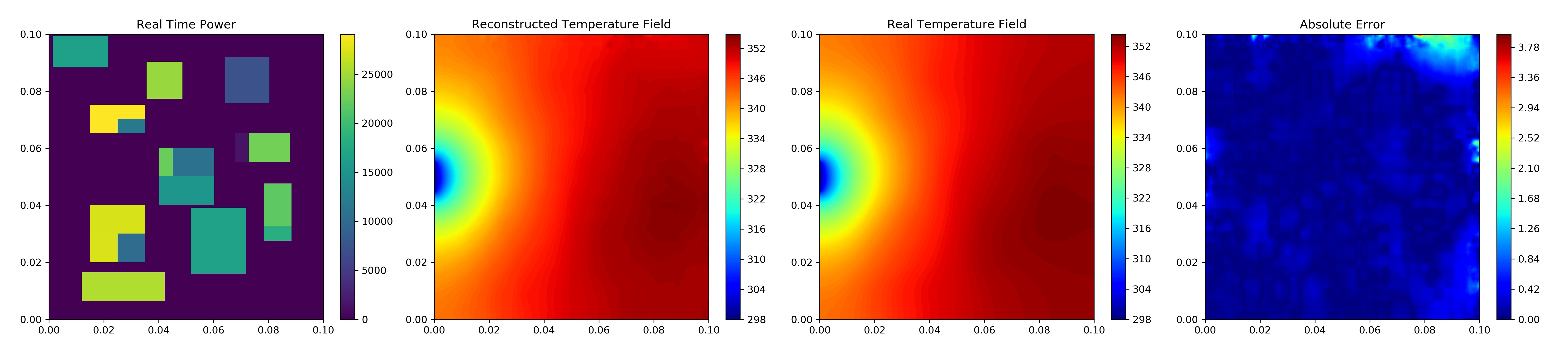}}
\subfigure[Data D]{\label{fig:1d}\includegraphics[width=0.9\linewidth]{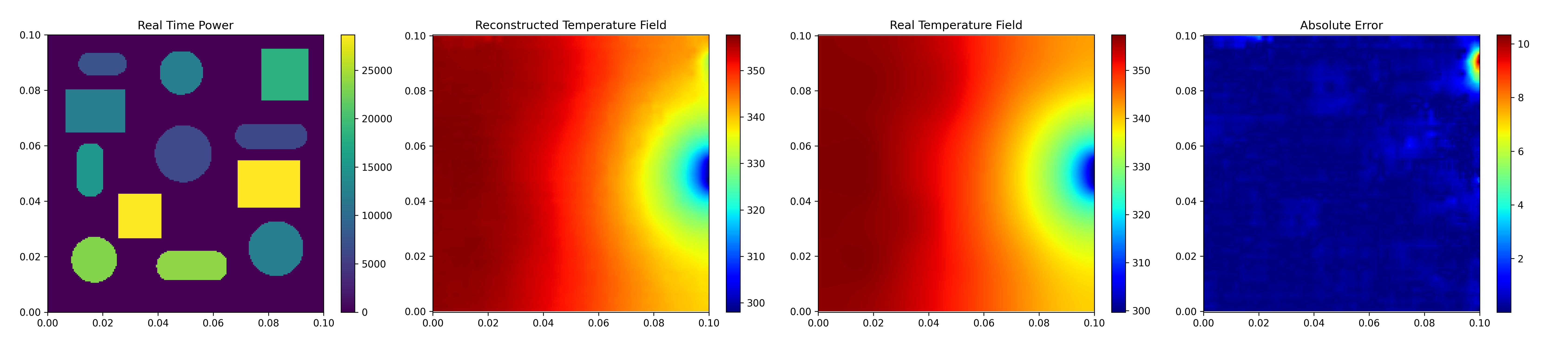}}
  \caption{Examples of proposed method over Data A, B, C and D for TFR-HSS task.}
\label{fig:monitoring_9}
\end{figure}

\begin{table}
\begin{center}
\caption{ Reconstruction performance of the proposed method under MAE (K), M-CAE  (K), CMAE (K) and BMAE (K) over datasets for TFR-HSS task (32000, 8000 samples for training, and validation, respectively). }
\label{table:general_performance}
\begin{tabular}{| c | c|c| c | c| }
\hline
{\bf Data} &  {MAE} & {M-CAE}   &  {CMAE} & BMAE  \\
\hline\hline
{Data A} &   0.3436 & 1.4340& 0.0883& 1.0483 \\
{Data B} & 0.2655 & 3.6059 & 0.1006 & 0.9003 \\
{Data C} & 0.1727 & 1.7394 & 0.1157 & 0.4560 \\
{Data D} & 0.2456 & 1.7043 & 0.1902 & 0.5751 \\
\hline
\end{tabular}
\end{center}
\end{table}

\subsubsection{Performance with Different Surrogate Models}\label{subsec:model_comparasion}

The base model of $Net_1$ and $Net_2$ in the proposed reversible regression model can significantly affect the reconstruction performance of the proposed method.
 {In this set of experiments, we compare the reconstruction performance under different vanilla surrogate models, namely the SegNet \cite{segnet}, Feature Pyramid Networks (FPN) \cite{fpn}, Fully Convolutional Networks (FCN) \cite{fcn}, and UNet \cite{unet}. The detailed architectures of the four models can be seen in the attachment.
}

{In this work, we construct four forms of the proposed reversible regression models, namely SegNet-SegNet, FCN-FCN, UNet-UNet, FPN-SegNet, respectively.  It should be noted that instead of FPN-FPN, we use the FPN-SegNet since the FPN-FPN does not converge for current TFR-HSS task.
32.6M, 10.5M, 62.1M, and 29.4M parameters  are existed in SegNet-SegNet, FCN-FCN, UNet-UNet, and FPN-SegNet, respectively.}
Table \ref{table:reversible_regression_model} lists the performance of the four configurations over the four datasets and Fig. \ref{fig:surrogate_models} shows the examples of the reconstruction results over Data A with different configurations. 

From the table, it can be find that FPN-SegNet performs better than other configurations over Data B and Data C under MAE while over Data A, UNet-UNet performs the best and over Data D, FCN-FCN perform the best. From the view of M-CAE, FPN-SegNet and UNet-UNet perform the best than other configurations while under CMAE and BMAE, UNet-UNet is better than other three configurations. 
The SegNet-SegNet performs the worst performance among all the four configurations.
 Overall, through thoroughly comparisons, the UNet-UNet architecture is most suited for the TFR-HSS task. 

For simplicity and better presenting the performance of the proposed method affected by other variables, the SegNet-SegNet is used as the base model in the proposed method unless otherwise specified.

\begin{figure}
\centering
\subfigure[Results by SegNet-SegNet]{\label{fig:1a}\includegraphics[width=0.9\linewidth]{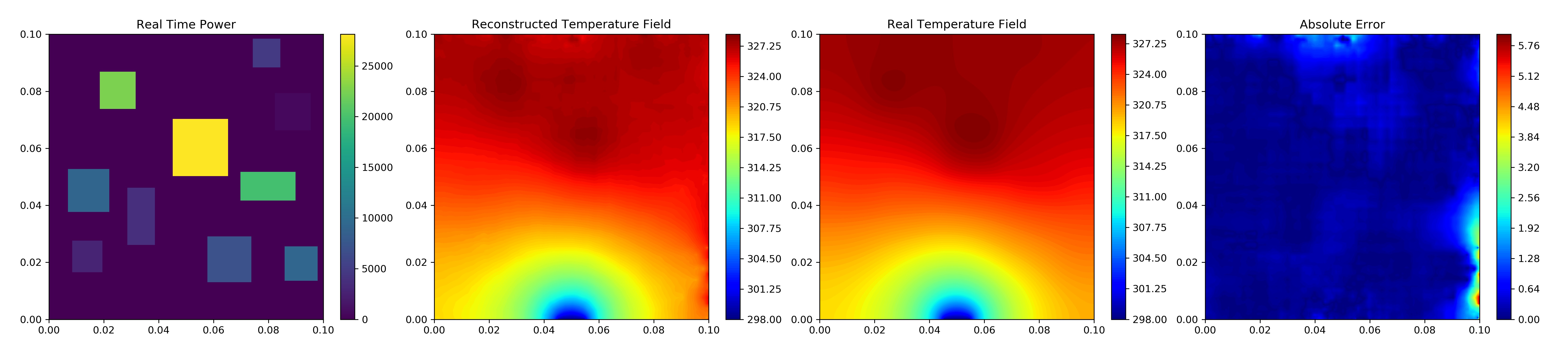}}
\subfigure[Results by FCN-FCN]{\label{fig:1b}\includegraphics[width=0.9\linewidth]{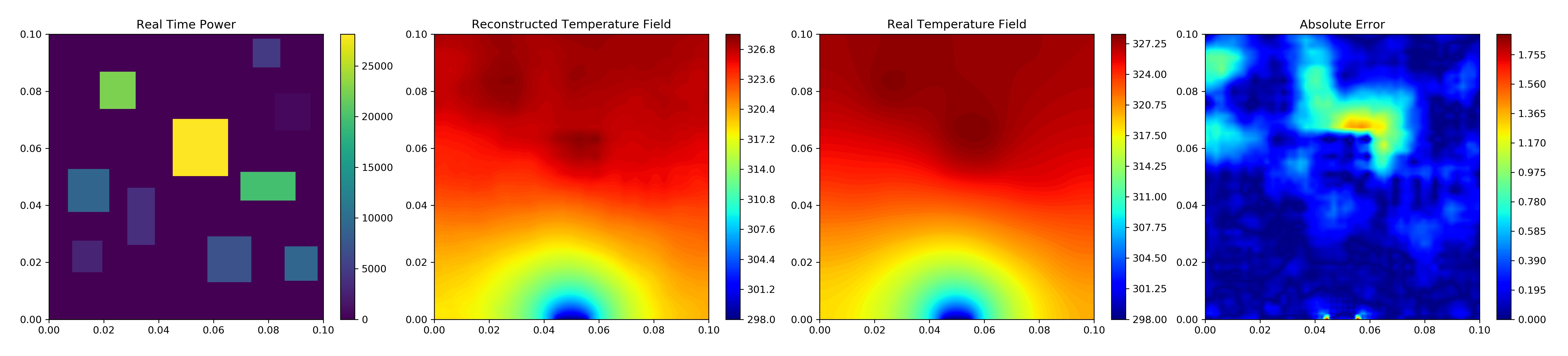}}
\subfigure[Results by FPN-SegNet]{\label{fig:1c}\includegraphics[width=0.9\linewidth]{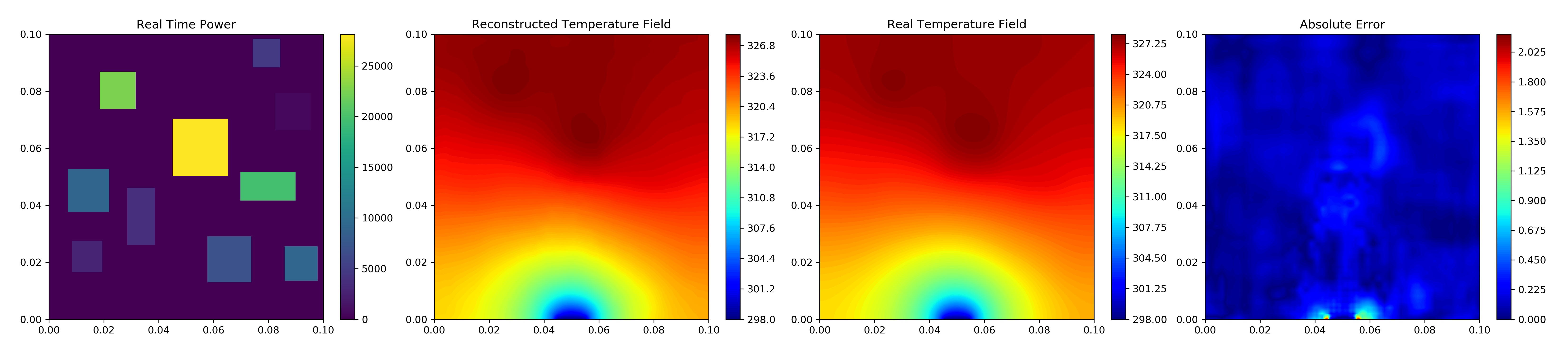}}
\subfigure[Results by Unet-Unet]{\label{fig:1d}\includegraphics[width=0.9\linewidth]{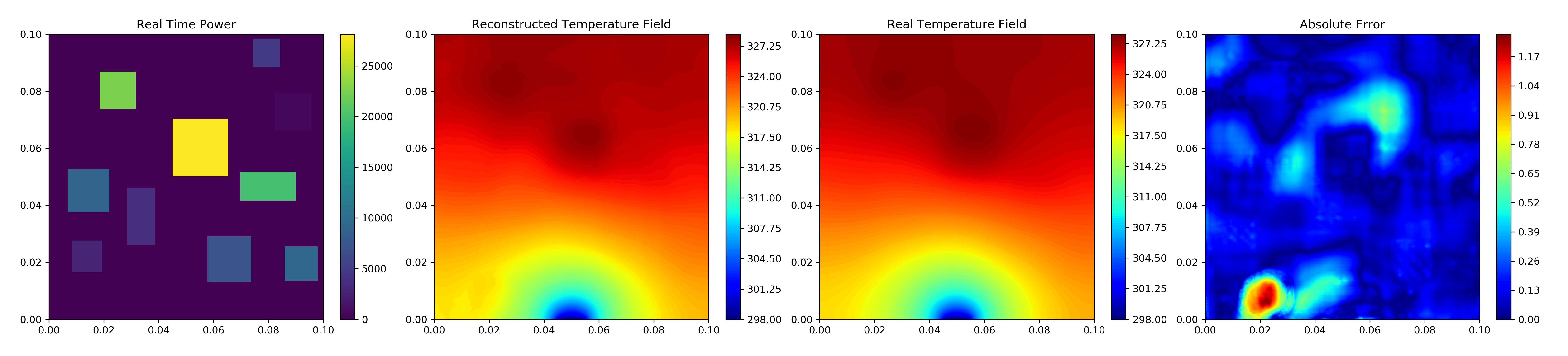}}
  \caption{Examples of the reconstruction results over Data A by proposed reversible regression models with different base models.}
\label{fig:surrogate_models}
\end{figure}

\begin{table}
\begin{center}
\caption{Reconstruction performance (K) of proposed physics-informed deep learning on reversible regression model with different base models for TFR-HSS task. }
\label{table:reversible_regression_model}
\begin{tabular}{| c| c | c|c| c | c | }
\hline
 {\bf Models}&  Metrics& {Data A} & {Data B}   &  {Data C} &  {Data D}  \\
\hline\hline
\multirow{4}{*}{{SegNet-SegNet}} & MAE & 0.3436 & 0.2655  & 0.1727 & 0.2456 \\
 & M-CAE & 1.4340 & 3.6059  & 1.7394 & 1.7043 \\
 & CMAE & 0.0883 & 0.1006  & 0.1157 & 0.1902 \\
& BMAE & 1.0483 & 0.9003  & 0.4560 & 0.5751 \\
\hline
\multirow{4}{*}{{FCN-FCN}} & MAE & 0.1309 & 0.1605  & 0.1051 & 0.1487 \\
 & M-CAE & 0.6310 & 0.6672  & 0.4913 & 0.6505 \\
 & CMAE & 0.0865 & 0.1065  & 0.1078 & 0.1674 \\
& BMAE & 0.1614 & 0.1847  & 0.1589 & 0.1941 \\
\hline
\multirow{4}{*}{{FPN-SegNet}} & MAE & 0.1503 & 0.1510  & 0.0931 & 1.1305 \\
 & M-CAE & 0.4059 & 0.4631  & 0.2941 & 4.5548 \\
 & CMAE & 0.1429 & 0.1393  & 0.0820 & 0.8288 \\
& BMAE & 0.1654 & 0.1656  & 0.1502 & 2.7322 \\
\hline
\multirow{4}{*}{{UNet-UNet}} & MAE & 0.1265 & 0.1540  & 0.1146 & 0.1979 \\
 & M-CAE & 0.3283 & 0.5230  & 0.3035 & 0.6429 \\
 & CMAE & 0.0703 & 0.0983  & 0.0700 & 0.2277 \\
& BMAE & 0.1582 & 0.1587  & 0.1250 & 0.2297 \\
\hline
\end{tabular}
\end{center}
\end{table}

\subsection{Comparisons with vanilla Deep Regression Models}

In this set of experiments, we compare the proposed reversible regression model with vanilla deep regression models which are both trained by the {proposed physics-informed reconstruction loss.}
Here, the proposed reversible regression model {is implemented under two different forms:} the reversible surrogate model with flip operation ($\text{RSM}_{w}$), and the reversible surrogate model without flip operation ($\text{RSM}_{wo}$). 

\begin{figure}
\centering
\subfigure[]{\label{fig:s1a}\includegraphics[width=0.49\linewidth]{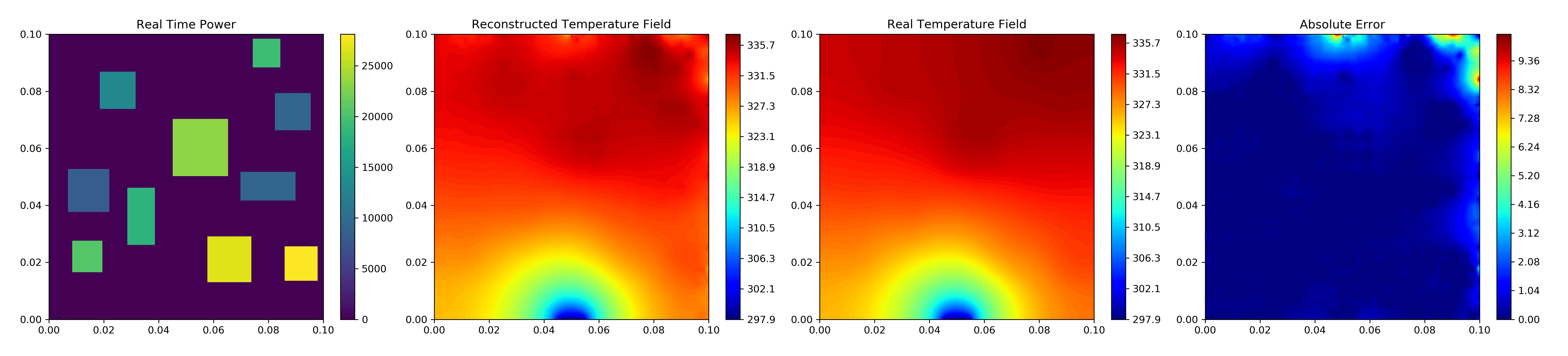}}
\subfigure[]{\label{fig:s1b}\includegraphics[width=0.49\linewidth]{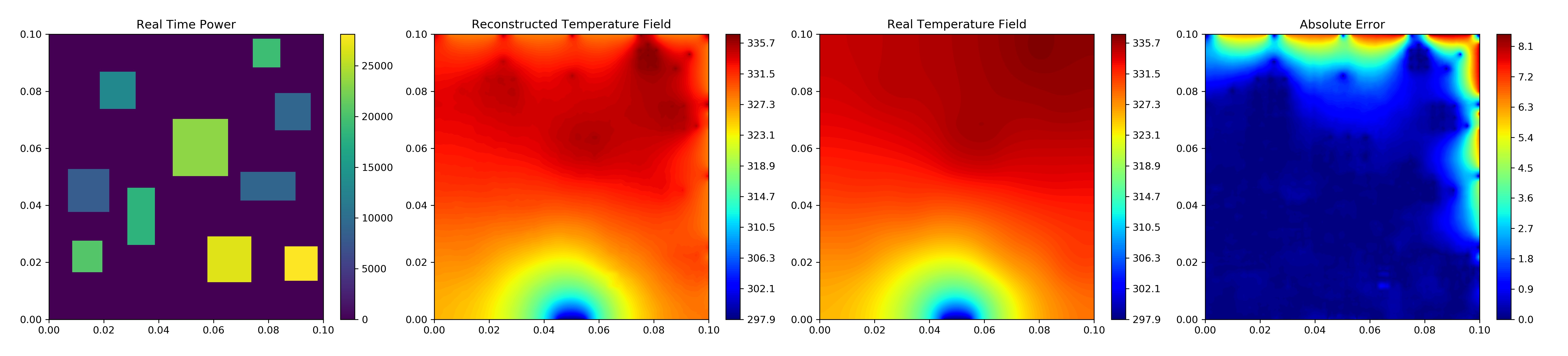}}
\subfigure[]{\label{fig:s2a}\includegraphics[width=0.49\linewidth]{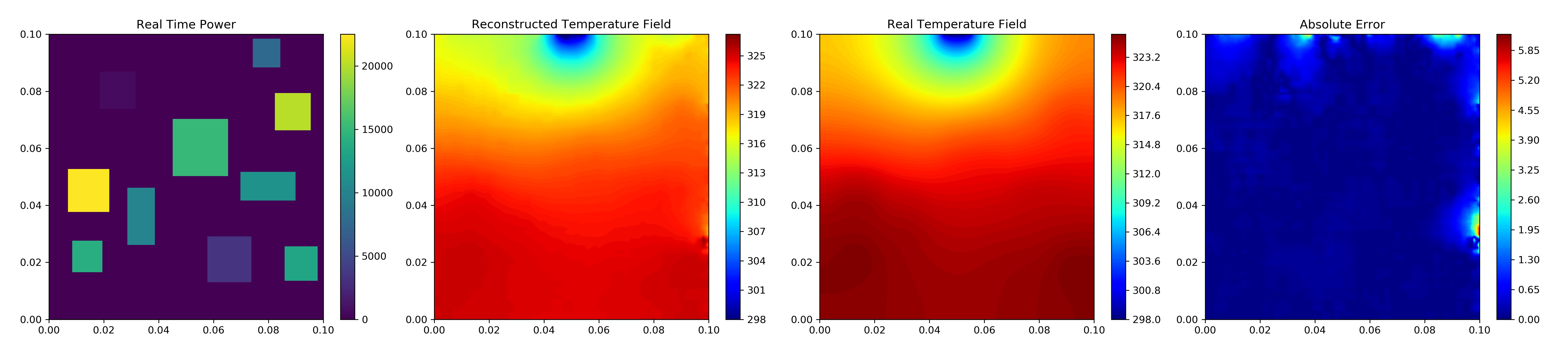}}
\subfigure[]{\label{fig:s2b}\includegraphics[width=0.49\linewidth]{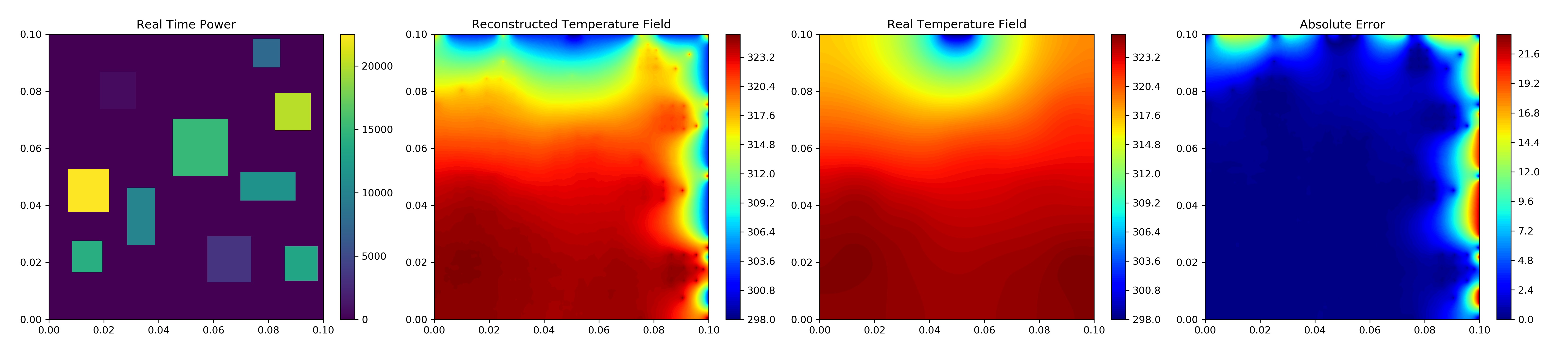}}
\subfigure[]{\label{fig:s3a}\includegraphics[width=0.49\linewidth]{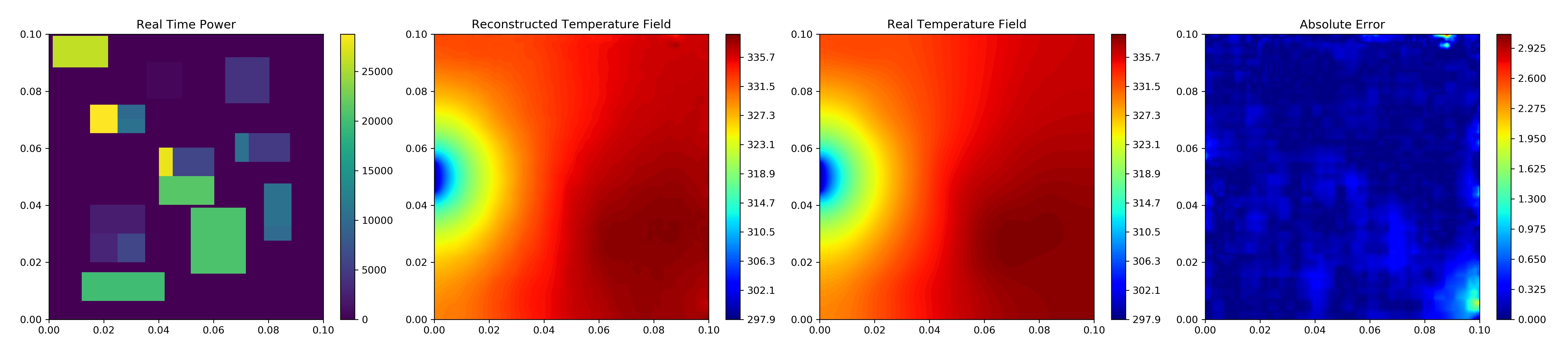}}
\subfigure[]{\label{fig:s3b}\includegraphics[width=0.49\linewidth]{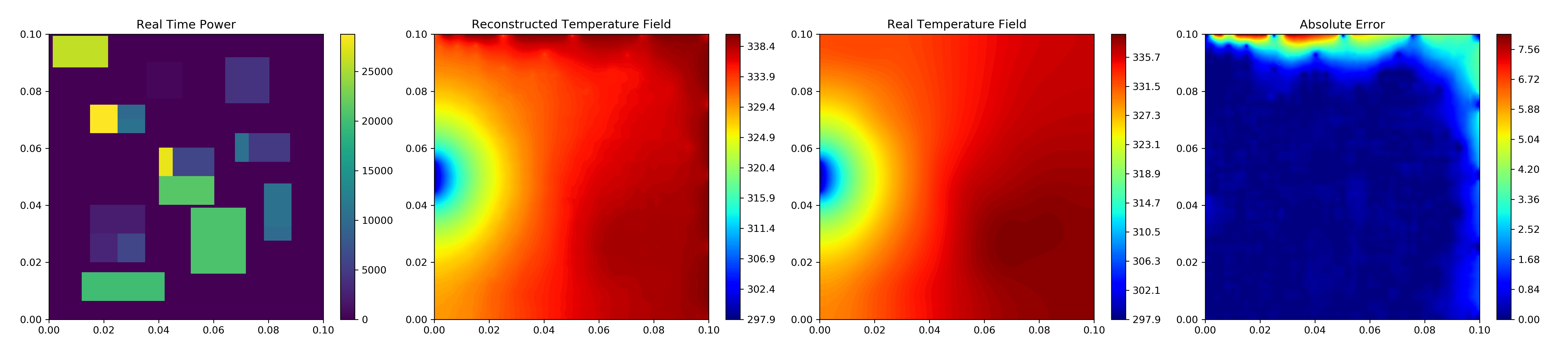}}
\subfigure[]{\label{fig:s4a}\includegraphics[width=0.49\linewidth]{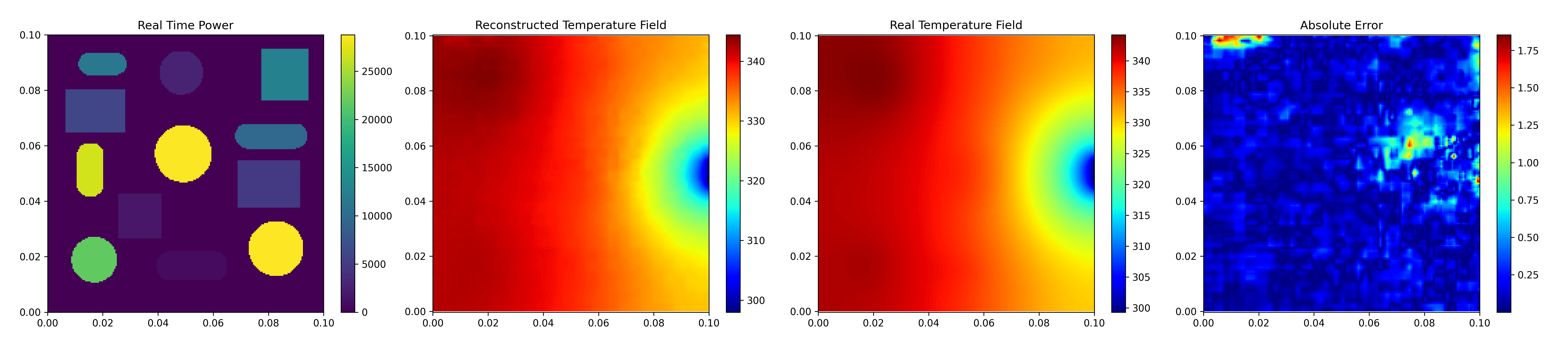}}
\subfigure[]{\label{fig:s4b}\includegraphics[width=0.49\linewidth]{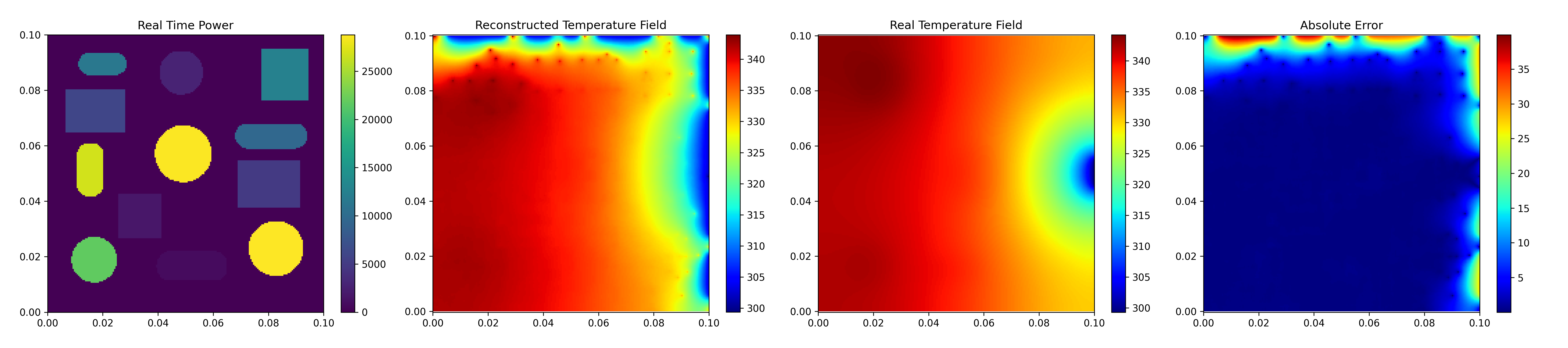}}
  \caption{Examples of the reconstruction results over different datasets by proposed reversible regression model with flip operation and vanilla deep regression model. (a), (c), (e), (g) describes example by proposed method over Data A, B, C, D, respectively; (b), (d), (f), (h) describes example by vanilla model over Data A, B, C, D, respectively.}
\label{fig:vanilla}
\end{figure}

Table \ref{table:surrogate_models} lists the comparison results. 
In the table the vanilla deep regression models is general SegNet model. For the proposed method, both $Net_1$ and $Net_2$ select the SegNet model.
{The proposed methods including the $\text{RSM}_{w}$ and $\text{RSM}_{wo}$ present better performance} than vanilla deep regression model. Especially, the proposed method can better reconstruct the temperature field near the boundaries. From the view of BMAE, the proposed method with flip operation presents {BMAE} of  1.0483K, 0.9003K, 0.4560K, 0.5751K over Data A, B, C, D, respectively, is better than the performance of vanilla regression models where the BMAE is 2.3134K, 8.2632K, 2.9737K, 14.0953K, respectively. 
From Fig. \ref{fig:vanilla}, it is also obvious that the proposed method can better reconstruct the temperature field, especially the field near the boundaries.
Furthermore, compare the experimental results between the  $\text{RSM}_{w}$ and $\text{RSM}_{wo}$, and we can find that the flip operation can improve the reconstruction performance. 
{With vanilla deep regression models, the reconstructed temperature field usually has jagged boundary temperature, especially the upside and rightside of the reconstructed temperature field as Fig. \ref{fig:vanilla} shows. 
This is because that the forward and backward of convolutional layer in deep model is conducted orderly, which makes the final calculated area cannot capture the physical information well. Just as Fig. \ref{fig:s1b}, \ref{fig:s2b}, \ref{fig:s3b}, and \ref{fig:s4b} shows, the temperature field near the upside and rightside boundary  cannot be well reconstructed.}

\begin{table}
\begin{center}
\caption{Reconstruction performance (K) of proposed reversible regression model and the vanilla deep regression model for TFR-HSS task. In the table, 'Vanilla' represents the vanilla deep regression model, '$\text{RSM}_{wo}$' means the reversible surrogate model without flip operation and '$\text{RSM}_{w}$' means the reversible surrogate model with flip operation.}
\label{table:surrogate_models}
\begin{tabular}{| c| c | c|c| c | c | }
\hline
 {\bf Models}&  Metrics& {Data A} & {Data B}   &  {Data C} &  {Data D}  \\
\hline\hline
\multirow{4}{*}{{Vanilla}} & MAE & 0.7640 & 2.4992  & 0.9532 & 3.5093 \\
 & M-CAE & 4.4597 & 14.1941  & 6.5920 & 15.7004 \\
 & CMAE & 0.2275 & 0.8505  & 0.3544 & 1.4107 \\
& BMAE & 2.3134 & 8.2632  & 2.9737 & 14.0953 \\
\hline
\multirow{4}{*}{{$\text{RSM}_{wo}$}} & MAE & {0.5729} & {0.3028}  & {0.6559} & {0.3325} \\
 & M-CAE & { 3.3061} & {\bf 1.8298}  & {5.6617} & {3.1053} \\
 & CMAE & { 0.1949} & {0.1220}  & {0.2570} & {0.3202} \\
& BMAE & { 1.7699} & {\bf 0.8976}  & {2.1814} & {\bf 0.3654} \\
\hline
\multirow{4}{*}{{$\text{RSM}_{w}$}} & MAE & {\bf 0.3436} & {\bf 0.2655}  & {\bf 0.1727} & {\bf 0.2456} \\
 & M-CAE & {\bf 1.4340} & { 3.6059}  & {\bf 1.7394} & {\bf 1.7043} \\
 & CMAE & {\bf 0.0883} & {\bf 0.1006}  & {\bf 0.1157} & {\bf 0.1902} \\
& BMAE & {\bf 1.0483} & { 0.9003}  & {\bf 0.4560} & { 0.5751} \\
\hline
\end{tabular}
\end{center}
\end{table}

\subsection{Ablation Studies}


\subsubsection{Performance with Different Training Samples}

\begin{figure}
\centering
\subfigure[MAE]{\label{fig:1a}\includegraphics[width=0.49\linewidth]{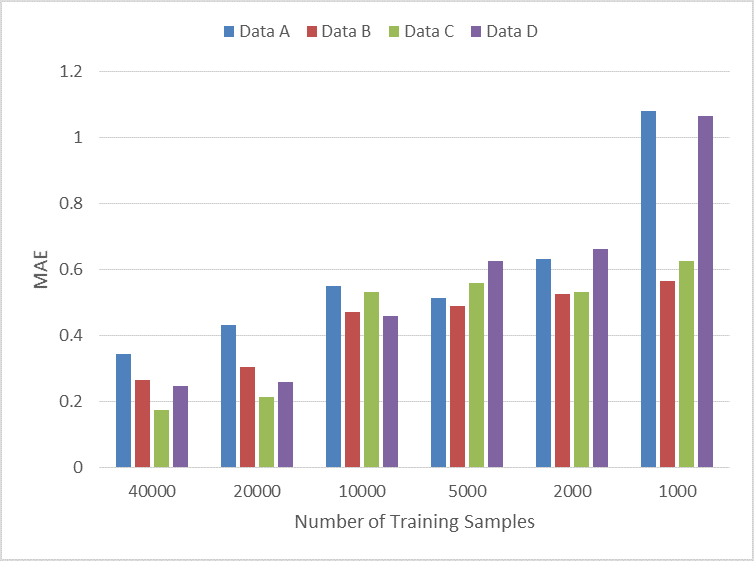}}
\subfigure[M-CAE]{\label{fig:1b}\includegraphics[width=0.49\linewidth]{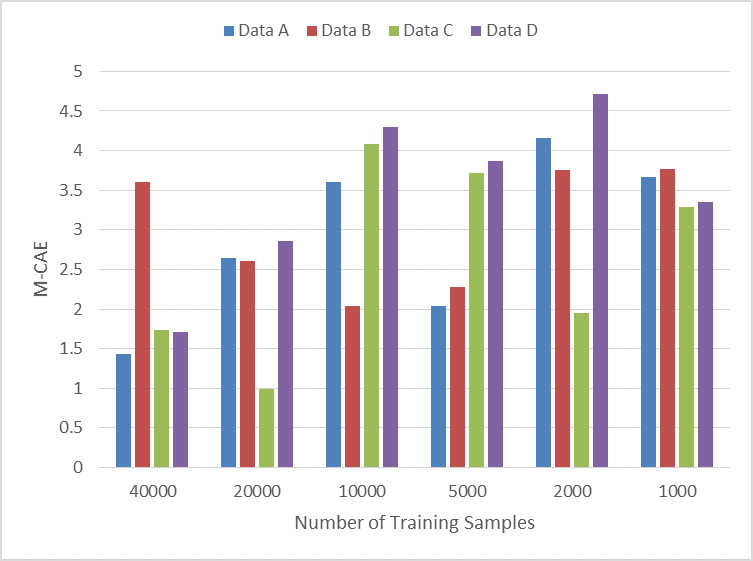}}
\subfigure[CMAE]{\label{fig:1a}\includegraphics[width=0.49\linewidth]{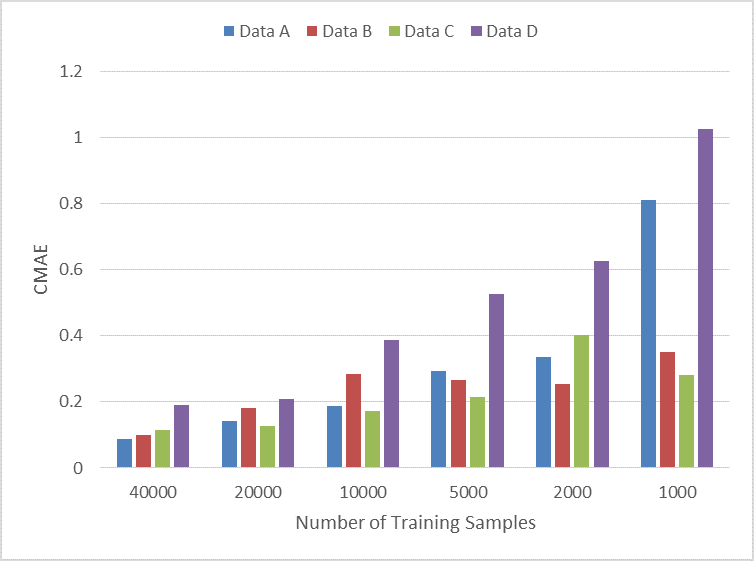}}
\subfigure[BMAE]{\label{fig:1b}\includegraphics[width=0.49\linewidth]{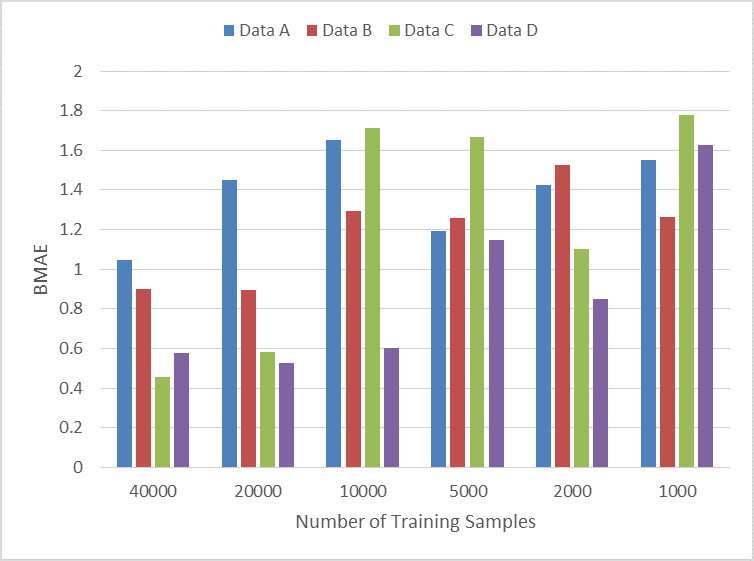}}
  \caption{Performance of proposed method with different number of unlabelled training samples for TFR-HSS task.}
\label{fig:number}
\end{figure}

In this subsection, we conduct experiments of the proposed method with different number of training samples.
The number of training samples is chosen from \{1000, 2000, 5000, 10000, 20000, 40000\}. Fig. \ref{fig:number} shows the reconstruction performance under different metrics over the four datasets.

Just the figure shows, the reconstruction performance tends to be better with the increase of the training samples. More training samples can help the proposed surrogate model to better learn the physical correlation between different points in the system.
It should also be noted that when the number of training samples increases to a certain level, the improvement of performance tends to be slight due to the limited of the training loss as well as the deep model. 
{Our proposed method can realize the unsupervised learning of the reconstruction model and the model can be constructed without labelled samples. However, unlabelled samples with monitoring information are also required for the better learning of the deep model.}

\subsubsection{Performance with Different Hyperparameters}

In order to reconstruct the temperature field unsupervisedly with only monitoring information, the proposed method {constructs} four different losses in order to use the physical information of temperature field, namely the Point loss, the BC loss, the Laplace loss, and the TV loss (see subsection \ref{subsec:loss} for details). These four losses play a different role in the training of the the deep model. To show the effect of {different losses on the reconstruction performance,} this subsection compares the performance of the proposed method {without  one of these losses.
Table \ref{table:loss_performance} shows the comparison results where $\alpha=0$ denotes the proposed training without BC loss, $\beta=0$ denotes the proposed training without Laplace loss, and $\gamma=0$ describes the proposed training without the TV loss.}

First of all, from the table {we can find that the proposed method can help to improve the most of the performance under all these four metrics.} For Data A, when $\beta=0$ the MAE is better than the original one. {Under $\beta=0$}, the BMAE tend to be 0.6108K which is better than {1.0483K obtained by} original one. This means that for Data A, the slight drop of performance of original training using laplace loss is mainly because of the BC loss instead of the laplace loss. 

{By inspecting the reconstruction performance with $\gamma=0$, we can} find that the TV loss, which takes advantage of the neighbor correlations, plays an important role in the reconstruction of temperature field near the boundary as well as the field over the component. From the table, it can be also noted that the performance drops the most without the TV loss when compared with all the other losses.

In addition, {by comparing} the performance when $\alpha=0$ with the original one, we can find that the BMAE drops the most. This indicates that the BC loss plays an important role in the temperature field reconstruction near the boundaries.   

\begin{table}
\begin{center}
\caption{Reconstruction performance (K) of proposed method with different hyperparameters for TFR-HSS task. }
\label{table:loss_performance}
\begin{tabular}{| c | c| c|c| c | c | }
\hline
{\bf Setting} & {\bf Metric}& {Data A} & {Data B}   &  {Data C} &  {Data D}  \\
\hline\hline
$\alpha=0$ & \multirow{4}{*}{{MAE}} & 0.5582 & 0.6050  & 0.6372 & 0.6942 \\
$\beta=0$ & & {\bf 0.2144} & 0.5583 & 0.6432& 0.3930 \\
$\gamma=0$ & & 1.3561 &1.7778 &1.5841 &2.6735 \\
original & & 0.3436 & {\bf 0.2655} & {\bf 0.1727} & {\bf 0.2456} \\
\hline
$\alpha=0$ & \multirow{4}{*}{{M-CAE}} & 2.4146 & {\bf 1.4672} & 2.2615 & 5.2712 \\
$\beta=0$ &  & {\bf 1.3816} & 6.6986&4.1077 &2.2869 \\
$\gamma=0$ & &9.3194 &14.1046 &6.9657 &25.2049 \\
original & &1.4340 &3.6059 &{\bf 1.7394} & {\bf 1.7043} \\
\hline
$\alpha=0$ & \multirow{4}{*}{{CMAE}} & 0.1471 & 0.1310 & 0.2799 & 0.3935 \\
$\beta=0$ &  & 0.1180 &0.1787 &0.2309 & 0.3156 \\
$\gamma=0$ & &0.7397 &0.6592 &0.6852 &2.9438 \\
original & & {\bf 0.0883} &{\bf 0.1006} &{\bf 0.1157} & {\bf 0.1902} \\
\hline
$\alpha=0$ & \multirow{4}{*}{{BMAE}} & 1.6682 & 1.5572 & 1.6145 & 1.9514 \\
$\beta=0$ &  &{\bf 0.6108} & 1.9804& 1.9854& 0.8724\\
$\gamma=0$ & &3.8190 & 9.5846&5.5920 &4.8308 \\
original & & 1.0483& {\bf 0.9003}& {\bf 0.4560} & {\bf 0.5751} \\
\hline
\end{tabular}
\end{center}
\end{table}

Overall, all the training losses in the proposed method play an important role in {unsupervised learning for temperature field reconstruction.
Under the jointly learning of all these losses, the proposed physics-informed reconstruction loss can train the deep model unsupervisedly without training samples. The proposed loss can alleviate the requirements of large amounts of labelled samples, which can be difficult to obtained or unavailable in real-world engineering.}  


\subsection{Comparisons with Other Methods}

This work uses the {global gaussian interpolation (GGI) \cite{tfrd_gong}, Gaussian Process Regression (GPR) \cite{ nr7}, support vector regression (SVR) \cite{svr}, polynominal regression(PR) \cite{pr}, neural networks (NN) \cite{nn} as baselines. All the baseline methods are reimplemented for TRF-HSS task.}

The global gaussian interpolation is proposed by our prior work \cite{tfrd_gong} which can utilize the global information instead of the local information. The reconstructed temperature at $(x_0,y_0)$ is related to all the monitoring points, and it can be formulated as
\begin{equation}
T(x_0,y_0)=\sum_{i=1}^m \frac{e^{-|(x_0-x_{s_i})^2+(y_0-y_{s_i})^2|_2}}{\sum_{j=1}^m e^{-|(x_0-x_{s_j})^2+(y_0-y_{s_j})^2|_2}}f(x_{s_i},y_{s_i}).
\end{equation}
The Gaussian process regression is implemented with the dace toolbox. For polynominal regression, the degree of the polynomial fit is set to 5. For neural network, the structure of the network is set to `2-10-10-1'.  The support vector regression is realised with the `sklearn.svm' package.  

Table \ref{table:comparisons} lists the comparison results with these former methods. Inspect the table and we can obtain the following conclusions.
First, the proposed method can obtain a better performance when compared with other methods. The proposed method can obtain a MAE of 0.1503K, 0.1510K, 0.0931K, and 0.1487K over the four datasets, respectively, outperforms all the other methods.
Besides, the proposed method costs less time for prediction.  Table \ref{table:comparisons} shows the prediction time of 10000 samples. We can find that for all the four datasets, the proposed method costs about 50s for prediction of 10000 testing samples benefiting from the use of GPUs while the fastest of other methods cost 4391.42s (NN over Data B). Especially, the SVR costs 88474.17s for prediction of Data C. {It should also be noted that our method costs less than 0.1s per testing samples with only CPUs which demonstrates that our method can also be applied in resource constrained environments.}

\begin{table}
\begin{center}
\caption{Reconstruction performance (K) of different methods for TFR-HSS task. The time, namely the prediction time is calculated over 10000 test samples.}
\label{table:comparisons}
\begin{tabular}{| c | c| c|c|c| c | c | }
\hline
{\bf Data} & {\bf Method}&{\bf  Time (s)}& {MAE} & {M-CAE}   &  {CMAE} &  {BMAE}  \\
\hline\hline
\multirow{6}{*}{{Data A}} & GGI & 13884.73s & 0.6148 & 3.3857  & 1.4215 & 2.6057 \\
 & GPR  & 8596.87s & 0.2799 & 1.2582  & 0.5800 & 2.4980 \\
 & SVR  & 36164.35s& 0.2664 & 0.4893  & {\bf 0.0787} & 2.5847 \\
 & PR  & 16421.41s & 0.5852 & 1.6478  & 1.3492 & 6.2793 \\
 & NN  & 4411.96s & 0.4232 & 2.4223  & 0.8504 & 3.2356 \\
 & PIRL & 52s & {\bf 0.1503} & {\bf 0.4059}  & 0.1429 & {\bf 0.1654} \\
\hline
\multirow{6}{*}{{Data B}} & GGI & 14143.03s & 0.7128 & 13.5829 & 2.6513 & 3.2147 \\
 & GPR  & 8783.92s & 0.3026 & 5.3661  & 1.0045 & 2.1256 \\
& SVR  & 30420.54s & 0.2938 & 0.4317  & {\bf 0.0824} & 3.5824 \\
 & PR  & 16766.82s & 0.6410 & 6.2734  & 2.0138 & 5.9980 \\
 & NN  & 4391.42s & 0.4938 & 8.7824  & 1.4837 & 3.2223 \\
 & PIRL & 47s & {\bf 0.1510} & {\bf 0.4631} & 0.1393 & {\bf 0.1656} \\
\hline
\multirow{6}{*}{{Data C}} & GGI & 16517.52s & 0.7693 & 19.3481 & 2.4615 & 3.6707 \\
 & GPR  & 11754.00s& 0.2998 & 7.9193  & 0.8648 & 2.0172 \\
& SVR  & 88474.17s & 0.2258 & 0.3524  & 0.0834 & 3.4857 \\
 & PR  & 18569.39s & 0.7478 & 9.0079 & 1.7457 & 8.8377 \\
 & NN  & 6796.88s & 0.5328 & 12.4782  & 1.5438 & 4.9832 \\
 & PIRL& 47s & {\bf 0.0931} & {\bf 0.2941} & {\bf 0.0820} & {\bf 0.1502} \\
\hline
\multirow{6}{*}{{Data D}}& GGI & 16812.54s & 0.541 & 3.9 & 0.331 & 0.709 \\
 & GPR  & 11673.16s & 0.458 & 1.7  & 0.303 & 1.26 \\
& SVR & 73571.12s & 0.894 & {13.4}  & {0.542} & 1.98 \\
 & PR  & 18654.51s & 0.813 & 3.48  & 0.585 & 2.43 \\
 & NN  & 6857.72s & 0.5874 & 3.283  & 0.559 & 1.852 \\
 & PIRL& 47s & {\bf 0.1487} & {\bf 0.6505}  & {\bf 0.1674}  & {\bf 0.1941} \\
\hline
\end{tabular}
\end{center}
\end{table}

{In conclusion, the proposed reconstruction method which joints the deep reversible regression model and physics-informed unsupervisedly learning can provide a nearly real-time prediction, which can improve the practicality of the proposed method for real-world engineering. Besides, the good reconstruction performance can guarantee the reliability for further thermal analysis.}

\subsection{Discussions}

{\bf Real-World Case Study}
To further validate the effectiveness of the propsed methods, we conduct additional experiments over real-world case study.

\begin{figure}
\centering
\subfigure[Components]{\label{fig:toa}\includegraphics[width=0.49\linewidth]{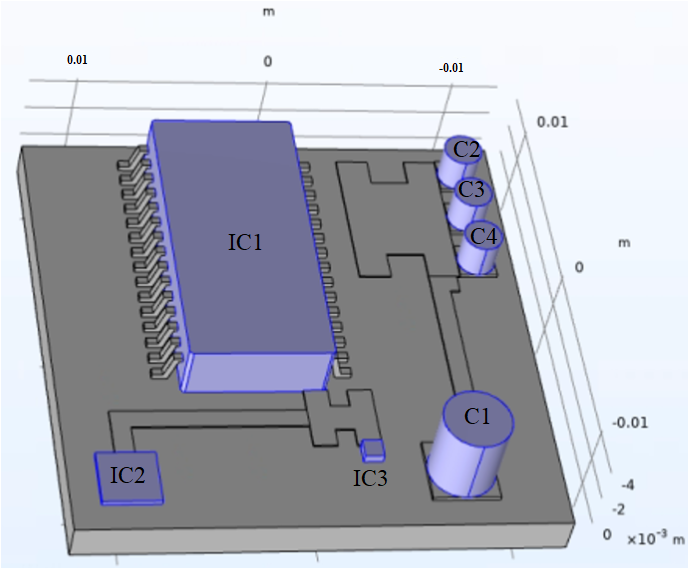}}
\subfigure[Fins]{\label{fig:vob}\includegraphics[width=0.49\linewidth]{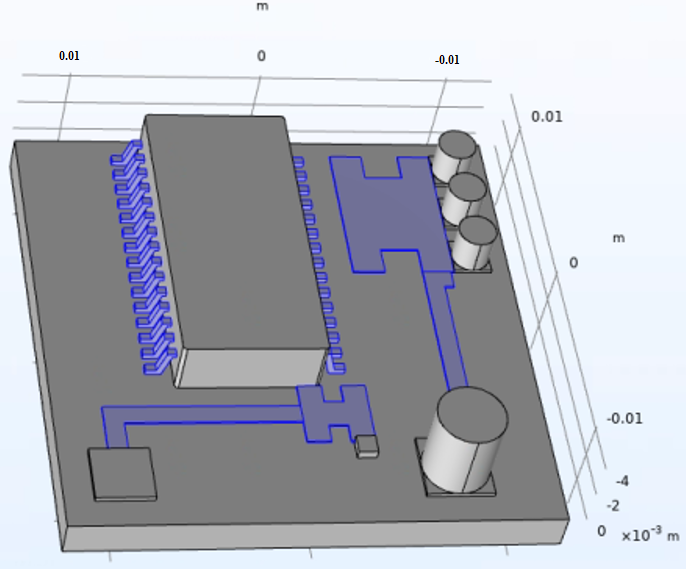}}
\subfigure[2-D Modeling]{\label{fig:voc}\includegraphics[width=0.49\linewidth]{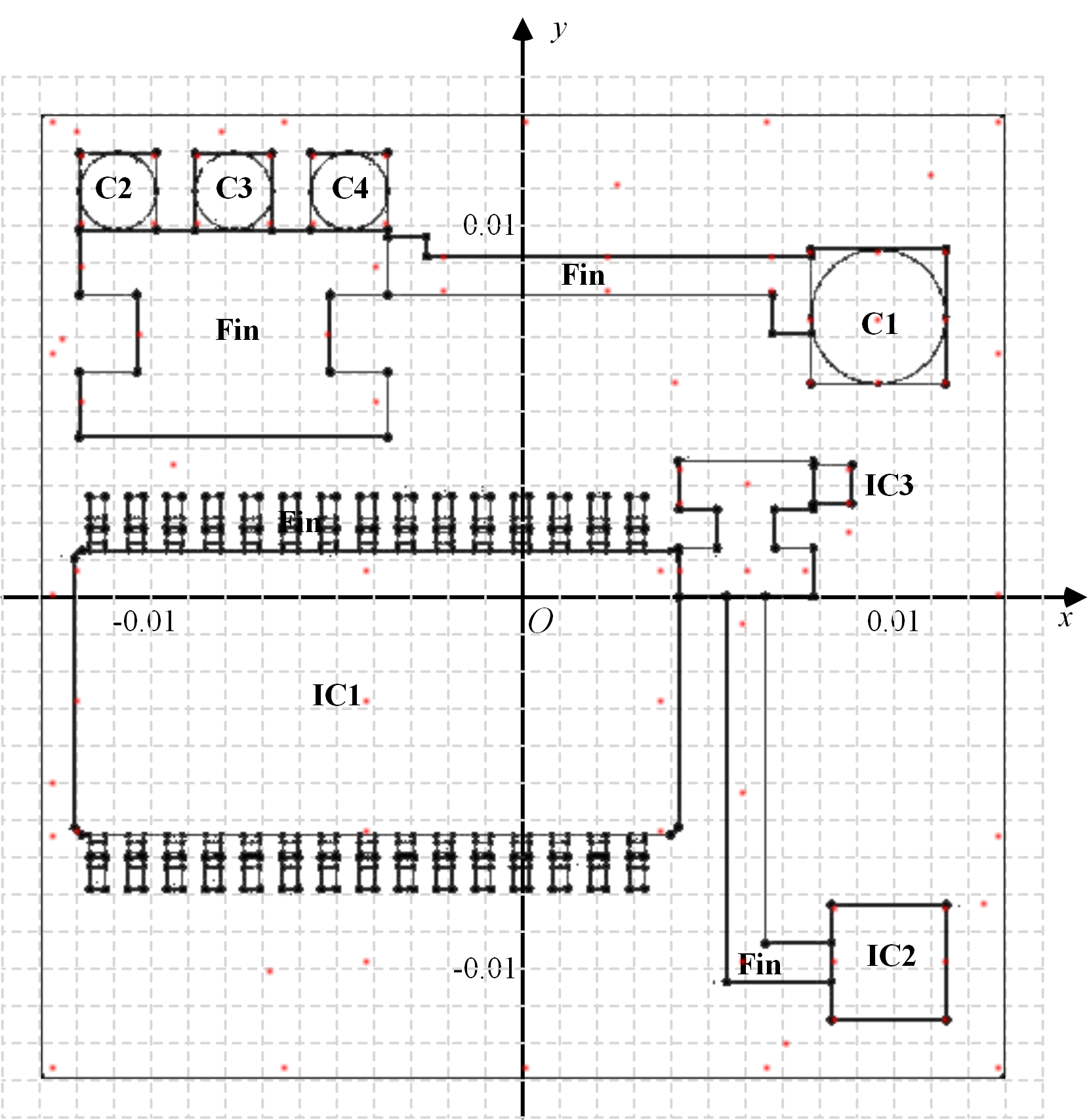}}
  \caption{3D heat-source system with four capacitances (e.g. C1, C2, C3, C4) and three chips (e.g. IC1, IC2, IC3) which are made of silicons and fins for heat dissipation through air convection. The red points in 2-D display mean the monitoring points.}
\label{fig:real_example}
\end{figure}

\begin{figure}
\centering
\subfigure[Sample 1 (MAE = 1.7143)]{\label{fig:t1a}\includegraphics[width=0.95\linewidth]{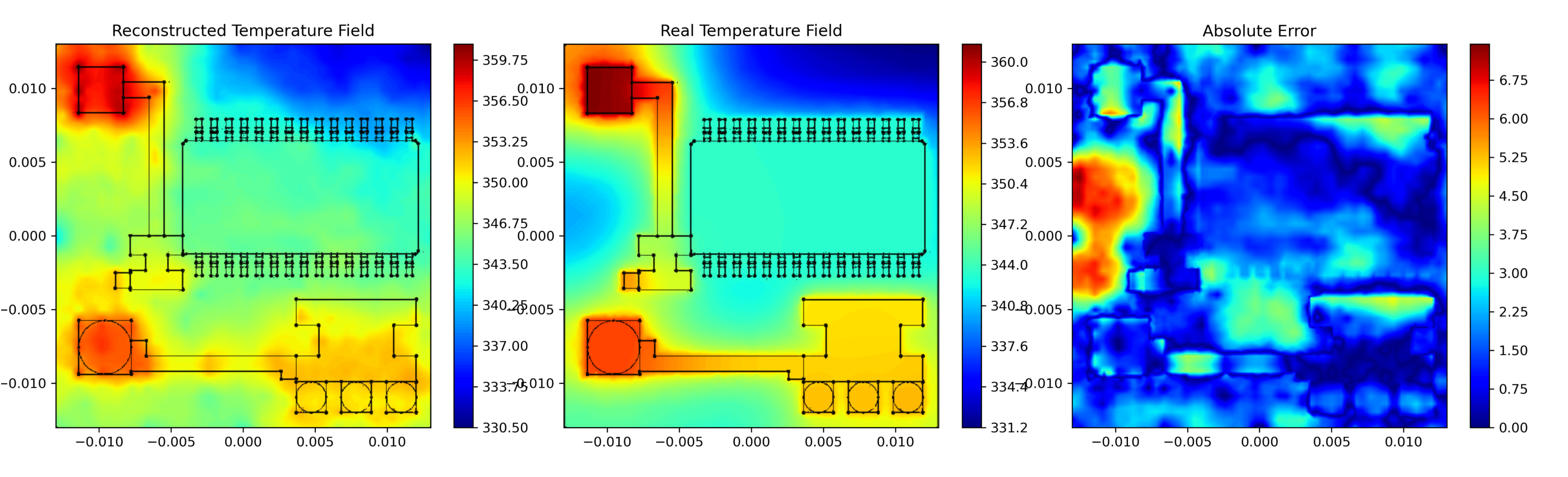}}
\subfigure[Sample 2 (MAE = 1.7668)]{\label{fig:v2b}\includegraphics[width=0.95\linewidth]{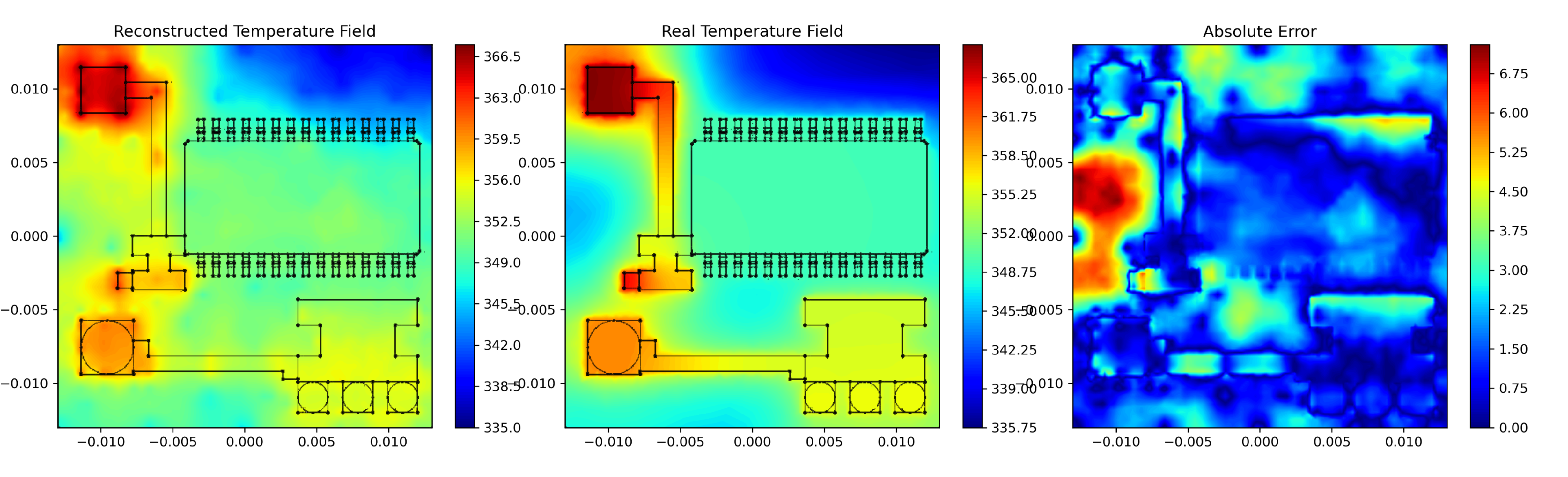}}
\subfigure[Sample 3 (MAE = 1.5251)]{\label{fig:v3b}\includegraphics[width=0.95\linewidth]{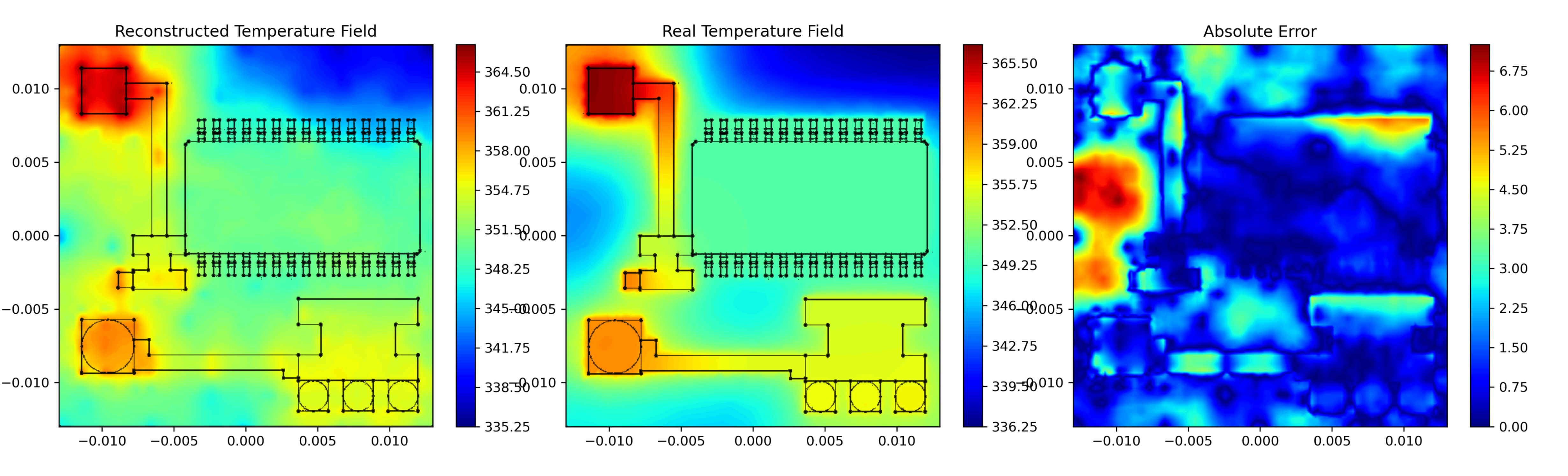}}
  \caption{Samples of temperature field reconstruction of 3D heat source system.}
\label{fig:real_results}
\end{figure}

Given a 3D heat-source system with four capacitances and three chips which are made of silicons. 
As Fig. \ref{fig:real_results} shows, the system is cooled with fins which are made of cuprum through air convection.
The layout information of the system can be seen in Fig. \ref{fig:voc}.
The powers of the chips in system, namely IC1, IC2, and IC3, and the capacitance C1, vary from 0.1w to 0.3w.
The powers of other capacitances, e.g. C2, C3, C4, vary from 0.1w to 0.5w. 

For convenience but without loss of generality, the temperature field is processed as the 2D dimensional plane. Therefore, the proposed method can also be used for the reconstruction task of the system.
As Fig. \ref{fig:voc} shows, 88 monitoring points are selected for the reconstruction task.
Since the generation the temperature of the system is time-consuming, this work generates 1600 samples with different working conditions to validate the effectiveness of the proposed method. Among these samples, 960 and 240 are used for training and validation separately, and the remainder 400 are for testing.

In the implementation, the hyperparameter $\alpha$ is set to 0 due to the different ways of heat dissipation. The system is also discreted to $200\times 200$ grids. 
The MAE, M-CAE, CMAE, and BMAE can be 2.12K, 11.69K, 1.56K, 2.30K {separately}.
Fig. \ref{fig:real_results} presents samples of the temperature reconstruction through the proposed method. From these samples, we can find that the proposed method can well reconstruct the temperature field of the complex system. Specifically, the high and low temperature area of the system can be well predicted.

It should be noted that under more training samples, the reconstruction performance would be significantly improved.
Furthermore, the data would be shared with the community \footnote{{https://pan.baidu.com/s/18wYsu9Sax0inqZeMfg5hJA?\_at\_=1667486387244\#list/path=\%2F, password: real}.}.

{
{\bf Performance with noise samples}  We also discuss the performance of the proposed method over noising observations as input samples. To study the effect of additive noise in observations, this work generates the noisy observation samples as follows: 
\begin{equation}
f_{noise}=f+\eta \delta(0,1) f,
\end{equation}
where $f$ denotes the monitoring matrix as subsection \ref{subsec:obser}, $f_{noise}$ defines the noised samples, $\eta$ determines the noise level, and $\delta(0,1)$ represents a random value sampled from Gaussian distribution with mean and standard deviation of 0 and 1 separately.

In the experiments, the noise is with the noise level $\eta$ of $1e^{-2}$ and enforced over Data D. Samples of the added noise are showed in Fig. \ref{fig:noise}. SetNet-SegNet is also used as the deep reversible regression model. 40000 noised samples are generated for training and 10000 samples are used for testing.

\begin{figure}
\centering
\subfigure[]{\label{fig:n1a}\includegraphics[width=0.48\linewidth]{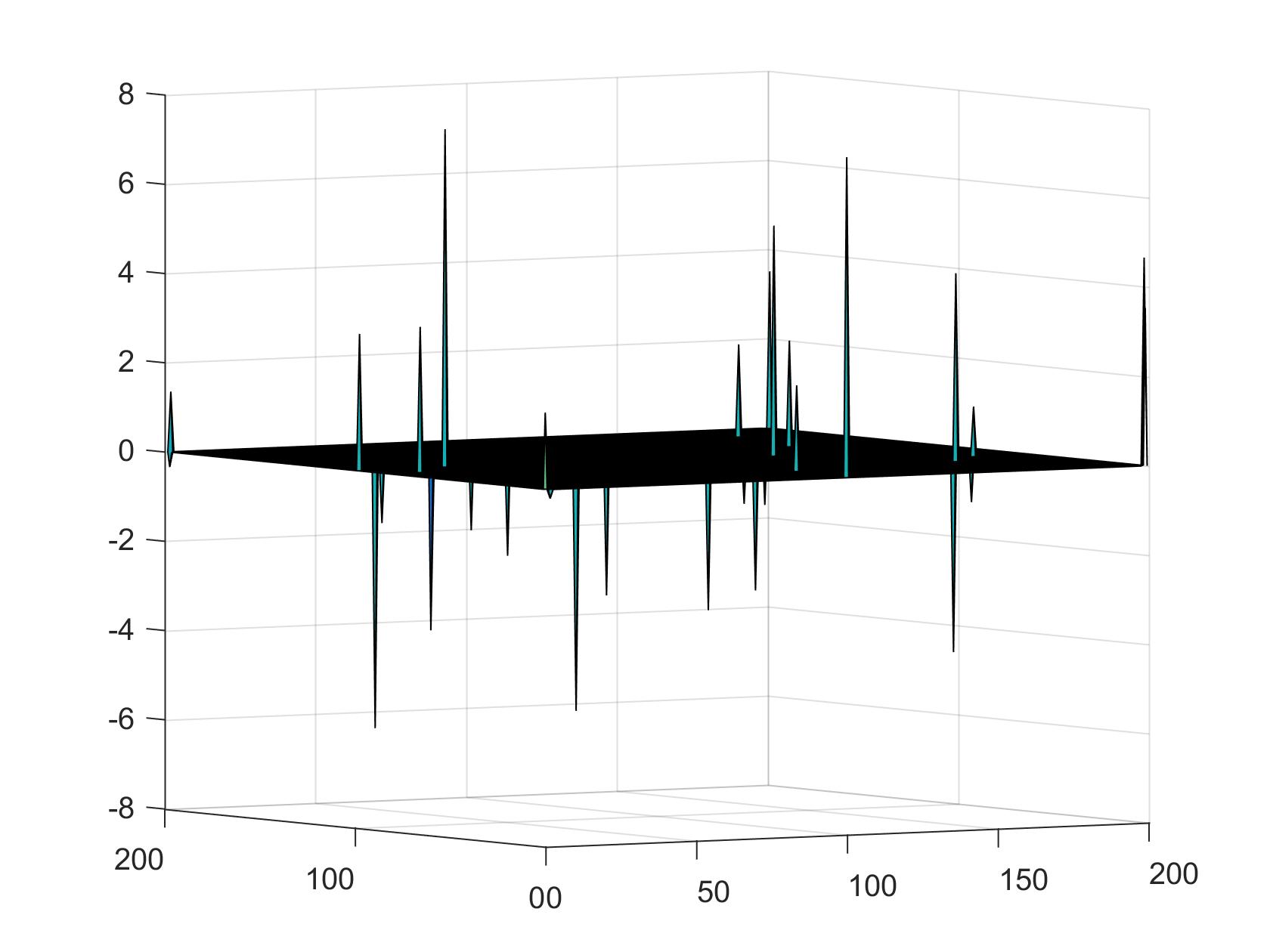}}
\subfigure[]{\label{fig:n2b}\includegraphics[width=0.48\linewidth]{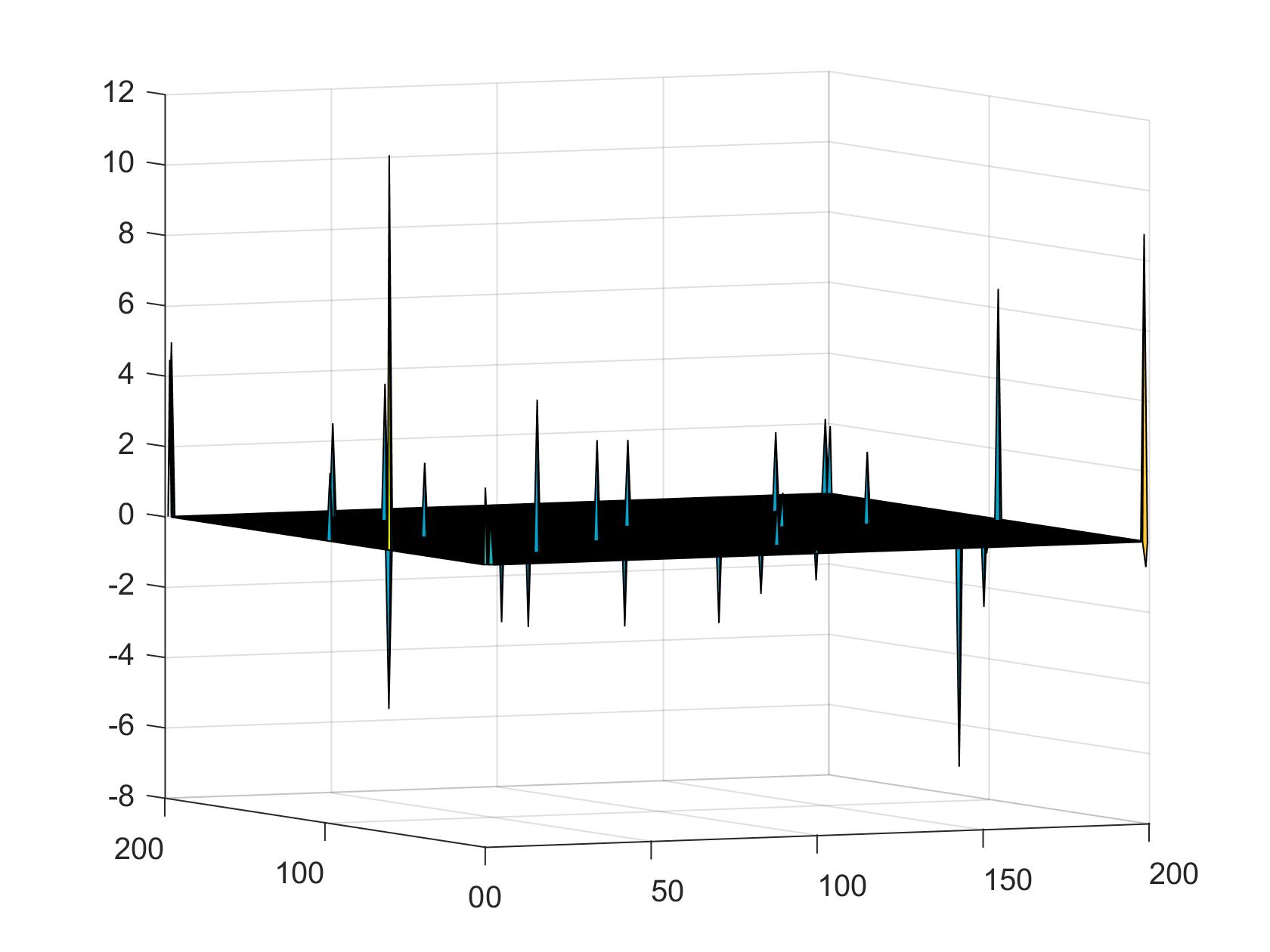}}
  \caption{Samples of noise added over the Data D.}
\label{fig:noise}
\end{figure}

From Fig. \ref{fig:noise}, we can find that about 10K deviation exist over monitoring points in each sample.
While through our proposed method, the MAE, M-CAE, CMAE, and BMAE can be 1.5324, 6.8861, 1.6478, 1.6502, respectively. Besides, Fig. \ref{fig:noise_results} shows the reconstruction examples. From the results, we can find that the noise samples have negative effect on the training process of the proposed method. However, our method behaves robustness over noise input and can work over task with such noise influence.

\begin{figure}
\centering
\subfigure[Sample 1]{\label{fig:nra}\includegraphics[width=0.95\linewidth]{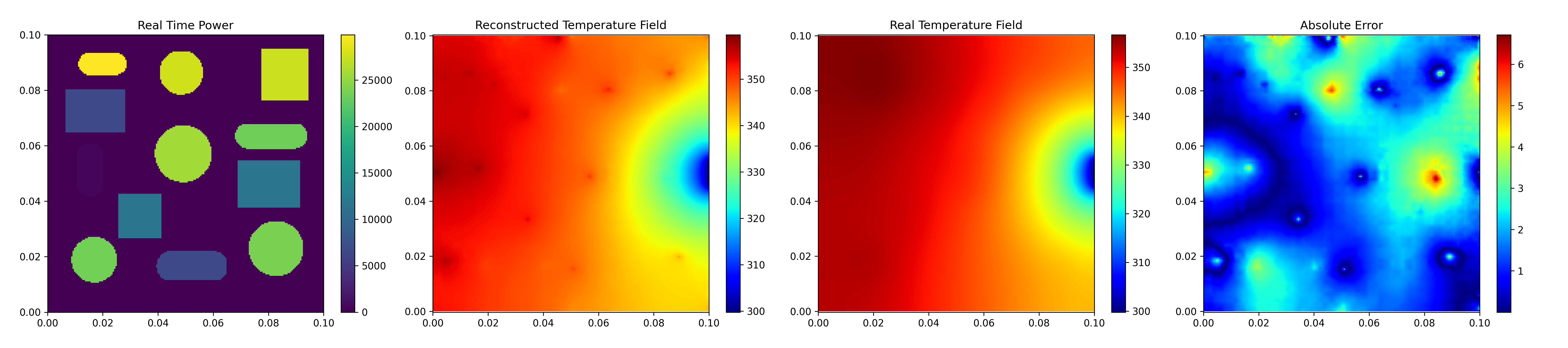}}
\subfigure[Sample 2]{\label{fig:nrb}\includegraphics[width=0.95\linewidth]{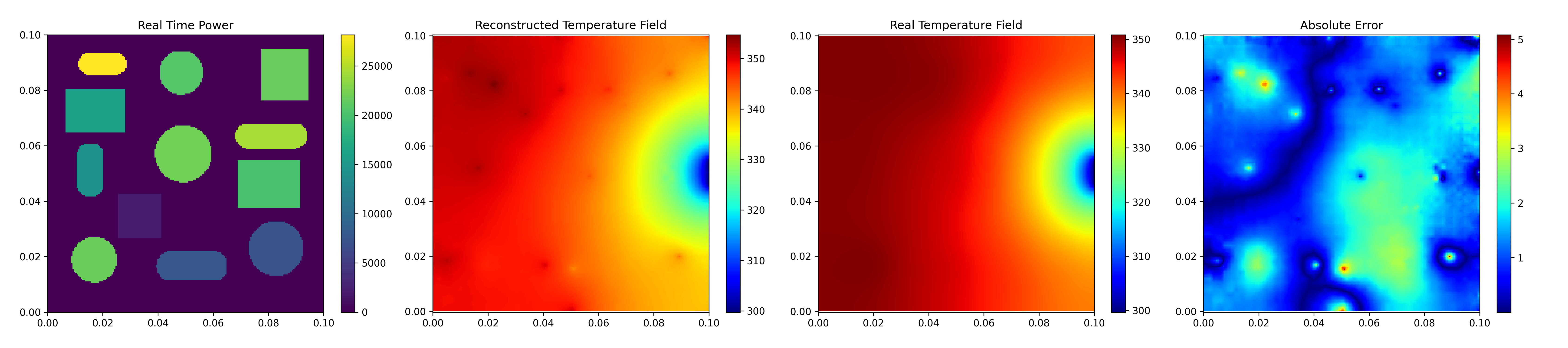}}
\subfigure[Sample 3]{\label{fig:nrc}\includegraphics[width=0.95\linewidth]{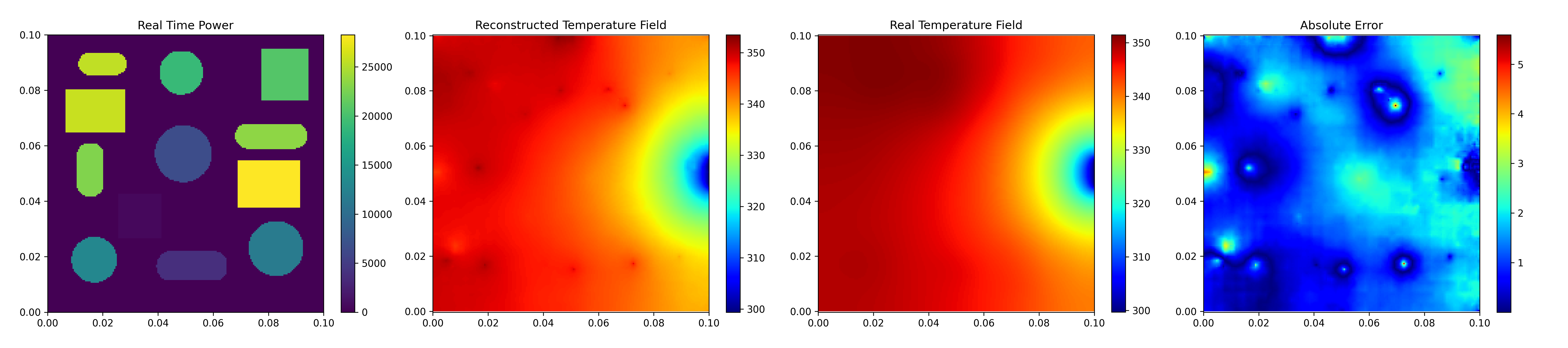}}
  \caption{Examples of reconstruction results through the proposed method over noise enforced Data D.}
\label{fig:noise_results}
\end{figure}

}

{\bf Monitoring points selective strategies} Generally, the more monitoring points the system has, the higher recontruction performance the model would provide. The number of monitoring points would be selected through the tradeoff between the reconstruction performance and the monitoring cost. However, given a certain number of monitoring points, points with more diversity and less redundant would provide more temperature information for the model which can result in a better reconstruction performance. Traditionally, clustering, correlation analysis, and diversity methods \cite{r4} are used for the selection of monitoring points.
{Besides, given a monitoring points arrangement, and Non-Maximum supression (NMS), coefficient matrix condition number\cite{r5},} and other redundancy analysis methods can be used to remove the most redundant monitoring points.

{\bf Structure of deep learning models} This work proposes a deep reversible regression model for the task. Subsection \ref{subsec:model_comparasion} presents the reconstruction performance with different base models. 
Generally, the M-CAE and MAE are the metrics we care most. Therefore, the UNet-UNet is the most recommended for the TFR-HSS task. 
However, it should also be noted that all these base models are designed for computer vision tasks. Other architectures or modules which can better represent the temperature field are suggested to investigate and be applied in the proposed framework.

{\bf Expanding applications}
This work mainly considers the thermal conduction, which is also one of the main type of heat transfer, to construct the physics-informed reconstruction loss. As for expanding applications with other energy sources such as radiant heat, our method can also be applied. Among the four physical {losses} we construct, the Point loss, BC loss, and TV loss can be applied directly for these expanding applications. Since the heat transfer of radiant heat follows other physical correlation, one should construct such physical-based loss instead of laplace loss for these applications.
Futhermore, our method can also be applied {in three-dimensional (3-D) systems.} Generally, what we care most of the 3-D systems is the temperature field of surface plane of the 3-D systems, which can be looked as the {two-dimensional (2-D)} reconstruction task. Therefore, our method can also work for such systems.
Overall, this work {provides a novel method} for using physical laws to reconstruct temperature field and others can construct the corresponding physical-informed losses based on the physical charasteristics of their systems.



\section{Conclusions}
\label{sec:conclusions}

{In this work, we develop a novel temperature field reconstruction method which joints the deep reversible regression model and physics-informed unsupervised learning for the TFR-HSS task.
First, a deep reversible regression model with two successive encoder-decoder processes and one flip operation is proposed as the deep surrogate model for the TFR-HSS task.
Experiments have showed that the proposed deep surrogate model can better reconstruct the temperature field than  vanilla deep  regression models, especially over the area near the boundaries.  Specifically, UNet-UNet is most recommended as base model for the task. 
Then, we develop a novel physics-informed reconstruction loss for TFR-HSS task, which can train the deep model unsupervisedly and alleviate the dependence of large amounts of labelled samples in the training process. The PIRL consists of the point loss, the bc loss, the laplace loss and the TV loss.
 Among these losses, the point loss, the bc loss, the TV loss are more general ones which can be applied in other physical system, such as systems with radiant sources.
 The experimental results demonstrate that our method can provide an accurate reconstructed field with the MAE less than 0.5K using about 5.2ms. 
Besides, compared with commonly used gaussian regression method, other machine learning methods, and the vanilla deep regression models, the proposed method can provide better reconstruction performance with the more representative deep architectures and embedded physical characteristics through physics-informed reconstruction loss.

 However, our method cannot be applied directly to the irregular system due to the types of input and output of the model.
Besides, even though the training process for the deep reversible model do not rely on large amounts of labelled samples with temperature field, it also requires massive unlabelled samples with monitoring information while collecting these monitoring information is also time-consuming.
These limitations point out the future directions of this work.
Inverstigating affine transformation, manifold embedding and other embedding-based methods to apply our method for irregular complex system would be can interesting topic.
As future work, it would be interesting to do researches on online learning to decrease the number of offline data for training process.  
}

\appendix
\section{Overview}

This document includes supplementary material to ``Joint Deep Reversible Regression Model and
Physics-Informed Unsupervised Learning for Temperature Field Reconstruction''. Included are other more results.


\section{More Results}

\subsection{Details of the Datasets}

{
The domain of the systems is denoted as $0.1m\times 0.1m$ board. For convenience but without loss of generality, the heat sinks with $0.01m$ are used as the boundary conditions for heat dissipation. Specifically, for Data A, Data B, Data C, Data D, the heat sinks exist over the center of the upside, downside, leftside, rightside of the boundary, respectively. 
For each dataset, the power of each heat source ranges from 0-30000 $W/m^2$. The layout information and characteristics of heat source components of each dataset are shown in Table \ref{table:componentA}, \ref{table:componentC} and \ref{table:componentD}.
The detailed configuration file of these datasets for our data generator \footnote{\url{https://github.com/shendu-sw/recon-data-generator}} would be released soon.  
Interested researchers could generate these datasets and conduct a deep study of our work.
}
\begin{table}
\centering
{  

    \caption{The layout information and characteristics of heat source components of Data A and Data B.}
    \label{table:componentA}
  \begin{tabular}{|c| c c c c|}
    \hline
  No. & Shape &   Length(m) & Width(m) & Location \\
	\hline
     {\bf c1} & rectangle &   0.01 & 0.01 & (0.0065,0.079) \\
     {\bf c2} & rectangle &   0.013 & 0.013 & (0.0195,0.025) \\
     {\bf c3} & rectangle &   0.013 & 0.013 & (0.0271,0.089) \\
     {\bf c4} & rectangle &   0.02 & 0.02 & (0.0397,0.0552) \\
     {\bf c5} & rectangle &   0.015 & 0.015 & (0.0548,0.0144) \\
     {\bf c6} & rectangle &   0.01 & 0.02 & (0.0533,0.08) \\
     {\bf c7} & rectangle &   0.02 & 0.01 & (0.0638,0.0336) \\
     {\bf c8} & rectangle &   0.016 & 0.016 & (0.079,0.0659) \\
     {\bf c9} & rectangle &   0.011 & 0.011 & (0.078,0.0139)  \\
    {\bf c10} & rectangle &   0.012 & 0.012 & (0.0805,0.0921) \\
    \hline
\end{tabular}
}
\end{table}

\begin{table}
\centering
{  

    \caption{The layout information and characteristics of heat source components of Data C.}
    \label{table:componentC}
  \begin{tabular}{| c  c| c c c c|}
    \hline
  \multicolumn{2}{|c|}{{\bf No.}} & Shape &   Length(m) & Width(m) & Location \\
	\hline
     \multicolumn{2}{|c|}{{\bf c1}}& rectangle &   0.011 & 0.02 & (0.006,0.011) \\
\hline
     \multicolumn{2}{|c|}{{\bf c2}} & rectangle &   0.008 & 0.016 & (0.012,0.072) \\
\hline
     \multicolumn{2}{|c|}{{\bf c3}} & rectangle &   0.013 & 0.013 & (0.016,0.042) \\
\hline
     \multirow{3}[0]{*}{{\bf c4}} & c41 & rectangle &   0.01 & 0.01 & (0.0297,0.0191) \\
 & c42& rectangle &   0.005 & 0.01 & (0.0271,0.0299) \\
 & c43& rectangle &   0.005 & 0.01 & (0.0321,0.0299) \\
\hline
     \multirow{2}[0]{*}{{\bf c5}} &c51 & rectangle &   0.01 & 0.005 & (0.0397,0.07) \\
 & c52& rectangle &   0.01 & 0.015 & (0.0397,0.08) \\
\hline
     \multirow{3}[0]{*}{{\bf c6}} &c61 & rectangle &   0.01 & 0.005 & (0.0447,0.0425) \\
 &c62 & rectangle &   0.01 & 0.015 & (0.0447,0.0526) \\
 & c63& rectangle &   0.01 & 0.02 & (0.0548,0.0501) \\
\hline
     \multirow{2}[0]{*}{{\bf c7}} & c71& rectangle &   0.015 & 0.01 & (0.0599,0.0834) \\
 & c72& rectangle &   0.005 & 0.01 & (0.0699,0.0834) \\
\hline
     \multirow{3}[0]{*}{{\bf c8}} & c81& rectangle &   0.01 & 0.02 & (0.0649,0.0249) \\
 &c82 & rectangle &   0.01 & 0.01 & (0.0749,0.0199) \\
 & c83& rectangle &   0.01 & 0.01 & (0.0749,0.0299) \\
\hline
     \multicolumn{2}{|c|}{{\bf c9}} & rectangle &   0.023 & 0.02 & (0.072,0.062)  \\
\hline
    \multicolumn{2}{|c|}{{\bf c10}} & rectangle &   0.01 & 0.03 & (0.089,0.027) \\
    \hline
\end{tabular}
}
\end{table}

\begin{table}
\centering
{  

    \caption{The layout information and characteristics of heat source components of Data D.}
    \label{table:componentD}
  \begin{tabular}{|c| c c c c|}
    \hline
  No. & Shape &   Length(m) & Width(m) & Location \\
	\hline
     {\bf c1} & rectangle &   0.0172 & 0.0186 & (0.0856,0.0853) \\
     {\bf c2} & rectangle &   0.0227 & 0.0173 & (0.0797,0.046) \\
     {\bf c3} & rectangle &   0.0154 & 0.016 & (0.0333,0.0343) \\
     {\bf c4} & rectangle &   0.0215 & 0.0155 & (0.0172,0.0721) \\
     {\bf c5} & circle &   0.0164 & 0.0164 & (0.0166,0.0186) \\
     {\bf c6} & circle &   0.0158 & 0.0158 & (0.0479,0.0858) \\
     {\bf c7} & circle &   0.0205 & 0.0205 & (0.0487,0.0566) \\
     {\bf c8} & circle &   0.0198 & 0.0198 & (0.0821,0.0226) \\
     {\bf c9} & capsule &   0.01 & 0.0195 & (0.0152,0.0509)  \\
    {\bf c10} & capsule &   0.011 & 0.0258 & (0.0518,0.0164) \\
    {\bf c11} & capsule &   0.026 & 0.0091 & (0.0804,0.0630)  \\
    {\bf c12} & capsule &   0.0174 & 0.008 & (0.0196,0.0889) \\
    \hline
\end{tabular}
}
\end{table}

\subsection{Performance with Different Monitoring Points on Each Component}

\begin{figure}[t]
\centering
\subfigure[Data A1]{\label{fig:1e}\includegraphics[width=0.48\linewidth]{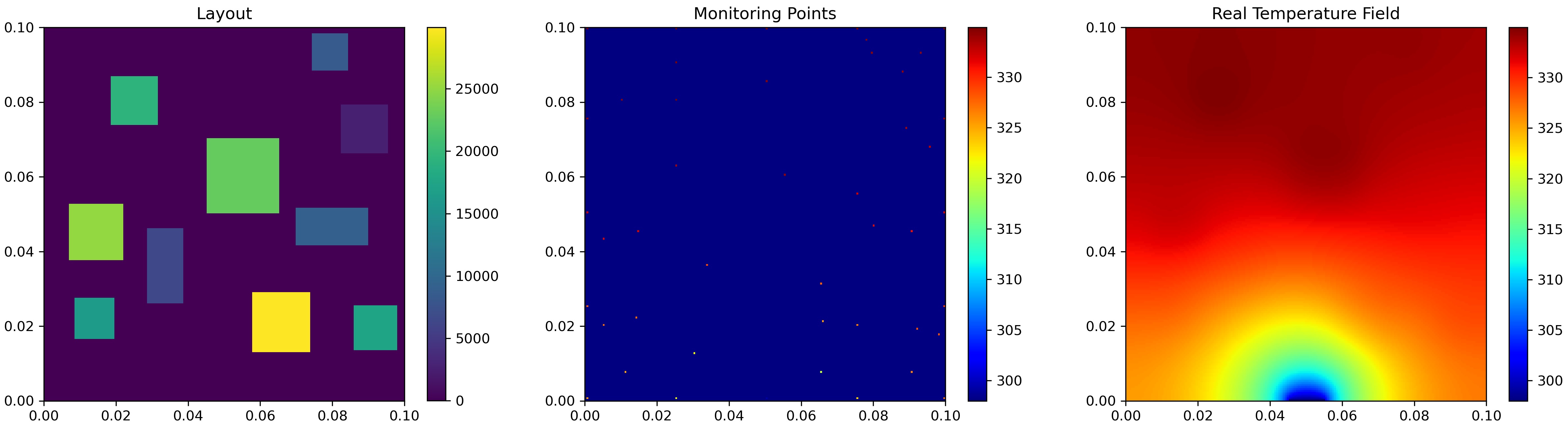}}
 \subfigure[Data B1]{\label{fig:1f}\includegraphics[width=0.48\linewidth]{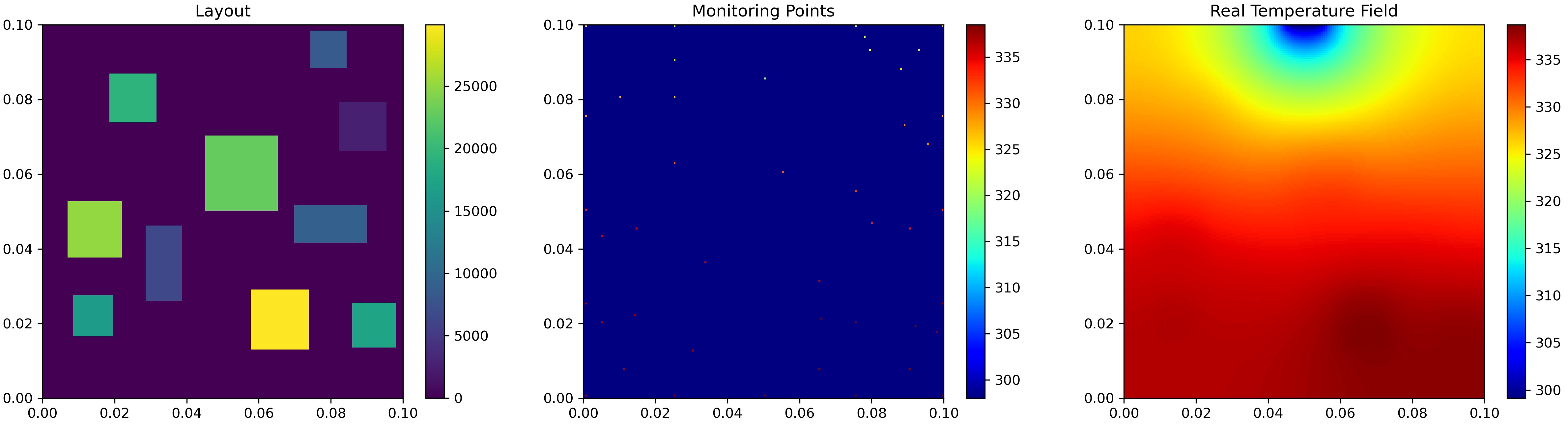}}
 \subfigure[Data C1]{\label{fig:1g}\includegraphics[width=0.48\linewidth]{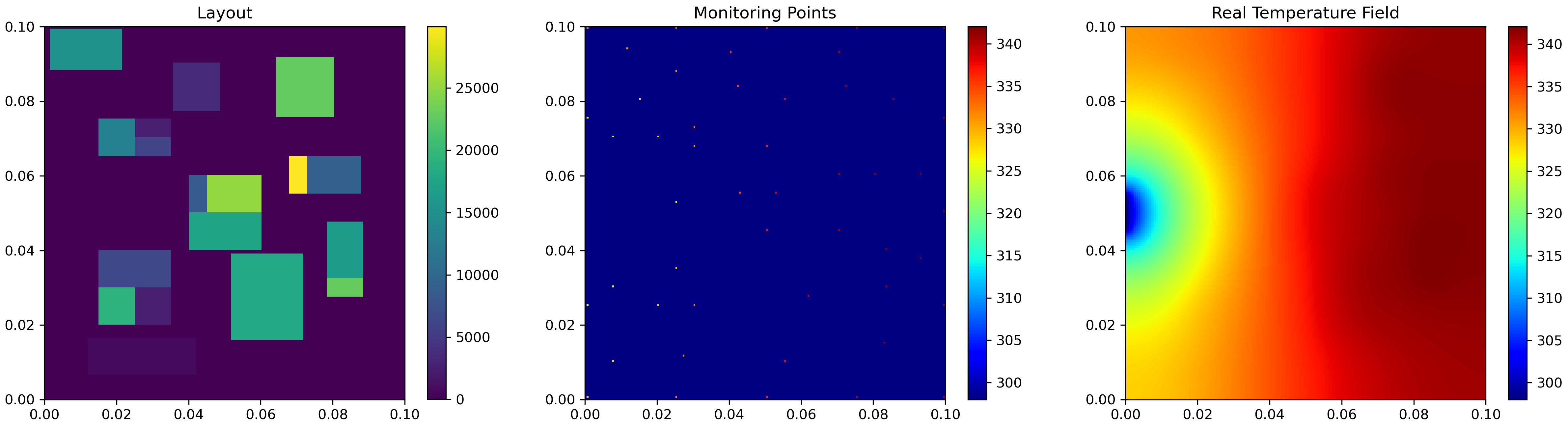}}
 \subfigure[Data D1]{\label{fig:1h}\includegraphics[width=0.48\linewidth]{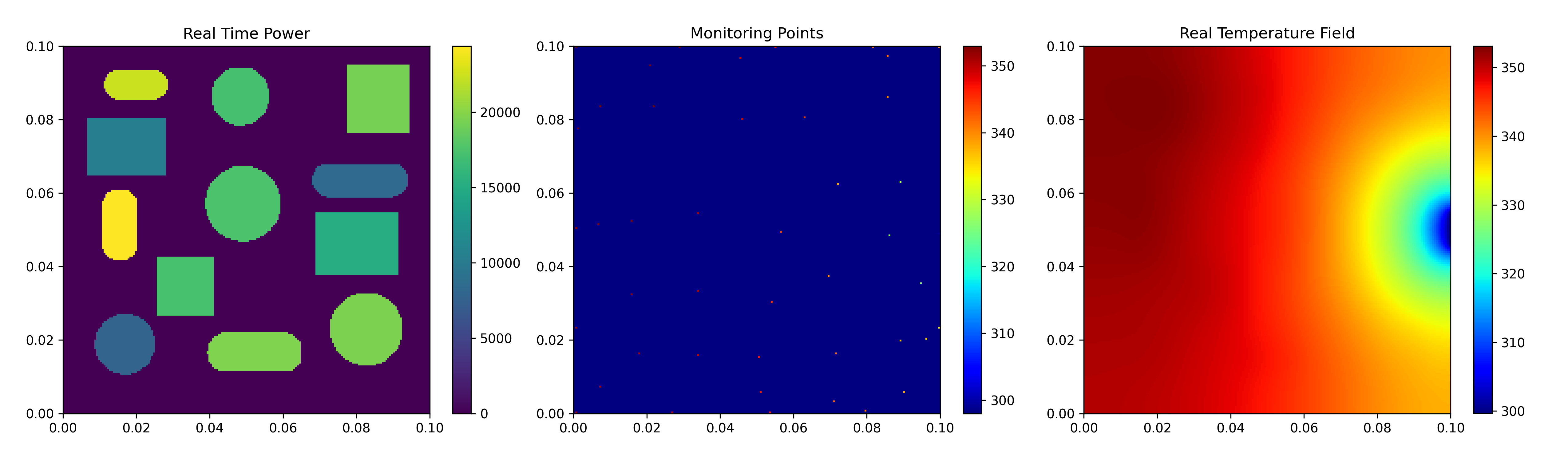}}
   \caption{Different simulation analysis datasets with less monitoring points over the heat sources in the system. Corresponding to Data A, B, C, D, Data A1, B1, C1, D1 place only one monitoring point over the component in the heat-source system.}
\label{fig:data1}
\end{figure}

For further comparison, as Fig. \ref{fig:data1} shows, we set Data A1, B1, C1, D1 corresponding to Data A, B, C, D, respectively.  
Data A1, B1, C1, D1 have the same heat sources layout as Data A, B, C, D, respectively but with different number of monitoring points. For Data A1, B1, C1, D1, we set one, two, three monitoring points for general heat sources, heat sources with two parts and with three parts, respectively.
The power of the heat sources in Data A1, B1, C1, D1 is distributed as Data A, B, C, D separately.

\begin{figure}
\centering
\subfigure[Data A]{\label{fig:1a}\includegraphics[width=0.48\linewidth]{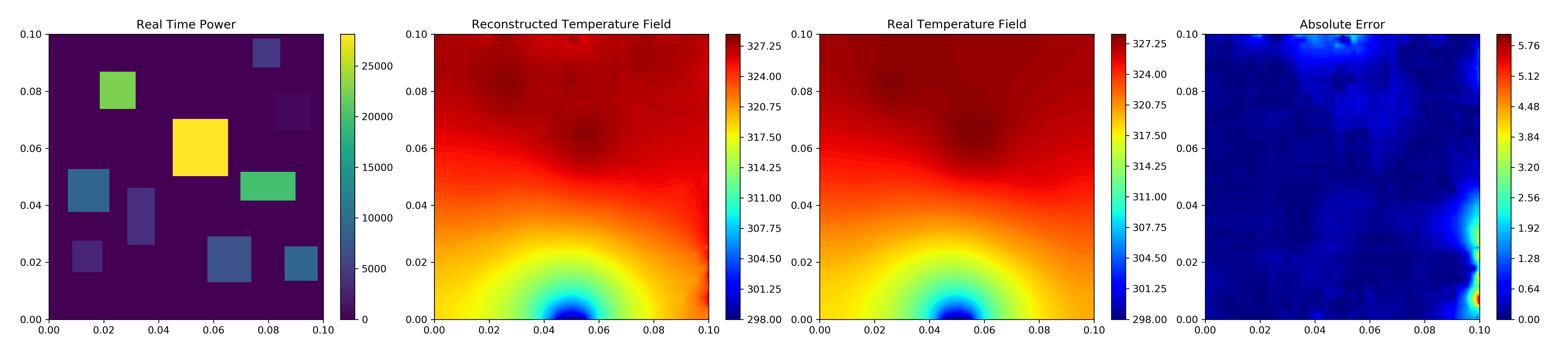}}
\subfigure[Data B]{\label{fig:1b}\includegraphics[width=0.48\linewidth]{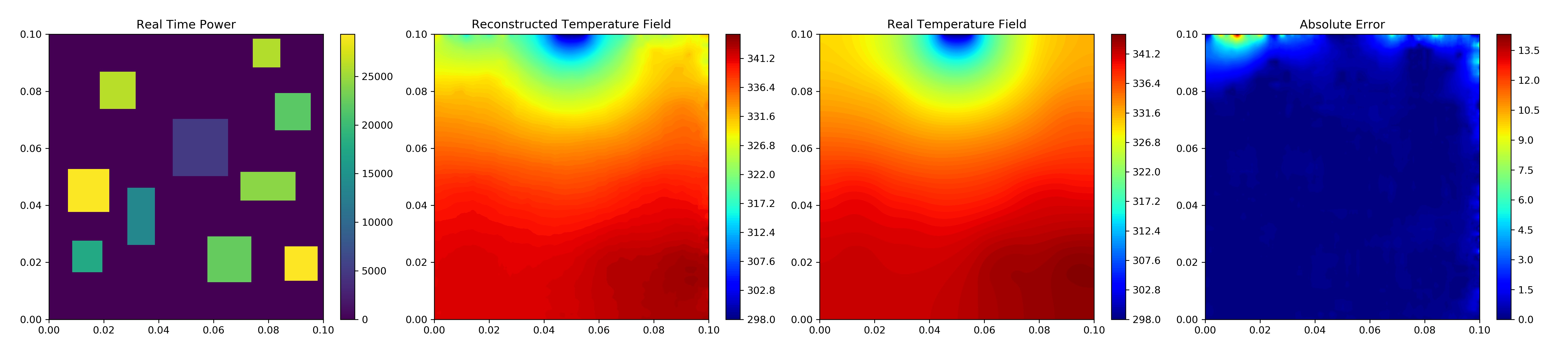}}
\subfigure[Data C]{\label{fig:1c}\includegraphics[width=0.48\linewidth]{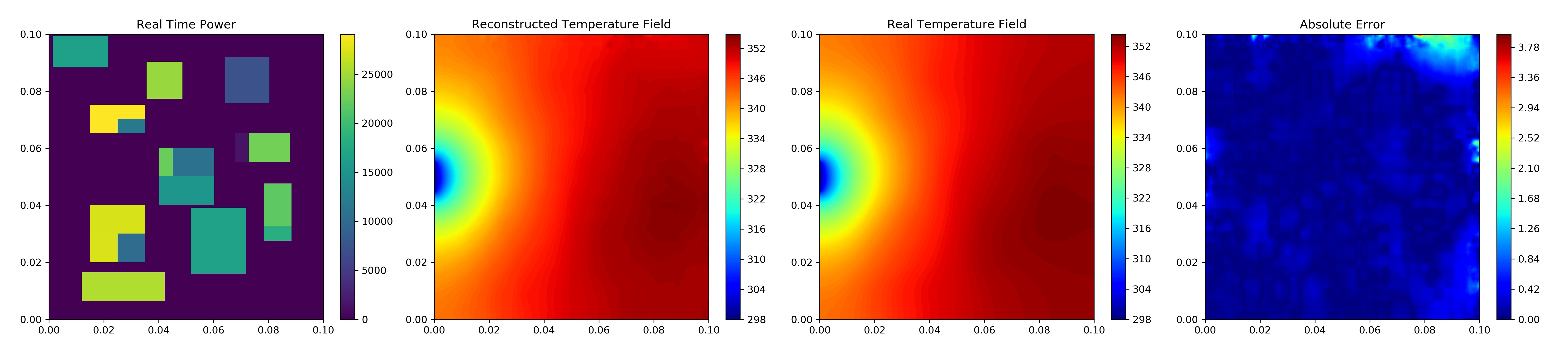}}
\subfigure[Data D]{\label{fig:1d}\includegraphics[width=0.48\linewidth]{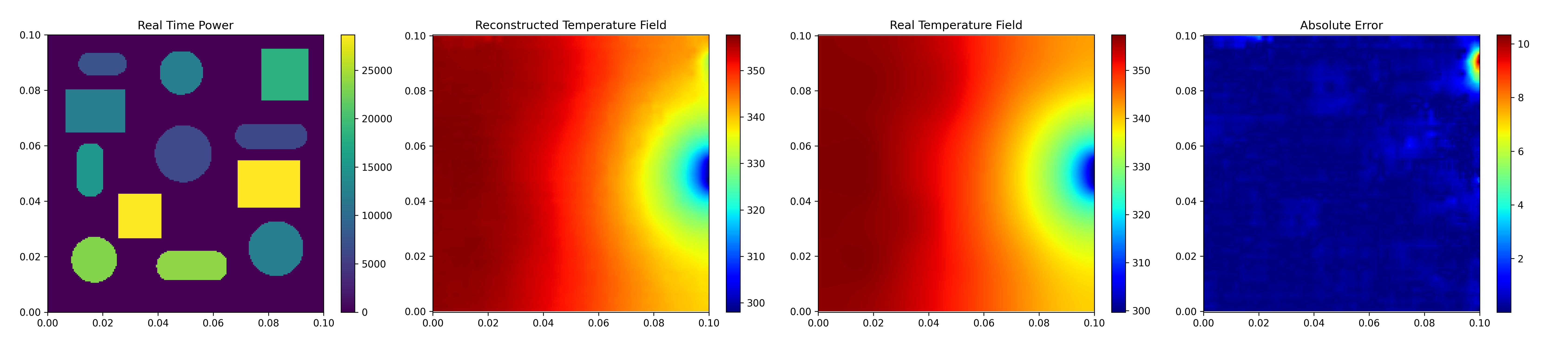}}
  \caption{Examples of proposed method over Data A, B, C and D for TFR-HSS task.}
\label{fig:monitoring_9}
\end{figure}

\begin{table}
\begin{center}
\caption{Number of monitoring points used in this work for Data A1, B1, C1, D1. NB, BC and OC represent near the boundary, between components and on the components, respectively. }
\label{table:general_performance}
\begin{tabular}{| c | c|c| c | c | }
\hline
{\bf Positions}    & {NB} & {BC}   &  {OC} & Total \\
\hline\hline
Data A1 \& B1 & 16 & 18 & $1\times 10$ & 44 \\
Data C1 & 16 & 16 & $1\times 9 + 3\times 3 + 1\times 2$ & 52 \\
Data D1 & 16 & 18 & $1\times 12$ & 52 \\
\hline
\end{tabular}
\end{center}
\label{table:point}
\end{table}

Since the temperature distribution over the heat sources is what we care most, we compare the performance of the performance with different monitoring points, that is we compare the performance between Data A and A1, B and B1, C and C1, D and D1. 

\begin{figure}
\centering
\subfigure[Data A1]{\label{fig:1a}\includegraphics[width=0.48\linewidth]{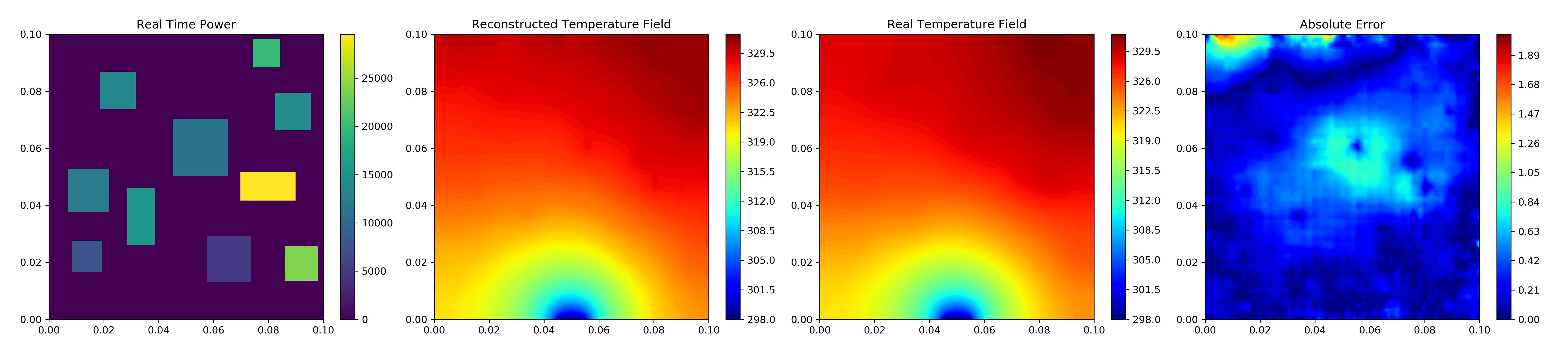}}
\subfigure[Data B1]{\label{fig:1b}\includegraphics[width=0.48\linewidth]{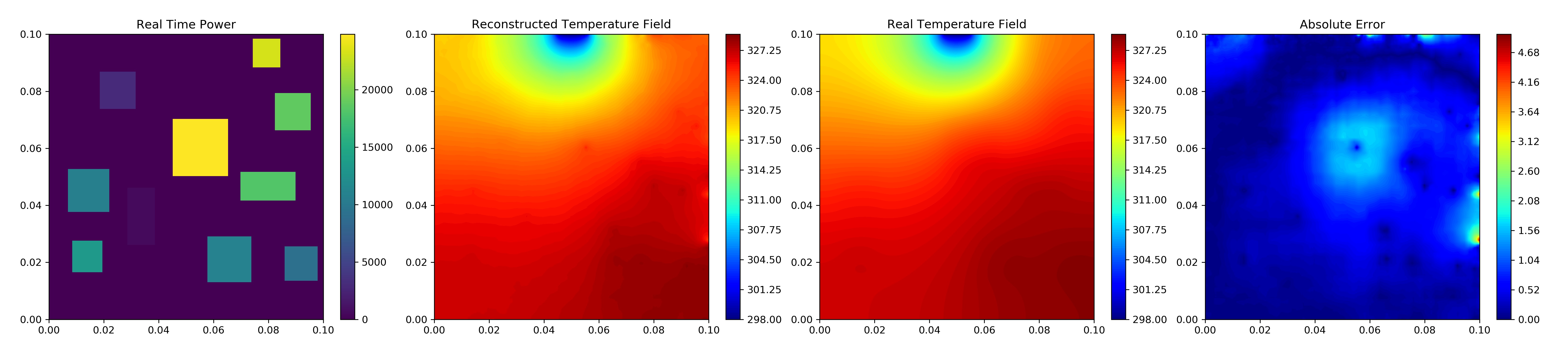}}
\subfigure[Data C1]{\label{fig:1c}\includegraphics[width=0.48\linewidth]{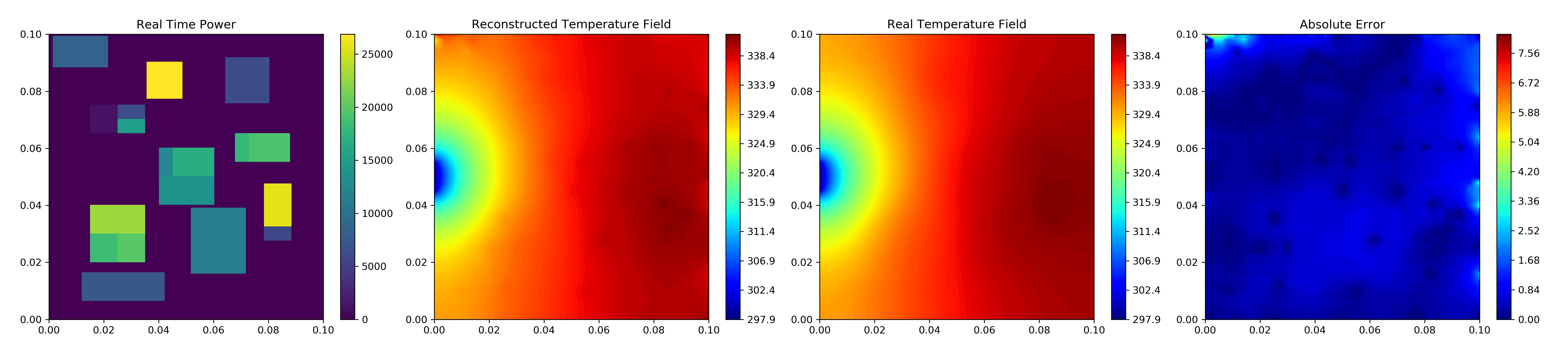}}
\subfigure[Data D1]{\label{fig:1d}\includegraphics[width=0.48\linewidth]{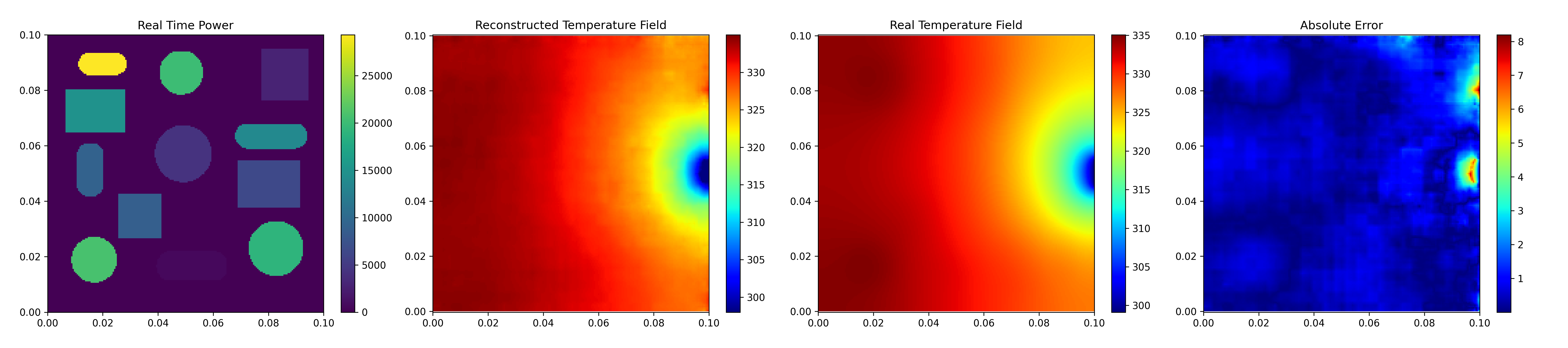}}
  \caption{Examples of proposed method over Data A1, B1, C1 and D1 which have only one monitoring point over each component for TFR-HSS task.}
\label{fig:monitoring_1}
\end{figure}

Fig. \ref{fig:monitoring_9} and \ref{fig:monitoring_1} shows the examples of proposed method over datasets with different monitoring points over the components. Table \ref{table:general_performance} lists the comparison results of the proposed method over these two circumstances. Obviously, the reconstruction performance with more monitoring points over the components is better than that with less ones. The global MAE of the proposed method over Data A, B, C, D is 0.3436K, 0.2655K, 0.1727K, 0.2456K which is better than 0.5371K, 0.5437K, 0.7840K, and 0.8328K over Data A1, B1, C1, D1, respectively. Especially, The CMAE which descripes the reconstruction error over the components which can obtain 0.0883K, 0.1006K, 0.1157K, 0.1902K is better than 0.4777K, 0.6210K, 0.6220K, 0.7743K. The M-CAE also shows that the proposed method can better reconstruct the temperature field with more monitoring points over the components.
Compare Fig. \ref{fig:monitoring_9} and \ref{fig:monitoring_1}, and we can find that under less monitoring points over each component, the error over the monitoring points is obvious lower than the error of other points. Especially, the reconstruction error of the neighboring points of the monitoring points is noticeably higher than that over the monitoring points. While under more monitoring points, there exists no such phenomenon and the reconstruction error is relatively uniform and very small. This means that under more monitoring points, the surrogate model obtains the physical properties of the temperature field.

Since the temperature field usually cannot achieve the expected performance with less monitoring points over components. Therefore, inspecting other methods which can reconstuct the expected temperature field with less monitoring points is  urgent in the futher research. In the following, we will mainly test the performance of the proposed method over Data A, B, C, and D.

\begin{table}
\begin{center}
\caption{General reconstruction performance (K) of proposed method over datasets with different number of monitoring points for TFR-HSS task (32000, 8000 samples for training, and validation, respectively). }
\label{table:general_performance}
\begin{tabular}{| c | c|c| c | c| }
\hline
{\bf Data} &  {MAE} & {M-CAE}   &  {CMAE} & BMAE  \\
\hline\hline
{Data A} &   0.3436 & 1.4340& 0.0883& 1.0483 \\
{Data B} & 0.2655 & 3.6059 & 0.1006 & 0.9003 \\
{Data C} & 0.1727 & 1.7394 & 0.1157 & 0.4560 \\
{Data D} & 0.2456 & 1.7043 & 0.1902 & 0.5751 \\
\hline
{Data A1} & 0.5371 & 2.1985 & 0.4777 & 0.9213 \\
{Data B1} &  0.5437 & 2.5530 & 0.6210 & 0.7909\\
{Data C1} & 0.7840 & 4.2049 & 0.6220 & 1.5538 \\
{Data D1} & 0.8328 & 2.8981 & 0.7743 & 1.2456 \\
\hline
\end{tabular}
\end{center}
\end{table}

\subsection{Performance with Different Surrogate Models}

\begin{figure}[t]
\centering
\includegraphics[width=0.9\linewidth]{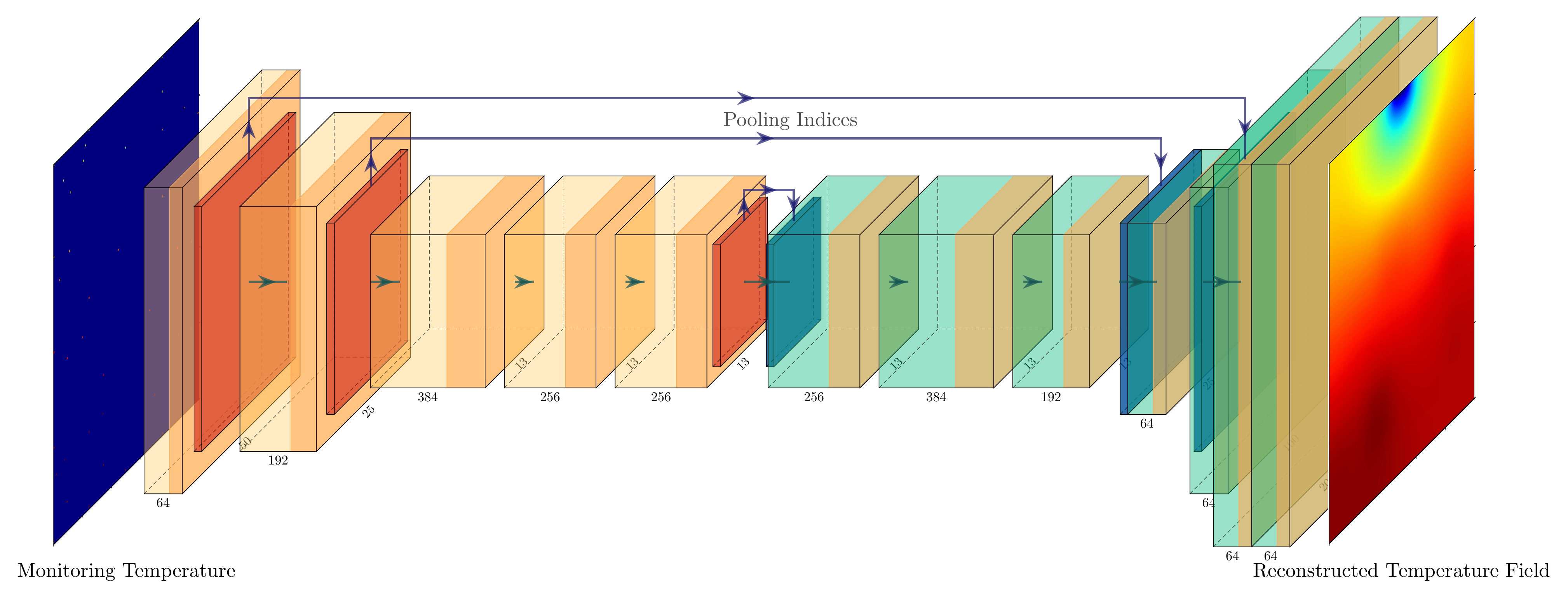}
	\caption{The specific SegNet for TFR-HSS task.}
	\label{fig:segnet}
\end{figure}

\begin{figure}
\centering
\includegraphics[width=0.9\linewidth]{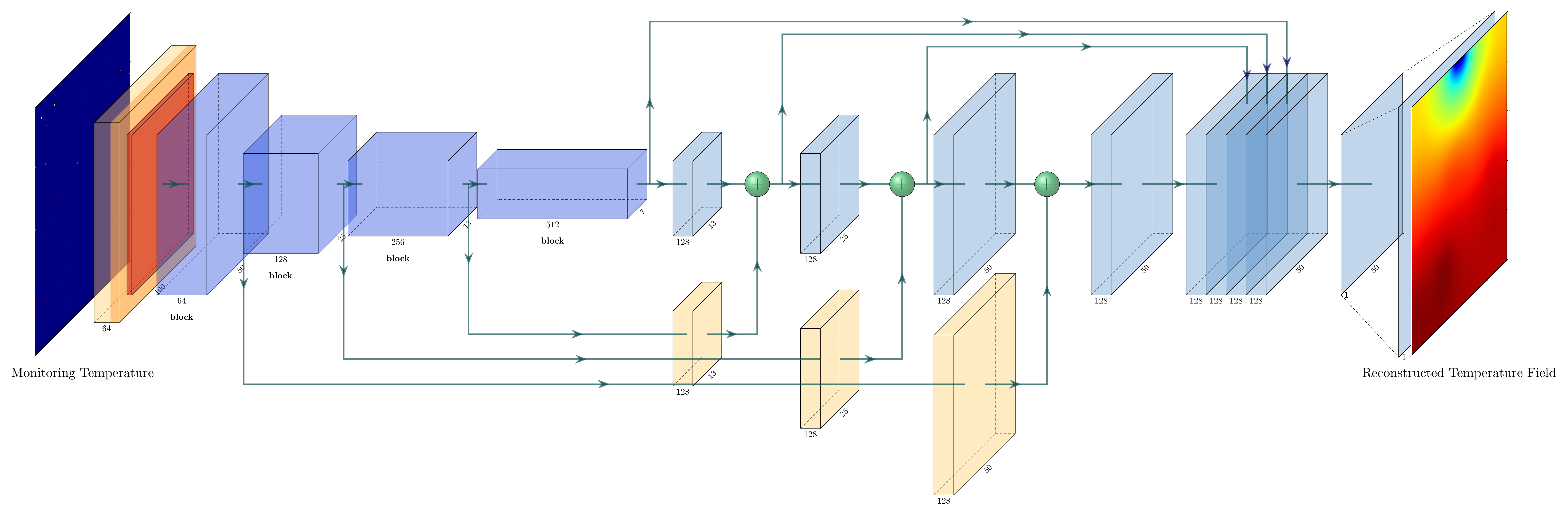}
	\caption{The specific FPN for TFR-HSS task.}
	\label{fig:fpn}
\end{figure}

\begin{figure}[t]
\centering
\includegraphics[width=0.9\linewidth]{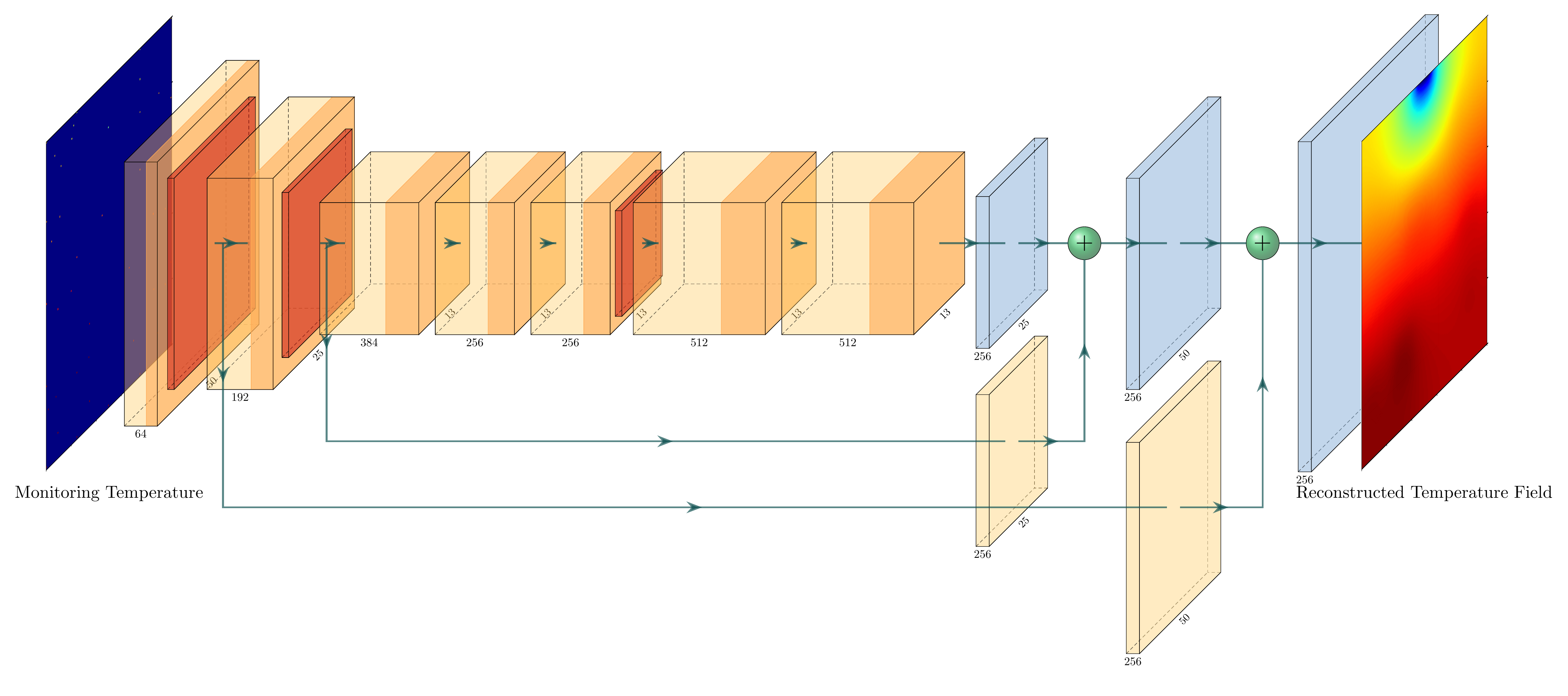}
	\caption{The specific FCN for TFR-HSS task.}
	\label{fig:fcn}
\end{figure}

\begin{figure}[t]
\centering
\includegraphics[width=0.9\linewidth]{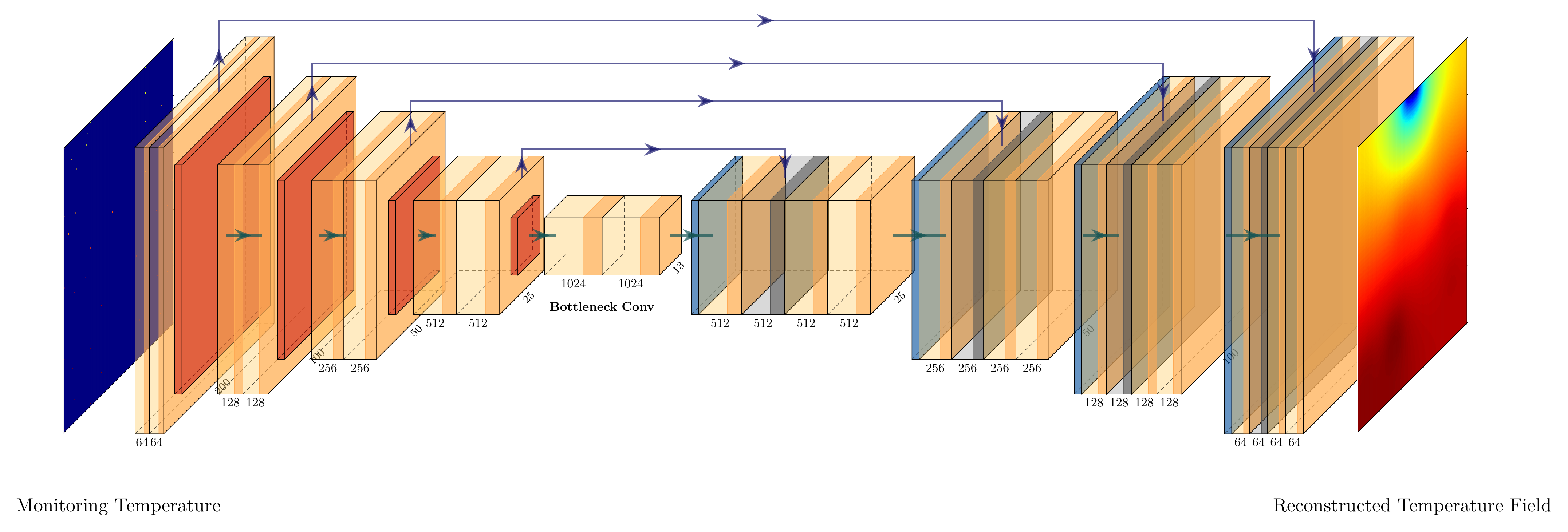}
	\caption{The specific Unet for TFR-HSS task.}
	\label{fig:unet}
\end{figure}

Based on the architecture of SegNet \cite{segnet}, Feature Pyramid Networks (FPN) \cite{fpn}, Fully Convolutional Networks (FCN) \cite{fcn}, and UNet \cite{unet} which are proposed for general computer vision task, such as object detection \cite{objectdetection} and image segmentation \cite{imagesegmentation}, this work designs several base model architectures for current task. { Based on AlexNet \cite{alexnet}, this work designs the specific  FCN  for current TFR-HSS task. Based on VGG \cite{vgg}, this work designs the specific UNet. Based on ResNet \cite{resnet}, this work designs the specific FPN for this task. Detailed architectures can be seen in Fig. \ref{fig:segnet}, \ref{fig:fpn}, \ref{fig:fcn}, and \ref{fig:unet}.}

As showed in the manuscript, the comparison results can be seen in table \ref{table:reversible_regression_model} and the examples of the reconstruction results over Data A by 
posed reversible regression models with different base models can be seen in Fig. \ref{fig:surrogate_models}.

\begin{figure}
\centering
\subfigure[Results by SegNet-SegNet]{\label{fig:1a}\includegraphics[width=0.48\linewidth]{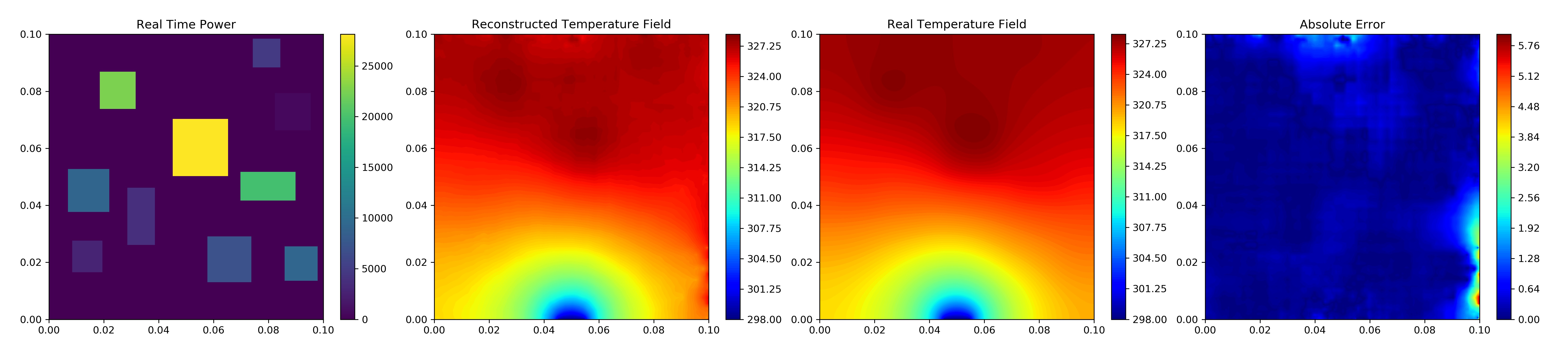}}
\subfigure[Results by FCN-FCN]{\label{fig:1b}\includegraphics[width=0.48\linewidth]{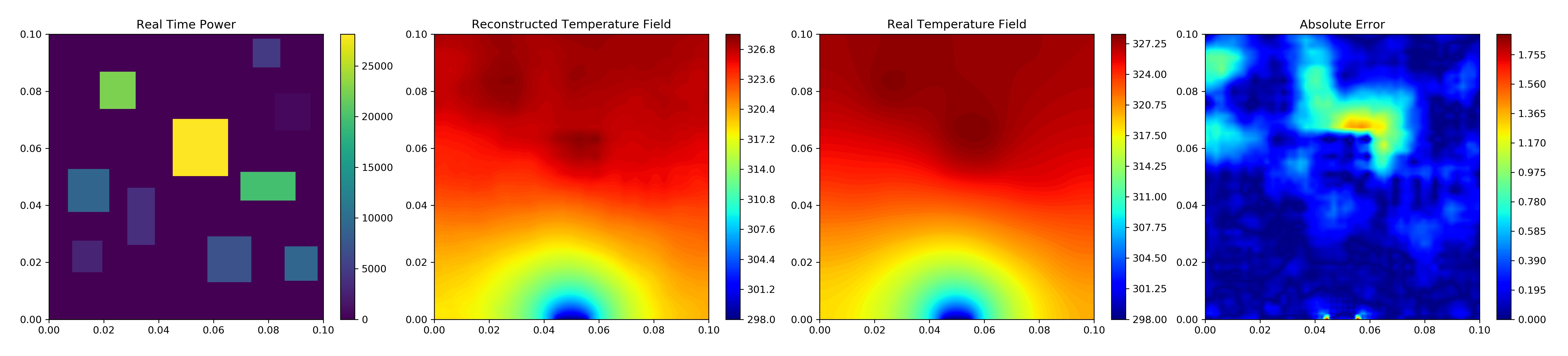}}
\subfigure[Results by FPN-SegNet]{\label{fig:1c}\includegraphics[width=0.48\linewidth]{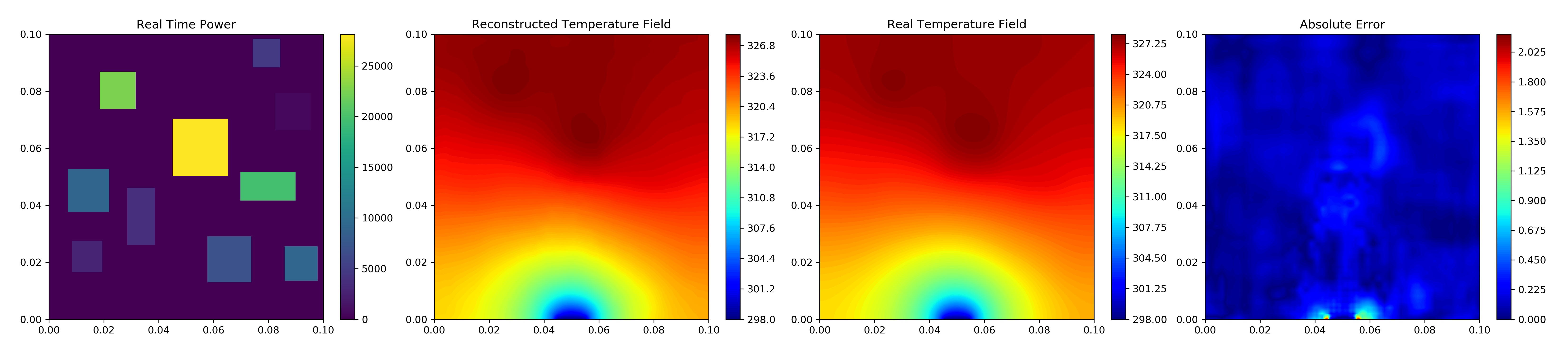}}
\subfigure[Results by Unet-Unet]{\label{fig:1d}\includegraphics[width=0.48\linewidth]{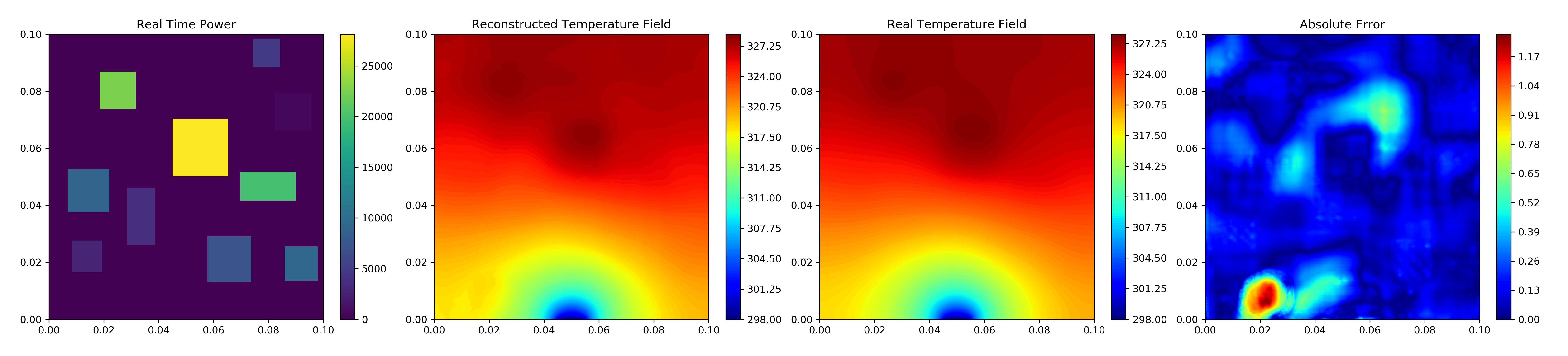}}
  \caption{Examples of the reconstruction results over Data A by proposed reversible regression models with different base models.}
\label{fig:surrogate_models}
\end{figure}

\begin{table}
\begin{center}
\caption{Reconstruction performance (K) of proposed physics-informed deep learning on reversible regression model with different base models for TFR-HSS task. }
\label{table:reversible_regression_model}
\begin{tabular}{| c| c | c|c| c | c | }
\hline
 {\bf Models}&  Metrics& {Data A} & {Data B}   &  {Data C} &  {Data D}  \\
\hline\hline
\multirow{4}{*}{{SegNet-SegNet}} & MAE & 0.3436 & 0.2655  & 0.1727 & 0.2456 \\
 & M-CAE & 1.4340 & 3.6059  & 1.7394 & 1.7043 \\
 & CMAE & 0.0883 & 0.1006  & 0.1157 & 0.1902 \\
& BMAE & 1.0483 & 0.9003  & 0.4560 & 0.5751 \\
\hline
\multirow{4}{*}{{FCN-FCN}} & MAE & 0.1309 & 0.1605  & 0.1051 & 0.1487 \\
 & M-CAE & 0.6310 & 0.6672  & 0.4913 & 0.6505 \\
 & CMAE & 0.0865 & 0.1065  & 0.1078 & 0.1674 \\
& BMAE & 0.1614 & 0.1847  & 0.1589 & 0.1941 \\
\hline
\multirow{4}{*}{{FPN-SegNet}} & MAE & 0.1503 & 0.1510  & 0.0931 & 1.1305 \\
 & M-CAE & 0.4059 & 0.4631  & 0.2941 & 4.5548 \\
 & CMAE & 0.1429 & 0.1393  & 0.0820 & 0.8288 \\
& BMAE & 0.1654 & 0.1656  & 0.1502 & 2.7322 \\
\hline
\multirow{4}{*}{{UNet-UNet}} & MAE & 0.1265 & 0.1540  & 0.1146 & 0.1979 \\
 & M-CAE & 0.3283 & 0.5230  & 0.3035 & 0.6429 \\
 & CMAE & 0.0703 & 0.0983  & 0.0700 & 0.2277 \\
& BMAE & 0.1582 & 0.1587  & 0.1250 & 0.2297 \\
\hline
\end{tabular}
\end{center}
\end{table}

\section*{References}

\bibliography{references}

\end{document}